\RequirePackage{fix-cm}%
\documentclass[twocolumn]{svjour3}%
\smartqed  % flush right qed marks, e.g. at end of proof

\usepackage[table,usenames,dvipsnames]{xcolor}%
\usepackage[utf8]{inputenc}%
\usepackage[T1]{fontenc}%
\usepackage{graphicx}%
\usepackage[english]{babel}%

\usepackage{epsfig}%
\usepackage{amssymb}%
\usepackage{mathtools}%
\usepackage[normalem]{ulem}%
\usepackage{pifont}%
\usepackage{tikz}%
\usepackage{natbib}%
\usepackage{multirow}%                  %% multirow/multicol cells in tables
\usepackage[strings]{underscore}%       %% get rid of stupid "missing $" errors for text with underscores
\usepackage{soul}%                      %% strikeout text using \sout{...}
\usepackage{xspace}%
\usepackage{array}%

\usepackage[pagebackref=true,breaklinks=true,colorlinks,bookmarks=false]{hyperref}
\hypersetup{citecolor=red} %Cites red (evil)
\hypersetup{linkcolor=green} %Local refs green (good)
\hypersetup{pdftitle={What Makes Good Synthetic Training Data for Learning Disparity and Optical Flow Estimation?},
            pdfauthor={Nikolaus Mayer, Eddy Ilg, Philipp Fischer, Caner Hazirbas, Daniel Cremers, Alexey Dosovitskiy, Thomas Brox}}

\newcolumntype{P}[1]{>{\centering\arraybackslash}p{#1}}%

\definecolor{grey}{rgb}{0.6,0.6,0.6}%

\newcommand{\printasis}[1]{\texttt{\detokenize{#1}}}%

\newcommand{\my}{\ding{51}}%
\newcommand{\mn}{}%

\newcommand{\pz}{\phantom{0}}%

\def\ijcvhruletop{\hline\noalign{\smallskip}}%
\def\ijcvhrulemid{\noalign{\smallskip}\hline\noalign{\smallskip}}%
\def\ijcvhrulebot{\noalign{\smallskip}\hline}%

\makeatletter
\DeclareRobustCommand\onedot{\futurelet\@let@token\@onedot}
\def\@onedot{\ifx\@let@token.\else.\null\fi\xspace}
\def\eg{\emph{e.g}\onedot}%
\def\ie{\emph{i.e}\onedot}%
\journalname{IJCV VISI}

\begin{document}

\title{What Makes Good Synthetic Training Data for Learning Disparity and Optical Flow Estimation?\thanks{This is a pre-print of an article published in the International Journal of Computer Vision. The final authenticated version is available online at: https://doi.org/10.1007/s11263-018-1082-6.\newline We acknowledge funding by the ERC Starting Grant ``VideoLearn'', the ERC Consolidator Grant ``3D Reloaded'', the DFG Grant BR-3815/7-1, the DFG Grant CR 250/17-1, and the EU Horizon2020 project ``TrimBot2020''. We thank Benjamin Ummenhofer for code that kick-started the creation of our 3D datasets.}}


\author{Nikolaus Mayer \and
        Eddy Ilg \and 
        Philipp Fischer \and 
        Caner Hazirbas \and
        Daniel Cremers \and
        Alexey Dosovitskiy \and
        Thomas Brox 
}


\institute{Nikolaus Mayer \and 
           Eddy Ilg \and 
           Philipp Fischer \and 
           Alexey Dosovitskiy \and 
           Thomas Brox \at
              University of Freiburg \\ 
              \email{\{mayern,ilg,fischer,dosovits,brox\}@cs.uni-freiburg.de}       %  \\
           \and
           Caner Hazirbas \and Daniel Cremers \at
               Technical University of Munich \\
               \email{\{hazirbas,cremers\}@in.tum.de}
}

\date{Received: date / Accepted: date}

\maketitle

\setlength{\fboxsep}{-0.5pt}%
%% TODO: understand why the enclosing {} are necessary
{\setlength{\fboxrule}{1pt}}%

\begin{abstract}
The finding that very large networks can be trained efficiently and reliably has led to a paradigm shift in computer vision from engineered solutions to learning formulations. 
As a result, the research challenge shifts from devising algorithms to creating suitable and abundant training data for supervised learning. How to efficiently create such training data?
The dominant data acquisition method in visual recognition is based on web data and manual annotation. Yet, for many computer vision problems, such as stereo or optical flow estimation, this approach is not feasible because humans cannot manually enter a pixel-accurate flow field. 
In this paper, we promote the use of synthetically generated data for the purpose of training deep networks on such tasks.
We suggest multiple ways to generate such data and evaluate the influence of dataset properties on the performance and generalization properties of the resulting networks. 
We also demonstrate the benefit of learning schedules that use different types of data at selected stages of the training process. 
  
\end{abstract}

\section{Introduction}
\label{sec:introduction}

Since the beginning of research in artificial intelligence, there has been a constant shift from engineering of solutions for special scenarios towards methodologies that are more generally applicable. 
The increased popularity of learning methods, which already started before the success of deep learning, is another chapter in this progress towards more generic approaches. 
These methods allow restricting the engineering on the algorithmic side, but at the same time much more influence is given to the data.
While classical computer vision methods have not required any training data, in which sense they are unsupervised methods\footnote{Clearly, the engineering and optimization of tuning parameters requires some training data. For a long time, test data was misused for this purpose.}, the final performance of approaches following the dominant supervised learning paradigm depends very much on the size and quality of training datasets. 
Especially the large potential of deep learning can only be fully exploited with large datasets.

Significant research efforts have been made to create such datasets, \eg ImageNet~\citep{imagenet}, MS COCO~\citep{mscoco}, CityScapes~\citep{cityscapes}, or the NYU dataset~\citep{nyu2}. These datasets have enabled most of the progress in computer vision in recent years. Notably, all these datasets cover the field of visual recognition such as object detection and classification. This has been one of the reasons why deep learning in computer vision has mainly focused on visual recognition for a long time. A notable exception is the NYU dataset. It contains a large collection of RGB-D images and enabled the seminal work of~\citet{eigen2014depth} on depth prediction from a single image. 

In this paper, we present and analyze one of the very first approaches to create training data for learning optical flow and disparity estimation---two classical computer vision tasks---on a large scale. Creating data for such tasks requires a different approach than in visual recognition, where web data in conjunction with manual annotation by mostly untrained persons can yield large datasets with reasonable time and monetary effort. In case of optical flow estimation, for instance, it is not obvious how to derive ground truth for large sets of real videos. Notable works include the Middlebury dataset~\citep{Baker-et-al-11} and the KITTI datasets~\citep{kitti2012,kitti2015}, but due to the difficulty of computing accurate ground truth, these have been restricted to $8$ and $2\cdot200$ pairs of frames, respectively (the latter are also limited by the special driving setting). This is far from what is needed to train a powerful deep network. Therefore, these datasets have been used mainly as test sets for classical, non-learning-based methodology. The situation is similar for disparity estimation from stereo pairs, or depth estimation and camera tracking from monocular videos. 

The MPI-Sintel dataset demonstrated the feasibility of using existing data from an open source movie to render videos together with the desired ground truth~\citep{Butler-et-al-12}. Sintel provides ground truth for optical flow, disparities and occlusion areas, and has been largely adopted as a serious benchmark dataset despite its synthetic nature. Still, the dataset is small by deep learning standards: The training set consists of $1041$ image pairs -- which is sufficient to train a network, but not a very powerful one, as we show in this paper. Moreover, the approach does not scale arbitrarily due to the limited availability of open source movies.

Inspired by the Sintel dataset, we take the approach of rendering images together with various ground truth outputs to a new level.
Embracing procedural generation and abstract instead of naturalistic data, we focus on training rather than testing and aim for a much larger scale. In contrast to the Sintel dataset, which was intended for benchmarking, this allows us to ignore many subtle problems that appear in the rendering of synthetic data~\citep{sintel-workshop}, and rather focus on the size and diversity of the dataset. 

There are many ways to generate training data in a synthetic manner: using existing scene data as in Sintel, manually designing new scenes, or creating randomized scenes in a procedural manner. In this paper, we investigate which of these options are most powerful and which properties of a dataset are most important for training a network that will generalize also to other data, particularly real data. We focus this investigation on networks that are trained for optical flow estimation, specifically FlowNet~\citep{flownet}, and disparity estimation, specifically DispNet~\citep{dispnet}. 

The study leads to many interesting findings. For instance, the data does not have to be realistic to make for a good training dataset. In fact, our simplistic 2D FlyingChairs dataset yields good data to start training a network for optical flow estimation.
Moreover, we find that:
\begin{itemize}
  \item[\textbullet] multistage training on multiple separate datasets works better than not only either one of the datasets by itself, but also than a full mix of both,
  \item[\textbullet] enabling more realism in the data via complex lighting does not necessarily help even if the test data is realistically lit and
  \item[\textbullet] simulating the flaws of a real camera during training improves network performance on images from such cameras.
\end{itemize}

\rowcolors{2}{gray!20}{white}%
\begin{table*}
  \centering
  %%
  %% ORIGINAL
  %%
  \begin{tabular}{ll|ccccrr}%
    \ijcvhruletop
    
     Dataset         & Published in    &  \!Synthetic/Natural\! & \!Flow\! & \!Depth\! & \!Stereo\! & \#Frames & \!Resolution \\ 
      
    \ijcvhrulemid
      UCL & \citet{UCL}                             &  S  & \my &     &     & $\phantom{100,\!00}4$ & $640\times480$ \\
      Middlebury &\citet{Baker-et-al-11}            & S/N & \my &     &     & $\phantom{100,\!00}8$ & $640\times480$ \\ 
      KITTI 2012 & \citet{kitti2012}                &  N  & \my & \my & \my & $\phantom{100,}194$   & $1,\!242\times375$ \\
      NYU v2 & \citet{nyu2}                         &  N  & \mn & \my & \mn & $408,\!473$           & $640\times480$ \\
      Sintel & \citet{Butler-et-al-12}              &  S  & \my &     & \my & $\phantom{10}1,\!064$ & $1,\!024\times436$ \\
      TUM    & \citet{tum}                          &  N  & \mn & \my & \mn & \!$\sim100,\!000$     & $640\times480$ \\
      UCL (extended) & \citet{UCL-extended}         &  S  & \my &     &     & $\phantom{100,\!0}20$ & $640\times480$ \\
      Sun3D & \citet{sun3d}                         &  N  & \mn & \my &     & \!$\sim2,\!500,\!000$ & $640\times480$ \\
      Middlebury 2014 & \citet{scharstein2014high}  &  N  & \mn & \my &     & $\phantom{100,\!0}23$ & \!$\sim6$MP \\
      KITTI 2015 & \citet{kitti2015}                &  S  & \my & \my & \my & $\phantom{100,}200$   & $1,\!242\times375$ \\ 
      FlyingChairs & \citet{flownet}                &  S  & \my & \mn & \mn & $\phantom{1}21,\!818$ & $512\times384$ \\
      FlyingThings3D & \citet{dispnet}              &  S  & \my & \my & \my & $\phantom{1}22,\!872$ & $960\times540$ \\
      Monkaa & \citet{dispnet}                      &  S  & \my & \my & \my & $\phantom{10}8,\!591$ & $960\times540$ \\
      Driving & \citet{dispnet}                     &  S  & \my & \my & \my & $\phantom{10}4,\!392$ & $960\times540$ \\
      Virtual KITTI & \citet{virtualkitti}          &  S  & \my & \my & \mn & $\pz21,\!260$         & $1,\!242\times375$ \\
      SYNTHIA & \citet{synthia}                     &  S  &     & \my & \my & \!$\sim200,\!000$     & $960\times720$ \\
      ScanNet & \citet{scannet}                     &  N  & \mn & \my & \mn & \!$\sim2,\!500,\!000$ & $640\times480$ \\
      SceneNet RGB-D & \citet{scenenet-rgbd}        &  S  & \my & \my & \mn & \!$\sim5,\!000,\!000$ & $320\times240$ \\
    \ijcvhrulebot
  \end{tabular}%
  \caption{\textbf{Datasets for optical flow and depth estimation.} The multitude and variation among the datasets makes choosing one nontrivial. This overview is ordered by date of publication. Many datasets focus on specific settings or features.}%
  \label{tab:datasets}%
\end{table*}
\rowcolors{2}{white}{white}%

\section{Related Work} 

Synthetic, rendered data has been used for benchmarking purposes for a long time. 
\citet{heeger1987} and~\citet{barron-et-al} used virtual scenes (including Lynn Quam’s famous \emph{Yosemite} 3D sequence) to quantitatively evaluate optical flow estimation methods. 
\citet{mccane-et-al} composed images of objects onto backgrounds and varied the complexity of scenes and motions to quantify how these changes affect estimation accuracy. 
Complementary to these are the works of~\citet{otte-nagel} who recorded real scenes with simple geometry and camera motion such that the ground truth flow could be annotated manually, and~\citet{meister2011real} who acquired ground truth for a real scene by recreating it as a virtual 3D model.
The Middlebury flow dataset~\citep{Baker-et-al-11} contains both real and synthetic scenes. 
\citet{vaudrey2008differences} created synthetic sequences for benchmarking scene flow estimation.
Somewhere between synthetic and real are datasets such as \citet{cutpastelearn} who created collages from real photographs to get training data for instance detection.

\begin{figure*}[!h]%
  \begin{tabular}{cccc}%
      \includegraphics[width=.24\linewidth]{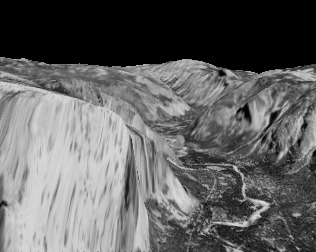}
    & \includegraphics[width=.24\linewidth]{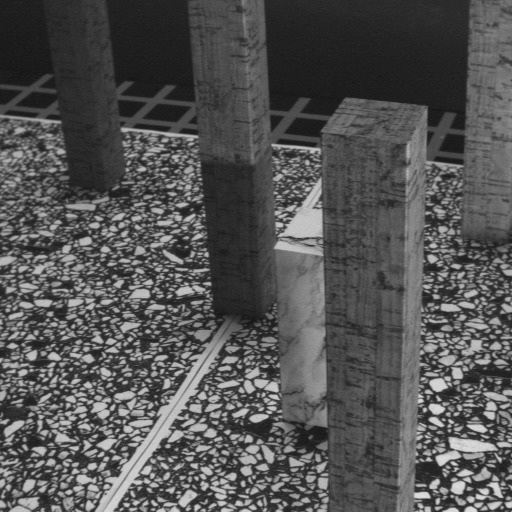}
    & \includegraphics[width=.24\linewidth]{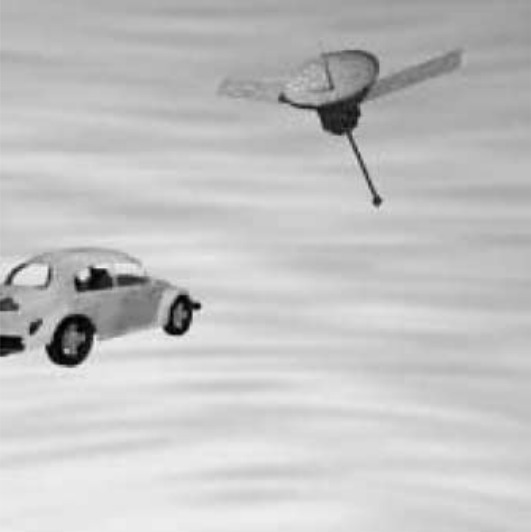}
    & \includegraphics[width=.24\linewidth]{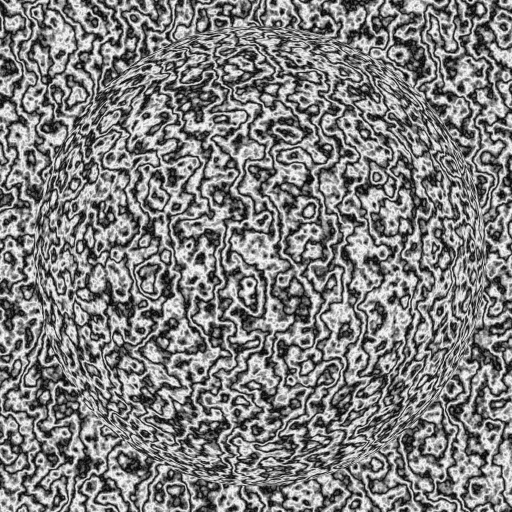} \\
      “Yosemite” (synth.)
    & “Marbled-Block” (real)
    & “Medium complexity” (synth.) 
    & “Split sphere” (synth.) \\
      \citet{barron-et-al} 
    & \citet{otte-nagel}
    & \citet{mccane-et-al}
    & \citet{HD07} \\[2mm]
      \includegraphics[width=.24\linewidth]{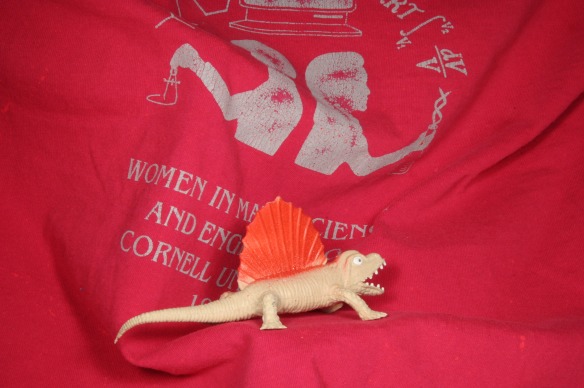}
    & \includegraphics[width=.24\linewidth]{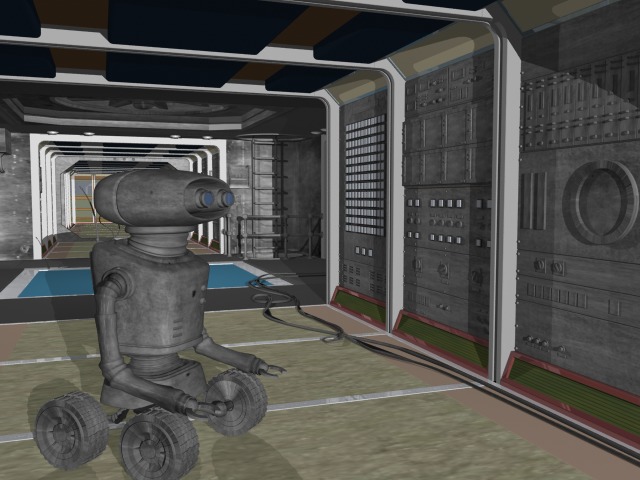}
    & \includegraphics[width=.24\linewidth]{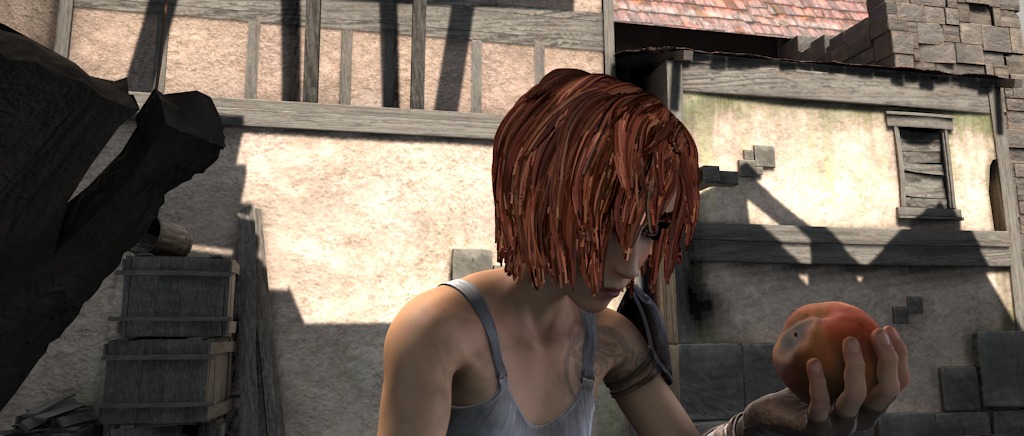}
    & \includegraphics[width=.24\linewidth]{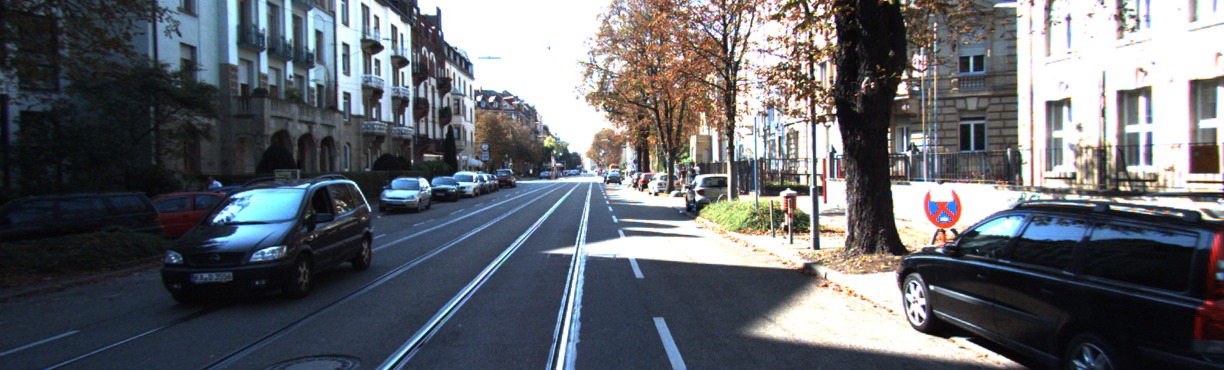} \\
      “Middlebury” (real+synth.)
    & “UCL Dataset” (synth.)
    & “Sintel” (synth.)
    & “KITTI 2015” (real) \\
      \citet{Baker-et-al-11}
    & \citet{UCL-extended}
    & \citet{Butler-et-al-12}
    & \citet{kitti2015} \\[2mm]
    
      \includegraphics[width=.24\linewidth]{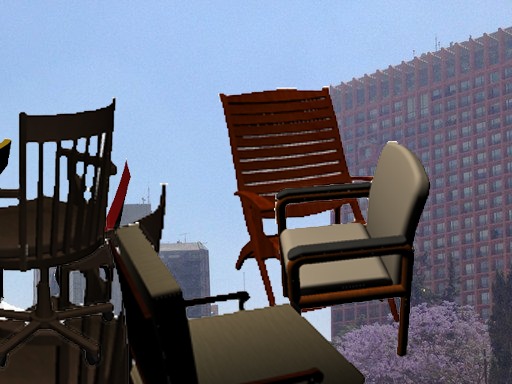}
    & \includegraphics[width=.24\linewidth]{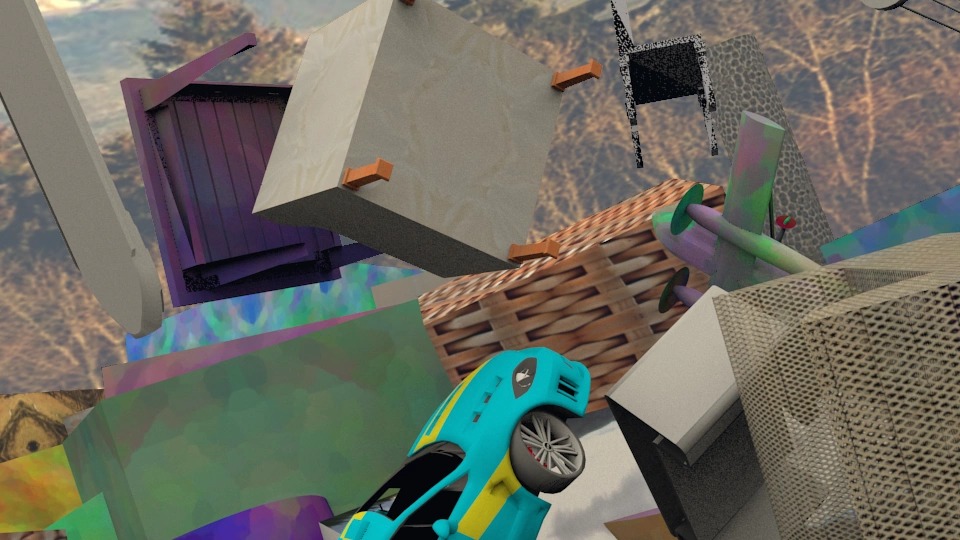}
    & \includegraphics[width=.24\linewidth]{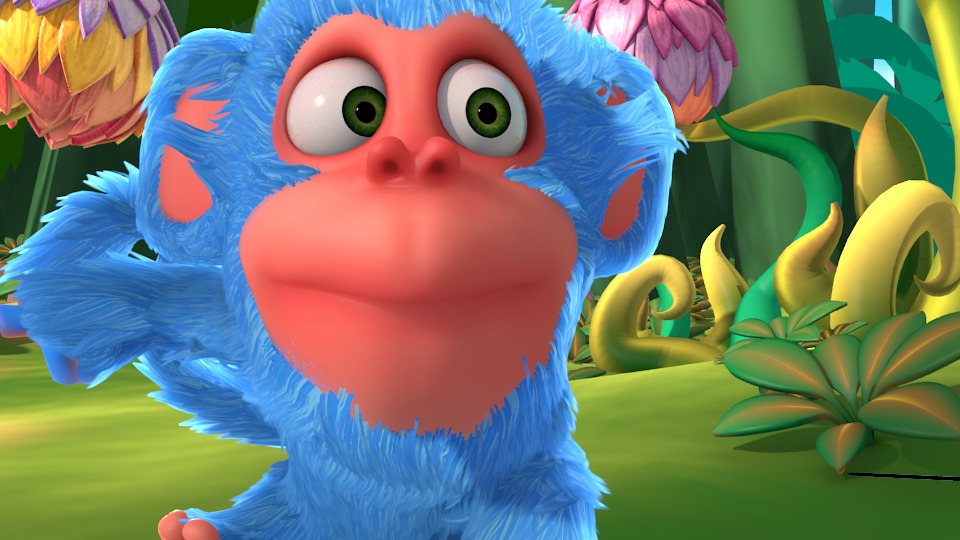}
    & \includegraphics[width=.24\linewidth]{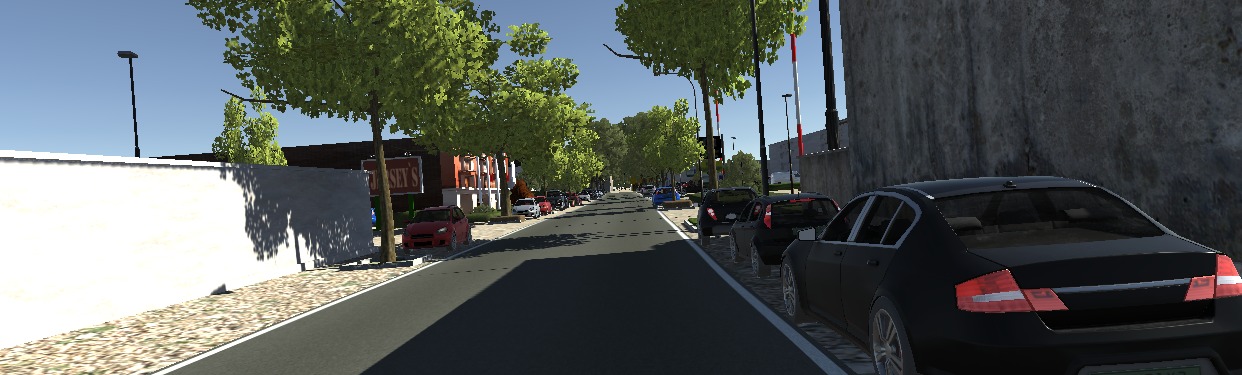} \\
      “FlyingChairs” (synth.)
    & “FlyingThings3D” (synth.)
    & “Monkaa” (synth.)
    & “Virtual KITTI” (synth.) \\
      \citet{flownet}
    & \citet{dispnet}
    & \citet{dispnet}
    & \citet{virtualkitti} \\
  \end{tabular}%
  \caption{\textbf{Samples from real and synthetic datasets with optical flow annotation.} % 
  Real datasets are small and restricted in diversity due to the difficulty to derive ground truth optical flow. Synthetic data can be based on manual scene modeling, such as the open-source Sintel and Monkaa movies, or procedural generation, such as the FlyingChairs and FlyingThings3D datasets. 
  }%
  \label{fig:datasets-gallery}%
\end{figure*}

Recent synthetic datasets include the UCL dataset by~\citet{UCL-extended} and the larger Sintel benchmark by \citet{Butler-et-al-12} where the 3D software \textit{Blender} was used to obtain dense and accurate flow and disparity ground truth for complex rendered scenes.
\citet{speed-and-texture} used the comparable 3D software \textit{Maya} to render a small driving-specific dataset for their study on optical flow accuracy. Table~\ref{tab:datasets} gives an overview of datasets for optical flow and depth estimation and their technical properties. Examples for most of them are shown in Fig.~\ref{fig:datasets-gallery}.

By hooking into the popular Unreal game engine, \citet{unrealcv} enabled data generation using existing assets made by game content creators.
Their technique was used by \citet{unrealstereo} to make a disparity evaluation dataset for difficult visual effects such as specularity.
The game engine approach was also taken by \citet{ovvv} to build test data for video surveillance systems, as well as by \citet{carla} to build the CARLA driving simulator.
Our work uses the concept of synthetic data, but focuses on large scale training data rather than benchmark datasets. Moreover, we present multiple ways of creating such data: using an open source movie as in~\citet{Butler-et-al-12}, manually building custom 3D scenes, and random procedural generation of scenes. 

Several recent and concurrent works have focused on building large scale synthetic training datasets for various computer vision tasks.
These datasets are typically specialized on narrow scenarios: automotive driving in the KITTI-lookalike Virtual KITTI dataset~\citep{virtualkitti}, the SYNTHIA dataset~\citep{synthia} and the dataset of~\citet{playingfordata}; indoor scenes in SceneNet~\citep{scenenet}, SceneNet RGB-D~\citep{scenenet-rgbd}, ICL-NUIM~\citep{ICL-NUIM} and SUNCG \citep{suncg}; isolated objects in ModelNet~\citep{modelnet} and ShapeNet~\citep{shapenet}; human action recognition in~\citet{proceduralvideogeneration}.
In contrast to these works, 
we are focusing on datasets which allow learning general-purpose optical flow and disparity estimation. 

There has been relatively little work on analyzing which properties of synthetic training data help generalization to real data.
\citet{renderforcnn} used rendered data to train a network on object viewpoint estimation; their results indicate better performance in real scenes when the training data contains real background textures and varying lighting.
\citet{Movshovitz-Attias2016} investigated how lighting quality in rendered images affects the performance of a neural network on viewpoint estimation, and found that more realistic lighting helps.
\citet{Zhang2017} studied the influence of synthetic data quality on semantic segmentation.
In this work, we perform an in-depth analysis of synthetic datasets for optical flow estimation, varying not only the lighting conditions, but also the shapes, motions and textures of the objects in the scene.

The present paper consolidates and extends three earlier conference papers~\citep{flownet,dispnet,flownet2} with the focus on training data. In the conference papers, datasets with 2D shapes (FlyingChairs,~\citet{flownet} and ChairsSDHom,~\citet{flownet2}) and more sophisticated datasets with 3D objects~\citep{dispnet} were introduced and used to train CNNs for optical flow and disparity estimation. In this work, we provide a principled study from simplistic to more sophisticated training data and analyze the effects of data properties on network training and generalization.

\section{Synthetic Data}

\setlength{\tabcolsep}{2pt}%
\renewcommand{\arraystretch}{1}%

Supervised training of deep neural networks requires large amounts of data.
For visual recognition, large training datasets can be acquired from web data with manual annotations, such as class labels, bounding boxes, or object outlines. 
For dense low-level tasks, such as optical flow or depth estimation, manual annotation is not feasible, especially when exact and pixel-accurate ground truth is desired.
This motivates the use of synthetic data, \ie data not taken from the real world but artificially generated: the key idea is that we have perfect knowledge and control of the virtual “world” from which we create our image data, and that we are thus able to produce perfect ground truth annotations along with the images. The difficulties are that creating photo-realistic image data is not trivial, and that it is not obvious which aspects of the real world are relevant and must be modeled (these problems arise because we are unable to perfectly simulate the real world).

This section recapitulates our prior works’ discussions about synthetic training data generation and data augmentation. 
We describe two ways to generate synthetic data: procedural randomization and manual modeling. Using our randomized 2D and 3D datasets as examples, we first discuss how starting off with a simple general scene and combining it with randomly sampled dynamic elements can yield datasets that are abstract but powerful and potentially of unlimited size. The dynamic elements are motivated by what the dataset should be able to teach, \eg optical flow due to object rotation.

Following this, we discuss manual data modeling in which more specialized datasets are created. This can be achieved either by rendering existing scenes (an approach used for the Sintel dataset of \citet{Butler-et-al-12} and our own Monkaa dataset \citep{dispnet}), or by manually recombining objects from existing scenes to get new, specially designed scenes\footnote{There is also the extreme case of handcrafting entire scenes and all their objects. We do not consider this option here because it yields even far less data for the same effort.}. We used the latter method in our Monkaa and Driving datasets. Mixtures of both methods are also possible, as demonstrated by Virtual KITTI where \eg cars can be randomly modified or left out of a fixed scene \citep{virtualkitti}. Video game engines allow tackling this from yet another perspective: using a game means offloading the model and scene design to its developers. The mechanics of the game world can then be used to generate large amounts of data, within the constraints of the game. This was used \eg by \citet{playingfordata}, \citet{playingforbenchmarks} and \citet{carla}. This approach is an alternative to ours and not considered in this paper.

\subsection{Randomized modeling of data}

\subsubsection{2D image pairs for optical flow: FlyingChairs}

Optical flow estimation is the task of taking two images which show basically the same content with some variation (\eg moving objects or different camera poses) and determining the apparent pixel motion of each pixel from the first to the second image. It is a low-level, nonsemantic task; any two images between which correspondences can be determined is a valid input. This means that to learn this task, any image pair with computable pixelwise ground truth displacements can serve as training data.

\begin{figure*}%
  \begin{center}%
    \begin{tikzpicture}[%
        arr/.style={thick,>=stealth},%
      ]%
      \node[rectangle,text width=1.75cm,align=center] (rngch) at (-2.5,5.5) {random\\sampling};
      \node (dice) at (-2.5,6.1) {\includegraphics[scale=0.04]{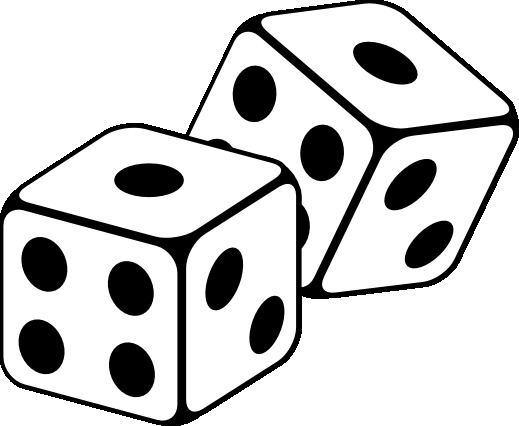}};
      \node[rectangle,text width=1.75cm,align=center] (rngbg) at (-2.5,0.0) {random\\sampling};
      \node (dice) at (-2.5,-0.6) {\includegraphics[scale=0.04]{images_flyingchairs-creation-schematics_dice.jpg}};
      
      \node (chproto) at (0.0,5.5) {\fbox{\includegraphics[scale=0.125]{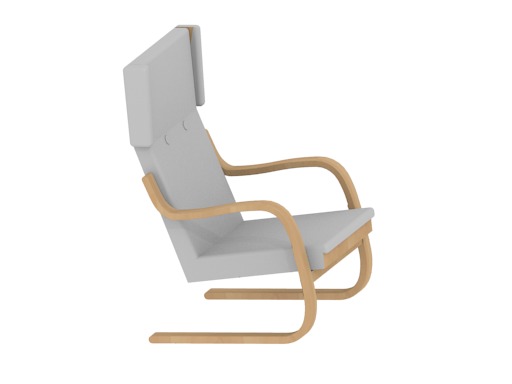}}};
      \node[text width=1.75cm,align=center] at (0,4.2) {object\\prototype};
      \node (bgproto) at (0.0,0.0) {\includegraphics[scale=0.125]{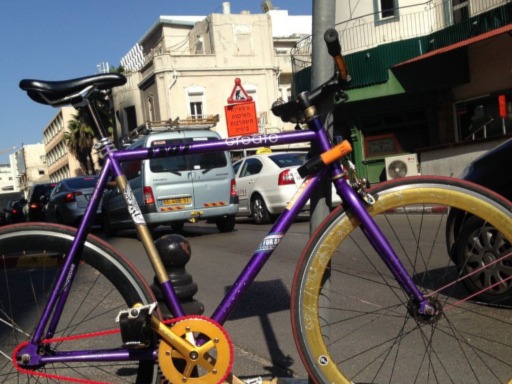}};
      \node[text width=1.75cm,align=center] at (0,-1.4) {background\\prototype};
      
      \draw[->,arr,shorten <=-2mm] (rngch) -- (chproto);
      \draw[->,arr,shorten <=-2mm] (rngbg) -- (bgproto);
      
      \node[rectangle,text width=1.75cm,align=center] (chopf0) at (3.0,5.5) {initial\\object\\transform};
      \draw[->,arr] (chproto) -- (chopf0);
      \node[rectangle,text width=1.75cm,align=center] (bgopf0) at (3.0,0.0) {initial\\background\\transform};
      \draw[->,arr] (bgproto) -- (bgopf0);
      
      \node (dice-chopf0) at (3.0,6.3) {\includegraphics[scale=0.04]{images_flyingchairs-creation-schematics_dice.jpg}};
      \node (dice-bgopf0) at (3.0,-0.8) {\includegraphics[scale=0.04]{images_flyingchairs-creation-schematics_dice.jpg}};
      
      \node (chf0) at (6.0,5.5) {\fbox{\includegraphics[scale=0.125]{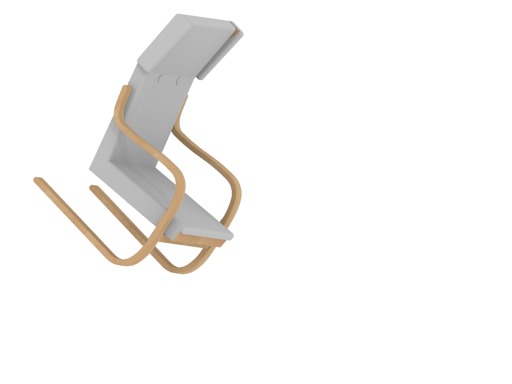}}};
      \draw[->,arr] (chopf0) -- (chf0);
      \node (bgf0) at (6.0,0.0) {\includegraphics[scale=0.125]{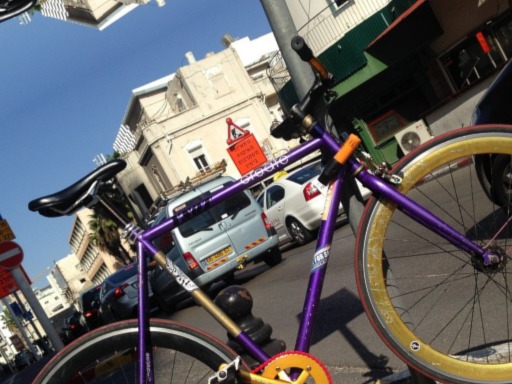}};
      \draw[->,arr] (bgopf0) -- (bgf0);

      \node[rectangle,text width=1.75cm,align=center] (chopf1) at (9.0,5.5) {object\\motion\\transform};
      \draw[->,arr] (chf0) -- (chopf1);
      \node[rectangle,text width=1.75cm,align=center] (bgopf1) at (9.0,0.0) {background\\motion\\transform};
      \draw[->,arr] (bgf0) -- (bgopf1);
      
      \node (dice-chopf1) at (9.0,6.3) {\includegraphics[scale=0.04]{images_flyingchairs-creation-schematics_dice.jpg}};
      \node (dice-bgopf1) at (9.0,-0.8) {\includegraphics[scale=0.04]{images_flyingchairs-creation-schematics_dice.jpg}};
      
      \node (chf1) at (12,5.5) {\fbox{\includegraphics[scale=0.125]{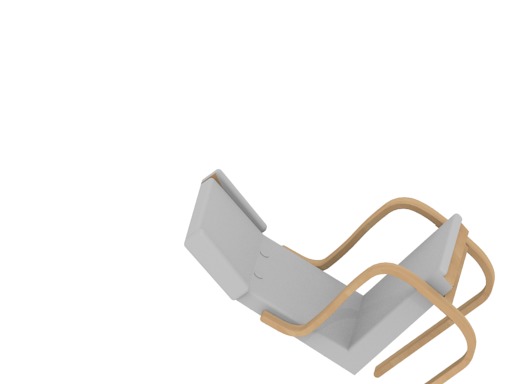}}};
      \draw[->,arr] (chopf1) -- (chf1);
      \node (bgf1) at (12,0.0) {\includegraphics[scale=0.125]{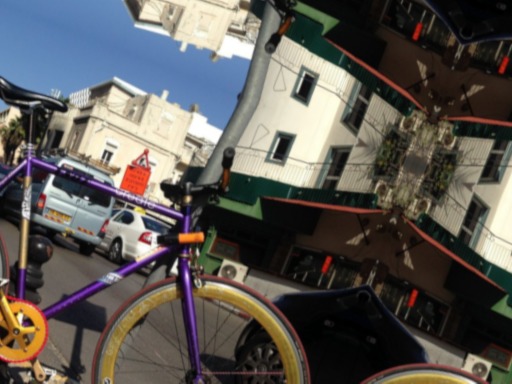}};
      \draw[->,arr] (bgopf1) -- (bgf1);

      \node (f0)   at (6.0,3) {\includegraphics[scale=0.125]{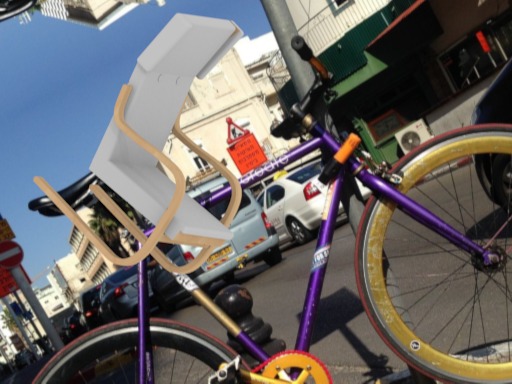}};
      \node        at (6.0,1.8) {first frame};
      \draw[->,arr] (chf0) -- (f0);
      \draw[->,arr,shorten >=5mm] (bgf0) -- (f0);

      \node (flow) at (9.0,3) {\includegraphics[scale=0.125]{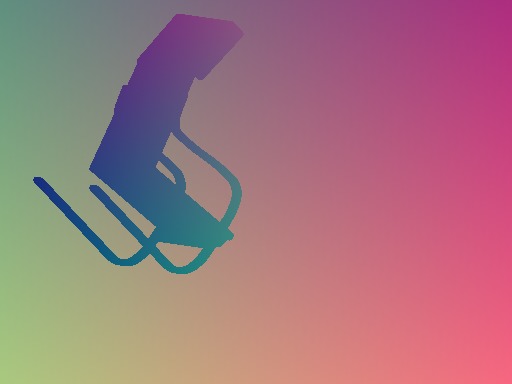}};
      \node        at (9.0,1.8) {optical flow};
      \draw[->,arr] (chopf1) -- (flow);
      \draw[->,arr,shorten >=5mm] (bgopf1) -- (flow);

      \node (f1)   at (12, 3)  {\includegraphics[scale=0.125]{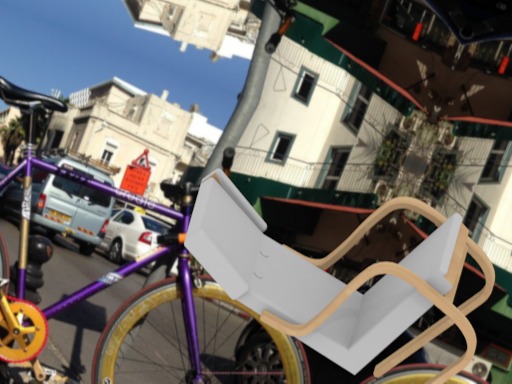}};
      \node        at (12,1.8) {second frame};
      \draw[->,arr] (chf1) -- (f1);
      \draw[->,arr,shorten >=5mm] (bgf1) -- (f1);

      \node[draw,rectangle,rounded corners=3mm,minimum width=12cm,minimum height=3cm] at (7.5,2.75) {};
      \node at (3.0,2.75) {Outputs:};
    \end{tikzpicture}%
  \end{center}%
  \caption{
    \textbf{FlyingChairs:} A two-frame sample is created by composing a foreground object onto backgrounds. Each object and background has an initial transform (to introduce scene variety), as well as a transform between frames which induces optical flow. All transforms are affine which makes computing the ground truth flow field easy. This schematic shows a simplification for a single foreground object; in our datasets there are multiple objects, each with an individual transform. Note that the foreground object’s transform is composed onto the one of the background, \ie the object will move “with the background” if its own transform is the identity. This correlates object motion with background motion and simulates optical flow induced by camera motion. Dice icons indicate randomization.
  }%
  \label{fig:flyingchairs_schema}%
\end{figure*}

A simple way of creating data with ground truth displacements is to take images of objects (segmented such that anything but the object itself is masked out) and paste them onto a background, with a randomized transformation applied from the first to the second image. 
Fig.~\ref{fig:flyingchairs_schema} illustrates how a FlyingChairs sample is created this way.
We used images of chairs created from CAD models by~\citet{Aubry-et-al-14}. 
We chose chairs because:  
(a) flying chairs offer nowhere near the semantic content of real-world scenes (thus, successful generalization to real data indicates that the estimation of point correspondences can be learned from abstract data) and (b) chairs have non-convex shapes with fine structure and are topologically diverse. 

We superimposed the chair images onto real-world background images of urban and landscape environments downloaded from Flickr. 
In more detail, to generate the first image of each sample pair, random affine transformations were applied to all objects, \ie the chairs were randomly scaled, rotated, and placed on the randomly transformed background.
To generate the second image, additional small random affine transformations were generated for all objects. 
These second transformations model the motion from the first to the second image. 
We simulate “camera motion” effects, \ie that objects tend to move with the scene, by composing the background’s motion onto each object’s own motion.
Since all transformation parameters are known, accurate ground truth optical flow is available at every single pixel, even in regions that become occluded in the second image. 
The resulting images resemble those from \citet{mccane-et-al} (\emph{cf.} Fig.~\ref{fig:datasets-gallery}), yet with the crucial difference that there are thousands of randomized scenes instead of a few manually created ones.

\subsubsection{Synthetic 3D data for optical flow, disparity and scene flow: FlyingThings3D}
\label{sec:flyingstuff3d}

The FlyingChairs were generated using simple 2D affine transformations. Since such datasets cannot contain information from depth in the scene, \ie 3D rotation and translation in space and camera motion, we took the randomization approach further and created FlyingThings3D: a 3D dataset rendered from true 3D models, with ground truth for stereo disparity, optical flow, and for the full scene flow task~\citep{dispnet}.

To obtain 3D scenes, we created a simple but structured background from simple geometric random shapes.
For the dynamic foreground objects, we used models from ShapeNet \citep{shapenet}.
All foreground objects follow linear trajectories in 3D space, and so does the camera.
To animate the camera’s viewing direction, an invisible moving object is added to the scene and the camera is constrained to always point or “look” at this object.
This 3D setting with combined object and camera motions allows for complex object flows and scene configurations. 

To render the 3D models into 2D images, we used the freely available Blender suite.
We modified its internal rendering engine to directly produce fully dense and accurate ground truth for depth, disparity, optical flow and object segmentation (among others), all for both views of a virtual stereo camera.

The FlyingThings3D dataset combines sophisticated 3D rendering with the procedural generation of scenes which allows for arbitrary amounts of data without manual effort. While the generated scenes are by no means realistic in the naturalistic sense of \eg Sintel, they allow for a large and diverse dataset. 

%%%%%%%%%%%%%%%%%%%%%%%%%%%%%%%%%%%%%%%%%%%%
\subsection{Manual modeling}

In the previous section, we presented approaches to generate large amounts of data automatically.
This section discusses the generation of synthetic data that involves manual engineering.
We use the 3D model rendering approach described in Section~\ref{sec:flyingstuff3d}. 

\subsubsection{Monkaa}

The publicly available open source movie Monkaa provides 3D scenes and assets that can be loaded into Blender.
We created a dataset that contains original scenes as well as new custom ones \citep{dispnet}.

In contrast to FlyingThings3D, data is generated from the movie in a deterministic way.
The original scenes and objects were modeled by 3D artists. For our custom scenes, we manually collected, composed and animated original set pieces and objects from the movie, producing entirely new environments and movements while keeping the visual style of the Monkaa movie.
To obtain a sufficient amount of data, we rendered longer scenes instead of procedurally generating many variations.
Contrary to the datasets mentioned above, Monkaa contains articulated non-rigid motion of animals and extremely complex fur.
Fig.~\ref{fig:datasets-gallery} shows a frame from our Monkaa dataset release.

During our work on Monkaa, we encountered many of the problems described by the creators of the Sintel benchmark dataset \citep{sintel-workshop}, such as changing focal length, scenes containing objects only in the exact locations viewed by the camera and optical tricks which break when using a stereo setup (\eg forced perspective or “fog” rendered as a flat 2D sprite). This greatly limited the amount of usable scenes and contributes to the fact that this approach to data generation cannot be scaled easily to produce more data.

\subsubsection{Driving}

Videos captured from a camera on a driving car provide a very special setting and usually differ significantly from other video material.
This is demonstrated well in the KITTI benchmark suite \citep{kitti2012}.
Wide-angle lenses are typically used to cover large areas, and the scene motion is dominated by the forward camera motion of the driving car. 
Most of the scene is static and the arrangement of objects in the scenes is very similar, with sky at the top, the road at the bottom and mostly walls and parked vehicles on the sides. 
These motion and object types are not covered in the previously presented data and hence require special treatment. 
To this end, we created a dataset that represents this setting and resembles the typical aspects of a real dataset like KITTI.
The 3D scenes contain simple blocks and cylinders to imitate buildings, and models of cars, trees, and street lamps taken from 3D Warehouse\footnote{\printasis{https://3dwarehouse.sketchup.com/}} and the ShapeNet database.
The stereo baseline, the focal length and properties such as scene lighting and object materials were chosen to match the KITTI dataset.
The virtual camera and other cars in the scene were manually animated into a simple urban traffic scenario. While the dataset does not make much sense on a high level (there are no road signs, traffic light situations, or pedestrians crossing the road), it does replicate the rough semantic structure of a typical KITTI scene.
The experiments in this paper confirm the intuition that fine-tuning on this data should teach a network useful priors, \eg a flat road surface at the bottom, and thus largely improve the performance on KITTI compared to training on data without such priors.

\setlength{\tabcolsep}{2pt}%
\renewcommand{\arraystretch}{1}%
\begin{figure*}[!h]%
  \includegraphics[width=.99\linewidth]{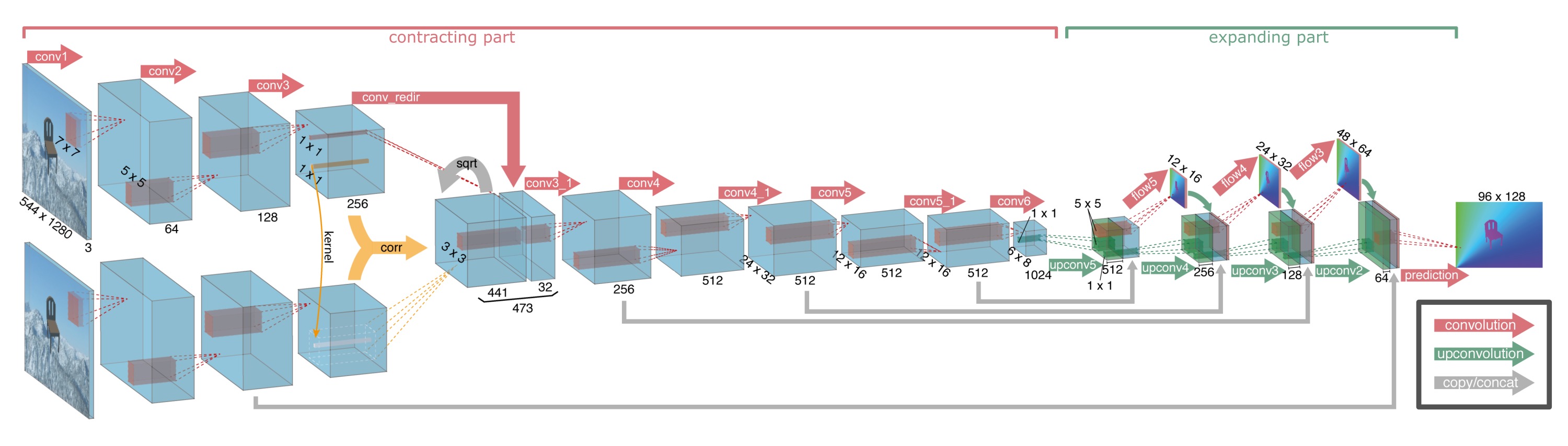}%
      \caption{\textbf{FlowNetC architecture} from \citet{flownet}. The figure shows the full 15-layer architecture with contracting and expanding part. The gray skip connections between both parts allow for high-resolution predictions without passing high-resolution information through the bottleneck. We use the same architecture with fewer feature channels in our optical flow experiments. The DispNetCorr1D from \citet{dispnet} (which we use for our disparity experiments in Section~\protect\ref{sec:datadegradation}) implements the same idea, but with only a 1D correlation.}
  \label{fig:flownet-arch}%
\end{figure*}%

\subsection{Data augmentation}

The term “augmentation” refers to artificially increasing the amount of data that is presented while training a machine learning algorithm, with the goal of improving generalization capabilities.
This is done by transforming the original data in small, incremental ways.
It relies on the assumption that the learning algorithm is unable to filter out the small changes and instead perceives the modified data as actual \textit{new} data with new information.
While our 2D and 3D datasets offer much more training data than had previously been available, the use of data augmentation within the training process is much more efficient: additional data need not be stored and read from disk during training but can be generated on the fly.
We split our augmentation options into color and geometry augmentations.
Our experiments indicate that the two types offer complementary benefits.

To train optical flow networks we used color augmentation (changing brightness, contrast and colors and adding color noise).
Options for geometric changes here include every change for which we can easily and directly compute an accurate ground truth flow field: shift, rotation and scaling.
We can simply apply the same transformation to both input images, or additionally apply a second transformation to only one of the images.
The latter can produce a greater variety of augmentations, \eg by inducing a scaling motion between frames.
Note that any of the described geometry augmentations still yields a valid optical flow training sample by simply composing the computed flow induced by the augmentation operation with the existing flow of the original sample.

For our disparity networks, we use the same color augmentations as for optical flow.
However, disparity estimation for stereo cameras is a more geometrically constrained setting than optical flow: the transformation between the two cameras is fixed, and \eg rotating both views of a sample pair or zooming one view relative to the other would disturb the epipolar geometry.
This means that for disparity the setting is more restricted and fewer augmentation options are available.

\section{Learning Tasks}

\subsection{Optical Flow Estimation}

The FlowNet was introduced by \citet{flownet}.
This network was the first to be trained end-to-end for optical flow estimation.
It was designed to receive two images of the video in full resolution as input and generate the corresponding optical flow field as output.

In \citet{flownet}, two architectures were proposed, FlowNetS and FlowNetC (see Fig.~\ref{fig:flownet-arch}), which share a common underlying idea.
The FlowNetS input consists of two RGB input images stacked into a six-channel input blob.
The processing starts with a contracting part that compresses the spatial image information into feature blobs with lower resolution but more feature channels. The subsequent expanding part uses up-convolutions to successively increase the resolution until it reaches the desired output size.
Skip-connections transfer higher resolution information directly from a resolution level of the contracting part to its corresponding resolution in the expanding part, thus this information does not have to pass through the architecture’s bottleneck. 

Each convolutional block consists of a convolution and a ReLU nonlinearity.
The filter size decreases while the number of feature maps increases as shown in Fig.~\ref{fig:flownet-arch}.
For spatial reduction of resolution, each convolution of the contracting part is strided with a factor of~2.
In the expanding part we use up-convolutions with stride 2 to increase the resolution again.
After each expansion step, an endpoint error (EPE) loss aids the network with additional training gradients in a deep-supervision manner.
For more details, we refer to \citet{flownet}.

The FlowNetC variant of this architecture (shown in Fig.~\ref{fig:flownet-arch}) employs a custom layer which computes the correlation scores between patches from two input feature streams. In this variant, the two input images are processed separately in the first three layers, but with shared weights, \ie as a Siamese network. The resulting features are fused using the correlation layer and then processed further as in the FlowNetS.

In \citet{flownet2}, we found that the FlowNetC consistently outperforms the FlowNetS. Therefore, unless mentioned otherwise, all experiments on optical flow in this paper are based on the FlowNetC architecture.\\
Specifically, we used “FlowNet2-c” \citep{flownet2}, a FlowNetC in which each layer’s number of channels was reduced to $\frac{3}{8}$ of the original size.
The reduced computational cost during training allows for the large set of evaluations in this paper. 
Unless specifically mentioned, we trained for $600$k mini-batch iterations with a batch size of $4$, following the $S_\text{short}$ learning rate schedule from~\cite{flownet2} (see also Fig.~\ref{fig:lr-schedules}): starting with learning rate $1\mathrm{e}{-4}$, then successively dropping to $5.0\mathrm{e}{-5}$, $2.5\mathrm{e}{-5}$, and $1.25\mathrm{e}{-5}$ at $300$k, $400$k, and $500$k iterations respectively.
Our experiments in Section~\ref{sec:learningschedules} use the $S_\text{long}$ and $S_\text{fine}$ schedules intended for pretraining and finetuning on different datasets.

\subsection{Disparity Estimation}
In our disparity experiments, we used the DispNet from \citet{dispnet} in its DispNetCorr1D variant.
Like FlowNetC, this network receives two images as input and initially processes them in two separate streams with shared weights (Siamese architecture).
The features at the end of both streams are correlated in a correlation layer and jointly processed further.
By exploiting the known and fixed epipolar geometry of rectified stereo pairs, the correlation is reduced to unidirectional 1D on horizontal scanlines.
Thus, compared to a FlowNet with a 2D correlation layer, the DispNetCorr1D can process a much wider receptive field at the same computational cost.

\section{Experiments}

\subsection{Testing environments}

In this paper, we aim for the analysis of relevant dataset properties that make a good dataset for training large networks for optical flow and disparity estimation. 
These might not be the same properties that make a good test set for benchmarking. 
Since benchmarking is not the aim of this paper, we use the established benchmark datasets Sintel~\citep{Butler-et-al-12} (the “clean” variant) and KITTI 2015~\citep{kitti2015} to measure the performance of networks trained on different data. 
While the evaluation on Sintel covers the generalization of the network to other synthetic datasets, the KITTI experiments cover the generalization to real-world data in the restricted setting of a driving scenario. 
In all cases we report the average endpoint error (EPE).

\rowcolors{3}{white}{gray!20}%
\setlength{\tabcolsep}{4pt}%
\renewcommand{\arraystretch}{1}%
\begin{table}
  \centering
  \begin{tabular}{l|P{1.5cm}P{1.5cm}P{1.5cm}}%
    \ijcvhruletop
    
                   & \multicolumn{3}{c}{Test data} \\[1.5mm]
    Training data  &  Sintel  &   KITTI2015   & FlyingChairs \\
    
    \ijcvhrulemid
    
    Sintel         &           $\pz6.42$  &          $18.13$ &          $5.49$ \\  %% mayern-datagen--sintelttrainsplit-testnet--04bb
    FlyingChairs   &  $\pz\mathbf{5.73}$  &          $16.23$ & $\mathbf{3.32}$ \\  %% mayern-datagen--chairs-testnet--04bb
    FlyingThings3D &           $\pz6.64$  &          $18.31$ &          $5.21$ \\  %% mayern-datagen--fs3d-testnet--04bb
    Monkaa         &           $\pz8.47$  &          $16.17$ &          $7.08$ \\  %% mayern-datagen--monkaa-testnet--04bb
    Driving        &             $10.95$  & $\mathbf{11.09}$ &          $9.88$ \\  %% mayern-datagen--driving-testnet--04bb
    
    \ijcvhrulebot
  \end{tabular}%
  \caption{\textbf{FlowNet trained on existing synthetic datasets.} Sintel and FlyingChairs were split into a training and a validation set. As expected, training on the Driving dataset works best for KITTI. The Sintel dataset, although very similar to its validation set, is too small to yield great performance. Surprisingly, the FlyingChairs datasets yields consistently better numbers than the more sophisticated FlyingThings3D and Monkaa datasets. This observation motivates our investigation of what makes a good dataset.}%
  \label{tab:introexperiment}%
\end{table}
\rowcolors{2}{white}{white}%

\subsection{What makes a good training dataset? \label{sec:good_training_dataset}}

In \citet{flownet}, we showed that the optical flow estimation task does not have to be trained on data which semantically matches the test data: there are no chairs in the Sintel dataset, but the network trained on the FlyingChairs dataset performs well on Sintel. In fact, Table~\ref{tab:introexperiment} shows that a network trained on the $22$k samples of the training split of the FlyingChairs dataset performs better than a network trained on a subset of the Sintel training dataset with $908$ samples and tested on the remaining $133$ samples for validation.

This positive result from \citet{flownet} does not yet clearly indicate what are the relevant dataset properties, apart from its size. 
Interestingly, Table~\ref{tab:introexperiment} also shows that training on the more diverse and more realistic FlyingThings3D dataset yields inferior performance to training on FlyingChairs. This is surprising and motivates our efforts to get more insights into: (1) which properties of the FlyingChairs dataset make it so successful for training optical flow networks and (2) how the training data can be potentially further improved.  

To this end, we performed an extensive ablation study to evaluate the contributions of object shape and types of motion. These aim primarily on explaining the generally good training behavior of the FlyingChairs dataset. 
Additionally, we tested the importance of surface features imparted by textures and the effect of lighting when lifting the FlyingChairs dataset into 3D, and what happens when combining the FlyingChairs and FlyingThings3D datasets.
This is to investigate the underlying reason why the FlyingChairs dataset outperforms the more sophisticated FlyingThings3D. 
This set of experiments was conducted on the optical flow task.

In a complementary step, focusing on real data, we looked at data characteristics that originate not in the observed scene but in the imaging system itself. In particular, we were interested in how explicit modeling of \eg camera lens distortion or Bayer artifacts in the training data may help improve the performance of the resulting network when applied to images from a real camera system showing these characteristics. 
For this set of experiments, we used the disparity estimation task because we found comparing disparity maps and assessing fine details by eye much easier than doing so on flow fields.

Due to time and compute power constraints, we could not evaluate both parts of our experiments suite on both optical flow and disparity estimation. Considering how closely related the two tasks are, the disparity results are presumably valid for optical flow as well, just as the optical flow results can likely be applied to disparity estimation.

\rowcolors{3}{gray!20}{white}%
\setlength{\tabcolsep}{4pt}%
\renewcommand{\arraystretch}{1}%
\begin{table*}
  \centering
  %%
  %% ORIGINAL
  %%
  \begin{tabular}{l|cccccccc|ccc}%
    \ijcvhruletop
    
    Training data
    & \rotatebox{90}{Polygons}
    & \rotatebox{90}{Ellipses}
    & \rotatebox{90}{Translation}
    & \rotatebox{90}{Rotation}
    & \rotatebox{90}{Scaling}
    & \rotatebox{90}{Holes in objects}
    & \rotatebox{90}{Thin objects}
    & \rotatebox{90}{Deformations}
    & \rotatebox{90}{Sintel train clean}
    & \rotatebox{90}{KITTI 2015 train}
    & \rotatebox{90}{FlyingChairs} \\
    
    \ijcvhrulemid

    %                     POLY    ROUND  TRA   ROT   SCAL  HOLE  THIN  WARP  SINTEL   KITTI     CHAIRS
      Boxes                                &\!(\my)\!&\mn&\my&\mn&\mn&\mn&\mn&\mn& $5.29$ & $17.69$ & $\!4.95$ \\ %A/1
      Polygons                                   &\my&\mn&\my&\mn&\mn&\mn&\mn&\mn& $4.93$ & $17.63$ & $\!4.60$ \\ %B/2
      Ellipses                                   &\mn&\my&\my&\mn&\mn&\mn&\mn&\mn& $4.88$ & $17.28$ & $\!4.87$ \\ %C/3
      Polygons+Ellipses                    &\my&\my&\my&\mn&\mn&\mn&\mn&\mn& $4.86$ & $17.90$ & $\!4.62$ \\ %D/8
      
    \ijcvhrulemid
      
      Polygons+Ellipses+Rotations          &\my&\my&\my&\my&\mn&\mn&\mn&\mn& $4.79$ & $18.07$ & $\!4.38$ \\ %E/4
      Polygons+Ellipses+Scaling            &\my&\my&\my&\my&\my&\mn&\mn&\mn& $4.52$ & $15.48$ & $\!4.22$ \\ %F/5
      Polygons+Ellipses+Holes in objects   &\my&\my&\my&\my&\my&\my&\mn&\mn& $4.71$ & $16.36$ & $\!4.20$ \\ %G/6
      Polygons+Ellipses+Thin objects       &\my&\my&\my&\my&\my&\my&\my&\mn& $4.60$ & $16.45$ & $\!4.13$ \\ %H/7
      Polygons+Ellipses+Deformations       &\my&\my&\my&\my&\my&\my&\my&\my& $\mathbf{4.50}$ & $\mathbf{14.97}$ & $\!4.23$ \\ %I/9
      
    \ijcvhrulemid
    
      FlyingChairs             &\!(\my)\!&\!(\my)\!&\my&\my&\my&\my&\my&\mn& $4.67$ & $16.23$ & $\!(\mathbf{3.32})$ \\
      
    \ijcvhrulebot
  \end{tabular}%
  \caption{\textbf{Object shape and motion.} Each row corresponds to one training set, containing certain object shapes and motion types. For reference, we also show results for a FlowNet trained on the FlyingChairs dataset \citep{flownet}. The trained networks are tested on three benchmark datasets. Even the simplest training dataset leads to surprisingly good results and adding more complex shapes and motions yields further improvement. Nonrigid deformations help on Sintel and KITTI, but not on the rigid FlyingChairs. On the other hand, the chairs contain holes whereas training with holes in objects weakens the results on Sintel and KITTI. Example images in Fig.~\protect\ref{fig:gallery-1} and Fig.~\protect\ref{fig:gallery-2} in the appendix.}%
  \label{tab:data-variation-types}%
\end{table*}
\rowcolors{2}{white}{white}%

\subsubsection{Object shape and motion}
\label{sec:shape-and-motion}%

In our first experiment, we investigated the influence of object shapes and the way they move.
The general setup here is similar to the FlyingChairs dataset---2D motions of objects in front of a background image---and we used the same composition approach, shown in Fig.~\ref{fig:flyingchairs_schema}. 
However, instead of chairs, we used different randomly shaped polygons and ellipses; 
and instead of arbitrary affine motions, we used different subsets thereof, plus optionally non-rigid deformations.
We used the same random Flickr images as in the FlyingChairs dataset, both for the background and for the foreground objects.
On every dataset variant we trained a network from scratch and evaluated its performance on three benchmark datasets: Sintel, KITTI and FlyingChairs.  

\begin{figure}[!h]%
  \begin{tabular}{ccc}%
      \includegraphics[width=.32\linewidth]{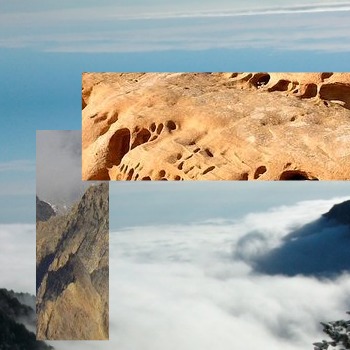}
    & \includegraphics[width=.32\linewidth]{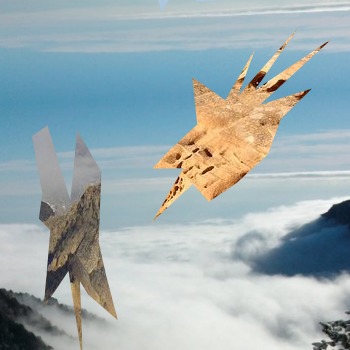}
    & \includegraphics[width=.32\linewidth]{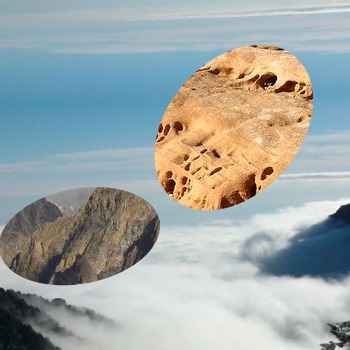} \\
    Boxes & Polygons & Ellipses \\[2mm]
    
      \includegraphics[width=.32\linewidth]{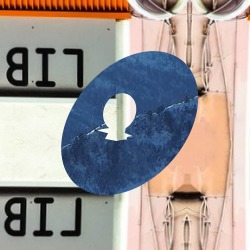}
    & \includegraphics[width=.32\linewidth]{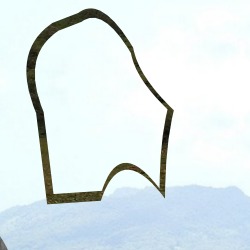}
    & \includegraphics[width=.32\linewidth]{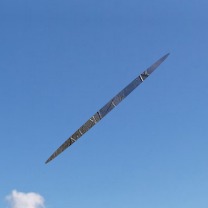} \\
    With holes & Outlines & Needles
  \end{tabular}%
  \caption{\textbf{Object shapes.} Distinct shape classes used in our “shapes and motions” ablation experiment. For more examples, see Fig.~\protect\ref{fig:gallery-1} and Fig.~\protect\ref{fig:gallery-2}.}%
  \label{fig:data-variation}%
\end{figure}

We designed a series of increasingly complex datasets, starting off with axis-aligned rectangular boxes and motions restricted to translation only, and going all the way to complex thin and non-convex objects with non-rigid motions. 
Examples for the tested scenarios are shown in Figure \ref{fig:data-variation}. 
During training we applied the same color and geometry augmentations used for optical flow training in \citet{flownet}; however, geometry augmentations were restricted to respect the class of motions of the training dataset; for instance, we applied no rotation augmentation if there was no rotation motion in the dataset.

\setlength{\tabcolsep}{2pt}%
\renewcommand{\arraystretch}{1}%

The results are shown in Table \ref{tab:data-variation-types}.
Even the simplest dataset consisting of translated axis-aligned rectangles leads to reasonable performance on the benchmark datasets.
As expected, the general trend is that more complexity and diversity in the shapes and motion during training improves the performance of the resulting networks. 
This can be best seen for the Sintel dataset, where, except for the addition of holes into objects, each additional level of complexity slightly improves the performance.  
While adding only rotations does not lead to much better scores and even decreases the score on KITTI, adding scaling motions on top yields a large improvement.
Holes in objects seem to be counter-productive, perhaps because objects in benchmark datasets rarely have holes.
Nonrigid deformations on objects and background lead to the overall best result. 

The observations can be explained by looking at the types of object motion present in each test dataset: adding rotation motion yields the biggest improvement on FlyingChairs which is also the testset with the strongest object rotation. The dominant motion in KITTI is scaling motion induced by the scene moving towards the camera. Thus, training on scaling motion helps in this case. The nonrigid 2D deformations can approximate the flow patterns exhibited by rotating 3D objects; the effect of training with this is noticeable in Sintel, and strongest in KITTI. Our results confirm that training data can be improved if it is possible to reason about the target domain. On the other hand, this is also disappointing because it indicates that there is no single best general-purpose training dataset.

\rowcolors{4}{white}{gray!20}%
\setlength{\tabcolsep}{2pt}%
\renewcommand{\arraystretch}{1}%
\begin{table}%
  \centering%
  \begin{tabular}{l|P{1.25cm}P{1.25cm}P{1.25cm}P{1.25cm}}%
    \ijcvhruletop
    
               & \multicolumn{4}{c}{Test data} \\
    Train data & Plasma  & Clouds  & Flickr  & Sintel \\
    
    \ijcvhrulemid
    
    Plasma   & $3.57$ & $6.10$ & $5.41$ & $5.85$ \\
    Clouds   & $3.74$ & $3.70$ & $5.05$ & $5.08$ \\
    Flickr   & $3.88$ & $4.13$ & $4.07$ & $\mathbf{4.57}$ \\
    
    \ijcvhrulebot
  \end{tabular}%
  \caption{\textbf{Object and background texture.} We trained FlowNet with three texture types illustrated in Table~\protect\ref{tab:textures} and tested on these datasets and on Sintel. Flickr textures (\ie real photographs) yield the best performance on Sintel. }%
  \label{tab:ablation-textures}%
\end{table}%
\rowcolors{2}{white}{white}%

\setlength{\tabcolsep}{2pt}%
\renewcommand{\arraystretch}{1}%
\begin{table*}%
  \centering%
  \begin{tabular}{c|cc|cc}%
    \ijcvhruletop
    
    Texture type & 
    \multicolumn{2}{c|}{Texture samples} &
    \multicolumn{2}{c}{Data samples} \\
    
    \ijcvhrulemid
    
    Plasma &
    \raisebox{-.5\height}{\includegraphics[clip,trim={0 64px 0 64px},width=.21\linewidth]{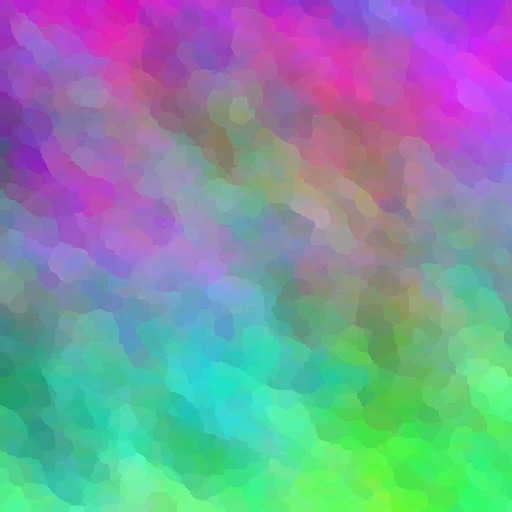}} &
    \raisebox{-.5\height}{\includegraphics[clip,trim={0 64px 0 64px},width=.21\linewidth]{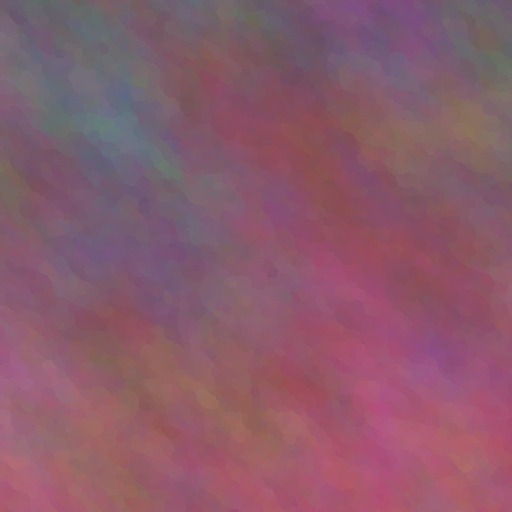}} &
    \raisebox{-.5\height}{\includegraphics[width=.21\linewidth]{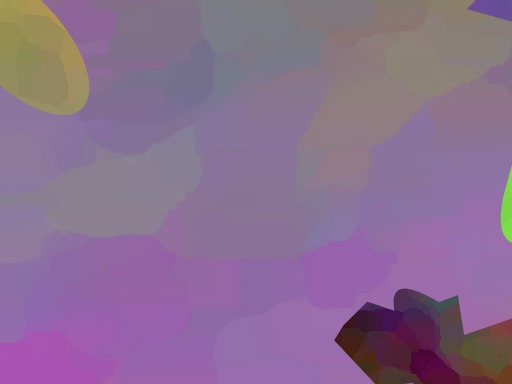}} &
    \raisebox{-.5\height}{\includegraphics[width=.21\linewidth]{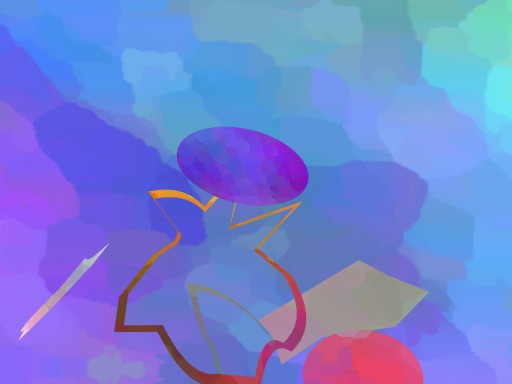}} \\
    
    &&&&\\
    
    Clouds &
    \raisebox{-.5\height}{\includegraphics[width=.21\linewidth]{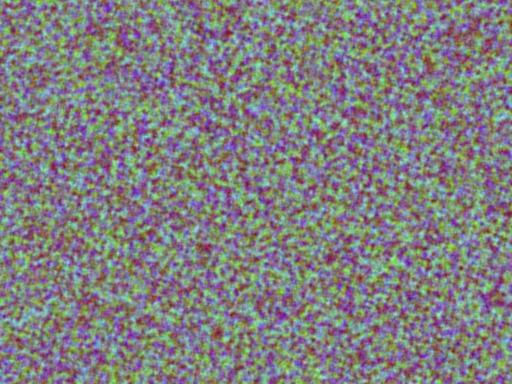}} &
    \raisebox{-.5\height}{\includegraphics[width=.21\linewidth]{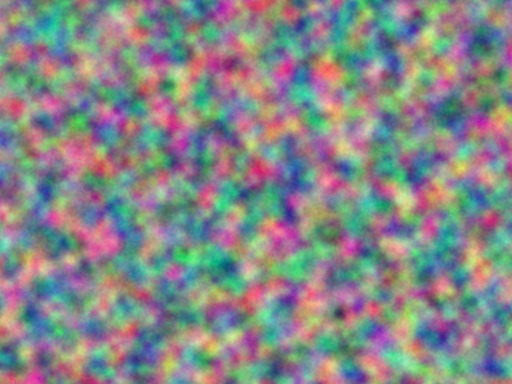}} &
    \raisebox{-.5\height}{\includegraphics[width=.21\linewidth]{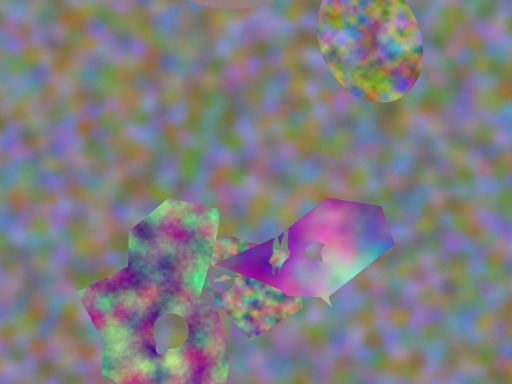}} &
    \raisebox{-.5\height}{\includegraphics[width=.21\linewidth]{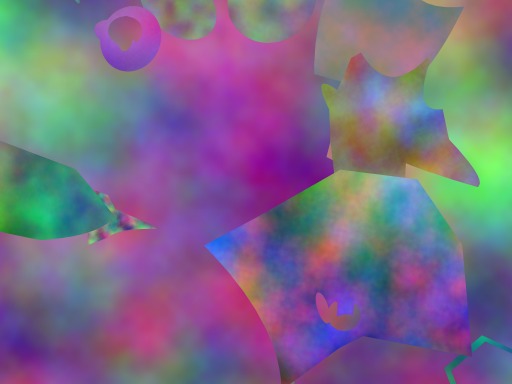}} \\
    
    &&&&\\
    
    Flickr &
    \raisebox{-.5\height}{\includegraphics[width=.21\linewidth]{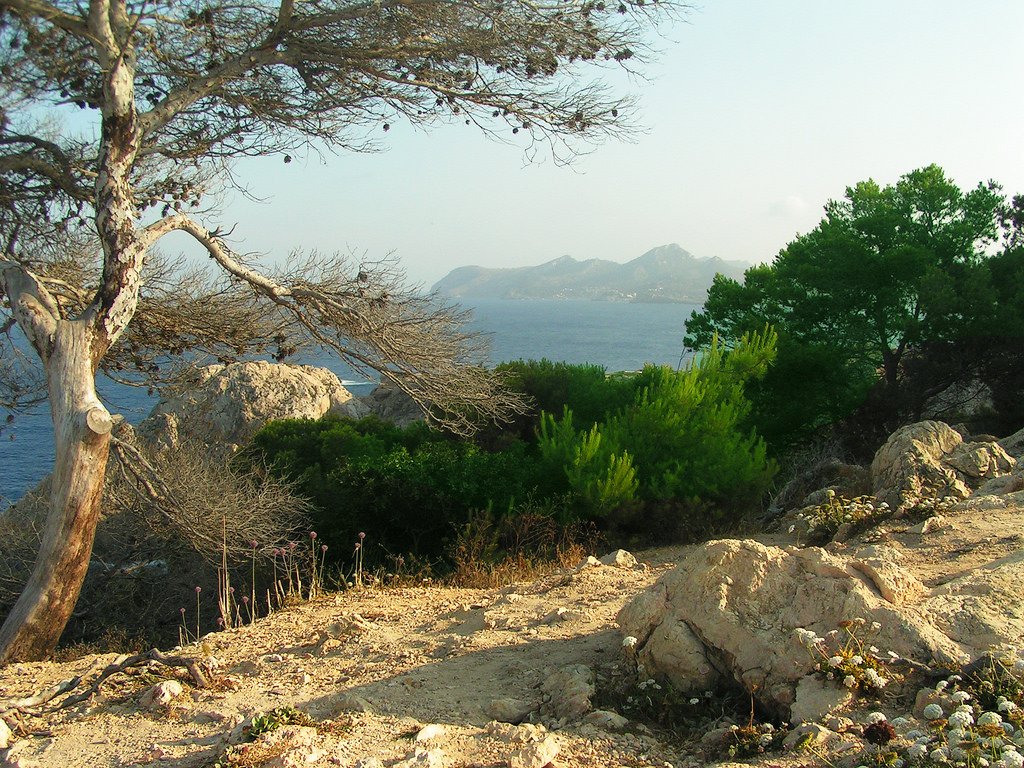}} &
    \raisebox{-.5\height}{\includegraphics[width=.21\linewidth]{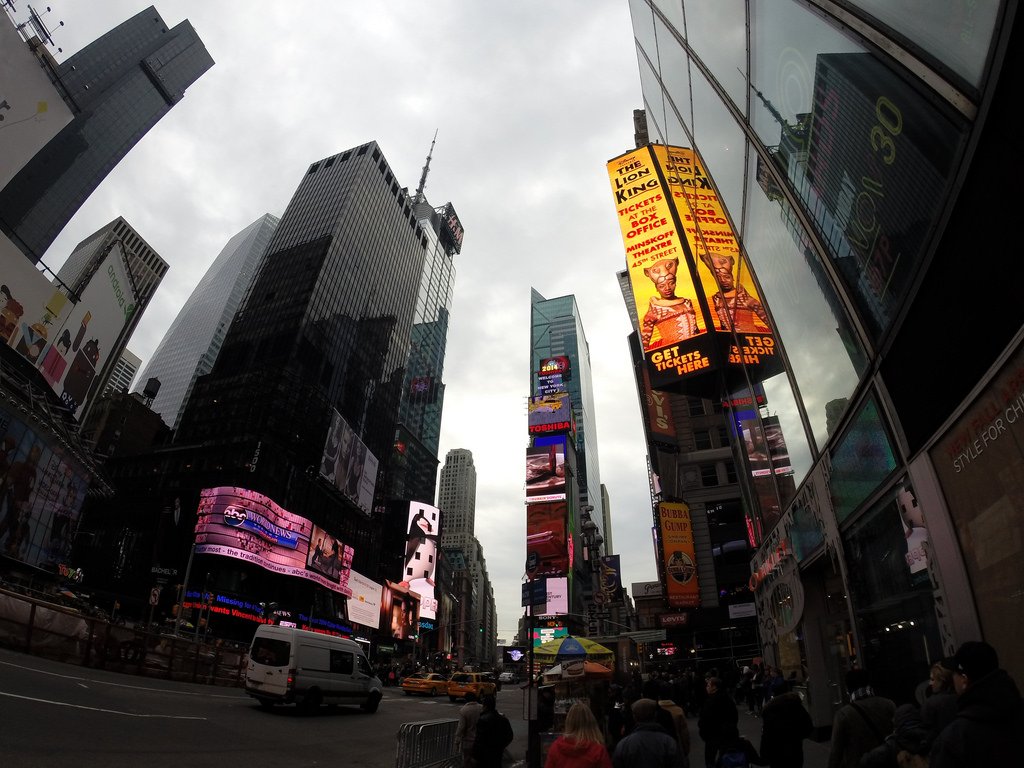}} &
    \raisebox{-.5\height}{\includegraphics[width=.21\linewidth]{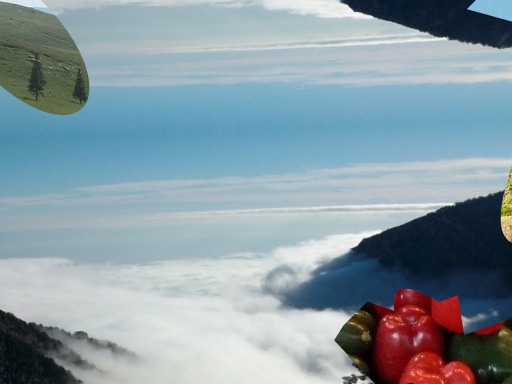}} &
    \raisebox{-.5\height}{\includegraphics[width=.21\linewidth]{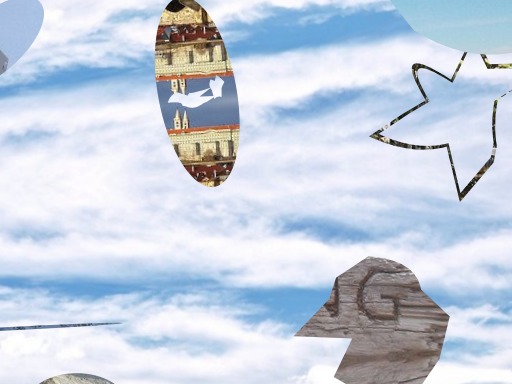}} \\
    
    \ijcvhrulebot
  \end{tabular}%
  \caption{\textbf{Textures.} Example texture images and training samples for the textures ablation experiment. Plasma and Clouds are two types of procedurally generated random textures, while Flickr corresponds to using random natural images from Flickr as textures. More examples in Fig.~\protect\ref{fig:gallery-2} and Fig.~\protect\ref{fig:gallery-3}.}
  \label{tab:textures}%
\end{table*}%

\subsubsection{Textures}
\label{sec:textures}%

In the experiments reported in the previous section we used the same real-world photographs obtained from Flickr both as the background and as object textures.
These provide textures with realistic image statistics, but lack natural semantic context.
But how much does the choice of textures matter for the performance of the trained networks?

In this section we compare three texture variants, illustrated in Table~\ref{tab:textures}.
In addition to Flickr images, we created two additional texture sets: \emph{Plasma} and \emph{Clouds}. 
“Plasma”-type textures\footnote{made by ImageMagick’s random “plasma” generator} are segmented into cells whose sizes range from few pixels to a quarter of the image width.
The cells have clear boundaries; each cell has a single flat color, and cell colors gradually change across the image.
The “clouds”-type images were made by composing color noise on multiple scales.
The resulting textures exhibit structure on all frequencies, but contain almost no constant-color regions or sharp outlines.
These images appear confusing and self-similar to the human eye.

We trained a network from scratch on each texture set and tested it on each other’s testing subsets (a fixed set of $1000$ samples each). 
Table~\ref{tab:ablation-textures} shows that the Flickr textures generalize best to new datasets. In particular, they yield the best results on the Sintel dataset.
This shows how important the diversity of textures during training is for the final network performance, and explains the poor performance when training a network on the Monkaa dataset, which comprises complex objects and motion patterns, but highly repetitive and monotonous textures.

\subsubsection{Displacement statistics}%
\label{sec:histograms}

\begin{figure*}%
  \begin{center}%
    \newcommand*{\HiXScale}{1/25.0}%
    \newcommand*{\HiYScale}{0.5}%
    \newcommand*{\HiYOffset}{3.5}%
      
    \begin{tikzpicture}[%
        xscale=15,%
        yscale=3,%
        plotline/.style={very thick,line cap=round}%
      ]%
      %% Draw grid
      \foreach \y in {1.5,2,...,3.5} {
        \draw[grey,thick,dotted] (0,\y)-- ++(1,0);
      }
      \foreach \x in {0,.1,...,1.1} {
        \draw[grey,thick,dotted] (\x,1.5)-- ++(0,2);
      }
      
      %% Frame
      \draw (0,1.5) -- ++(1,0) -- ++(0,2) -- ++(-1,0) -- cycle;
      %% X base ticks
      \foreach \x in {0.,.1,...,1} {%
        \draw (\x,1.5) -- ++(0, .1);%
        \draw (\x,3.5) -- ++(0,-.1);%
      }
      %% Y base ticks
      \foreach \y in {1.5,2.5,...,3.0} {%
        \draw (0,\y) -- ++( .01,0);%
        \draw (1,\y) -- ++(-.01,0);%
      }
      %% Y logarithm ticks
      \foreach \logbase in {1.5,2.5,...,3.0} {%
        \foreach \log in {.3,.48,.6,.7,.78,.85,.9,.95} {%
          \draw (0,{\logbase+\log}) -- ++( .006,0);%
          \draw (1,{\logbase+\log}) -- ++(-.006,0);%
        }%
      }%
      
      %% Axes labels
      \node at (0.0,1.4) {\small$0$};
      \node at (0.2,1.4) {\small$5$};
      \node at (0.4,1.4) {\small$10$};
      \node at (0.6,1.4) {\small$15$};
      \node at (0.8,1.4) {\small$20$};
      \node at (1.0,1.4) {\small$25$};
      \node at (0,3.5) [anchor=east] {\small$10^{0}$};
      \node at (0,3.0) [anchor=east] {\small$10^{-1}$};
      \node at (0,2.5) [anchor=east] {\small$10^{-2}$};
      \node at (0,2.0) [anchor=east] {\small$10^{-3}$};
      \node at (0,1.5) [anchor=east] {\small$10^{-4}$};
      
      \node at (-.075,2.5) [rotate=90] {\small Fraction of pixels within dataset};
      \node at (0.5,1.25) {\small Displacement magnitude (pixels)};
      
      %% Legend
      \draw[fill=white] (0.75,2.6) rectangle ++(.25,0.9);
      \foreach \x/\y/\color/\text in {%
        0.77/3.35/grey/FlyingThings3D,%
        0.77/3.20/black/3x Sintel-like,%
        0.77/3.05/red/2x Sintel-like,%
        0.77/2.90/orange/FlyingChairs,%
        0.77/2.75/blue/Sintel%
      } {%
        \draw[draw=\color,ultra thick] (\x,\y) -- ++(.05,0);
        \node at ({\x+.05},\y) [anchor=west] {\scriptsize\text};
      }
      
      %% From here on, clip everything to grid area
      \clip (0,1.5) rectangle (1.0,3.5);%
      
      %% Plot functions
      
      %% SINTEL
      \foreach \ax/\ay/\bx/\by in {0.00/-1.0344/0.07/-1.0344, 0.07/-1.0344/0.22/-1.3125, 0.22/-1.3125/0.38/-1.6612, 0.38/-1.6612/0.53/-1.6883, 0.53/-1.6883/0.67/-1.7352, 0.67/-1.7352/0.82/-1.5598, 0.82/-1.5598/0.97/-1.6063, 0.97/-1.6063/1.12/-1.5264, 1.12/-1.5264/1.27/-1.4774, 1.27/-1.4774/1.43/-1.5277, 1.43/-1.5277/1.57/-1.6010, 1.57/-1.6010/1.72/-1.6641, 1.72/-1.6641/1.88/-1.7105, 1.88/-1.7105/2.02/-1.7782, 2.02/-1.7782/2.18/-1.7973, 2.18/-1.7973/2.33/-1.8106, 2.33/-1.8106/2.48/-1.8286, 2.48/-1.8286/2.62/-1.8519, 2.62/-1.8519/2.78/-1.8824, 2.78/-1.8824/2.93/-1.8997, 2.93/-1.8997/3.08/-1.9131, 3.08/-1.9131/3.23/-1.9489, 3.23/-1.9489/3.38/-1.9654, 3.38/-1.9654/3.53/-2.0129, 3.53/-2.0129/3.68/-2.0361, 3.68/-2.0361/3.83/-2.0576, 3.83/-2.0576/3.98/-2.0762, 3.98/-2.0762/4.12/-2.0837, 4.12/-2.0837/4.28/-2.0938, 4.28/-2.0938/4.42/-2.1027, 4.42/-2.1027/4.58/-2.1176, 4.58/-2.1176/4.73/-2.1485, 4.73/-2.1485/4.88/-2.1678, 4.88/-2.1678/5.03/-2.1662, 5.03/-2.1662/5.17/-2.1816, 5.17/-2.1816/5.33/-2.1830, 5.33/-2.1830/5.48/-2.2068, 5.48/-2.2068/5.62/-2.2428, 5.62/-2.2428/5.78/-2.2501, 5.78/-2.2501/5.92/-2.2195, 5.92/-2.2195/6.08/-2.2437, 6.08/-2.2437/6.23/-2.2913, 6.23/-2.2913/6.38/-2.2998, 6.38/-2.2998/6.53/-2.3134, 6.53/-2.3134/6.67/-2.3798, 6.67/-2.3798/6.83/-2.3483, 6.83/-2.3483/6.98/-2.3661, 6.98/-2.3661/7.12/-2.4130, 7.12/-2.4130/7.28/-2.3376, 7.28/-2.3376/7.42/-2.3650, 7.42/-2.3650/7.58/-2.3738, 7.58/-2.3738/7.73/-2.3590, 7.73/-2.3590/7.88/-2.3749, 7.88/-2.3749/8.03/-2.3594, 8.03/-2.3594/8.17/-2.3832, 8.17/-2.3832/8.32/-2.4072, 8.32/-2.4072/8.47/-2.4097, 8.47/-2.4097/8.62/-2.3458, 8.62/-2.3458/8.77/-2.3914, 8.77/-2.3914/8.92/-2.3874, 8.92/-2.3874/9.07/-2.3534, 9.07/-2.3534/9.22/-2.3212, 9.22/-2.3212/9.38/-2.3474, 9.38/-2.3474/9.52/-2.3342, 9.52/-2.3342/9.67/-2.2946, 9.67/-2.2946/9.82/-2.2719, 9.82/-2.2719/9.97/-2.3112, 9.97/-2.3112/10.12/-2.3360, 10.12/-2.3360/10.27/-2.3141, 10.27/-2.3141/10.42/-2.3254, 10.42/-2.3254/10.57/-2.3203, 10.57/-2.3203/10.72/-2.3644, 10.72/-2.3644/10.88/-2.3846, 10.88/-2.3846/11.02/-2.3916, 11.02/-2.3916/11.17/-2.4391, 11.17/-2.4391/11.32/-2.4451, 11.32/-2.4451/11.47/-2.4746, 11.47/-2.4746/11.62/-2.4140, 11.62/-2.4140/11.77/-2.4625, 11.77/-2.4625/11.92/-2.5022, 11.92/-2.5022/12.07/-2.5193, 12.07/-2.5193/12.22/-2.5310, 12.22/-2.5310/12.38/-2.5225, 12.38/-2.5225/12.52/-2.5764, 12.52/-2.5764/12.67/-2.5557, 12.67/-2.5557/12.82/-2.6097, 12.82/-2.6097/12.97/-2.5588, 12.97/-2.5588/13.12/-2.6091, 13.12/-2.6091/13.27/-2.6477, 13.27/-2.6477/13.42/-2.6690, 13.42/-2.6690/13.57/-2.6767, 13.57/-2.6767/13.72/-2.6174, 13.72/-2.6174/13.88/-2.6716, 13.88/-2.6716/14.02/-2.6870, 14.02/-2.6870/14.17/-2.7200, 14.17/-2.7200/14.32/-2.7284, 14.32/-2.7284/14.47/-2.6599, 14.47/-2.6599/14.62/-2.6928, 14.62/-2.6928/14.77/-2.7127, 14.77/-2.7127/14.92/-2.7396, 14.92/-2.7396/15.07/-2.7559, 15.07/-2.7559/15.22/-2.6692, 15.22/-2.6692/15.38/-2.7230, 15.38/-2.7230/15.52/-2.7080, 15.52/-2.7080/15.67/-2.7261, 15.67/-2.7261/15.82/-2.7412, 15.82/-2.7412/15.97/-2.7415, 15.97/-2.7415/16.12/-2.7877, 16.12/-2.7877/16.27/-2.7867, 16.27/-2.7867/16.43/-2.8015, 16.43/-2.8015/16.57/-2.8042, 16.57/-2.8042/16.72/-2.7654, 16.72/-2.7654/16.88/-2.8073, 16.88/-2.8073/17.02/-2.8569, 17.02/-2.8569/17.18/-2.8462, 17.18/-2.8462/17.32/-2.8448, 17.32/-2.8448/17.47/-2.7984, 17.47/-2.7984/17.62/-2.8264, 17.62/-2.8264/17.77/-2.8566, 17.77/-2.8566/17.93/-2.8760, 17.93/-2.8760/18.07/-2.8749, 18.07/-2.8749/18.22/-2.9070, 18.22/-2.9070/18.38/-2.9187, 18.38/-2.9187/18.52/-2.9106, 18.52/-2.9106/18.68/-2.8806, 18.68/-2.8806/18.82/-2.8922, 18.82/-2.8922/18.97/-2.9254, 18.97/-2.9254/19.12/-2.9247, 19.12/-2.9247/19.27/-2.9258, 19.27/-2.9258/19.43/-2.8548, 19.43/-2.8548/19.57/-2.9182, 19.57/-2.9182/19.72/-2.9332, 19.72/-2.9332/19.88/-2.9252, 19.88/-2.9252/20.02/-2.9426, 20.02/-2.9426/20.18/-2.9470, 20.18/-2.9470/20.32/-2.9421, 20.32/-2.9421/20.47/-2.9233, 20.47/-2.9233/20.62/-2.9117, 20.62/-2.9117/20.77/-2.9207, 20.77/-2.9207/20.93/-2.9221, 20.93/-2.9221/21.07/-2.8950, 21.07/-2.8950/21.22/-2.9249, 21.22/-2.9249/21.38/-2.9510, 21.38/-2.9510/21.52/-2.9087, 21.52/-2.9087/21.68/-2.8817, 21.68/-2.8817/21.82/-2.9358, 21.82/-2.9358/21.97/-2.9533, 21.97/-2.9533/22.12/-2.9566, 22.12/-2.9566/22.27/-2.9697, 22.27/-2.9697/22.43/-2.9974, 22.43/-2.9974/22.57/-3.0124, 22.57/-3.0124/22.72/-3.0228, 22.72/-3.0228/22.88/-2.9384, 22.88/-2.9384/23.02/-2.9919, 23.02/-2.9919/23.18/-3.0297, 23.18/-3.0297/23.32/-2.9872, 23.32/-2.9872/23.47/-3.0200, 23.47/-3.0200/23.62/-3.0551, 23.62/-3.0551/23.77/-3.0774, 23.77/-3.0774/23.93/-2.9621, 23.93/-2.9621/24.07/-3.0608, 24.07/-3.0608/24.22/-3.0934, 24.22/-3.0934/24.38/-3.0228, 24.38/-3.0228/24.52/-3.0486, 24.52/-3.0486/24.68/-3.1277, 24.68/-3.1277/24.82/-3.1408, 24.82/-3.1408/24.97/-3.1451, 24.97/-3.1451/25.12/-3.1623, 25.12/-3.1623/25.27/-3.1800, 25.27/-3.1800/25.43/-3.1730, 25.43/-3.1730/25.57/-3.1695, 25.57/-3.1695/25.72/-3.1637, 25.72/-3.1637/25.88/-3.1560, 25.88/-3.1560/26.02/-3.1594, 26.02/-3.1594/26.18/-3.1661, 26.18/-3.1661/26.32/-3.1753, 26.32/-3.1753/26.47/-3.1601, 26.47/-3.1601/26.62/-3.1814, 26.62/-3.1814/26.77/-3.1950, 26.77/-3.1950/26.93/-3.2019, 26.93/-3.2019/27.07/-3.2029, 27.07/-3.2029/27.22/-3.2131, 27.22/-3.2131/27.38/-3.1782, 27.38/-3.1782/27.52/-3.1469, 27.52/-3.1469/27.68/-3.1227, 27.68/-3.1227/27.82/-3.0551, 27.82/-3.0551/27.97/-3.0410, 27.97/-3.0410/28.12/-3.0866, 28.12/-3.0866/28.27/-3.1176, 28.27/-3.1176/28.43/-3.1413, 28.43/-3.1413/28.57/-3.1624, 28.57/-3.1624/28.72/-3.1722, 28.72/-3.1722/28.88/-3.1812, 28.88/-3.1812/29.02/-3.1953, 29.02/-3.1953/29.10/-3.1953}{
        \draw[blue,plotline] ({\ax*\HiXScale},{\ay*\HiYScale+\HiYOffset}) -- ({\bx*\HiXScale},{\by*\HiYScale+\HiYOffset});
      }

      %% FlyingChairs
      \foreach \ax/\ay/\bx/\by in {0.00/-0.8645/0.07/-0.8645, 0.07/-0.8645/0.22/-1.4633, 0.22/-1.4633/0.38/-1.5276, 0.38/-1.5276/0.53/-1.6552, 0.53/-1.6552/0.67/-1.6948, 0.67/-1.6948/0.82/-1.7277, 0.82/-1.7277/0.97/-1.7508, 0.97/-1.7508/1.12/-1.7977, 1.12/-1.7977/1.27/-1.8011, 1.27/-1.8011/1.43/-1.8792, 1.43/-1.8792/1.57/-1.8265, 1.57/-1.8265/1.72/-1.9214, 1.72/-1.9214/1.88/-1.9016, 1.88/-1.9016/2.02/-1.9524, 2.02/-1.9524/2.18/-1.9594, 2.18/-1.9594/2.33/-1.9191, 2.33/-1.9191/2.48/-1.9363, 2.48/-1.9363/2.62/-1.9690, 2.62/-1.9690/2.78/-1.9577, 2.78/-1.9577/2.93/-2.0029, 2.93/-2.0029/3.08/-1.9638, 3.08/-1.9638/3.23/-1.9908, 3.23/-1.9908/3.38/-2.0178, 3.38/-2.0178/3.53/-2.0503, 3.53/-2.0503/3.68/-2.0892, 3.68/-2.0892/3.83/-2.0336, 3.83/-2.0336/3.98/-2.0431, 3.98/-2.0431/4.12/-2.0775, 4.12/-2.0775/4.28/-2.0833, 4.28/-2.0833/4.42/-2.1358, 4.42/-2.1358/4.58/-2.0892, 4.58/-2.0892/4.73/-2.0697, 4.73/-2.0697/4.88/-2.1108, 4.88/-2.1108/5.03/-2.1062, 5.03/-2.1062/5.17/-2.1499, 5.17/-2.1499/5.33/-2.1656, 5.33/-2.1656/5.48/-2.1732, 5.48/-2.1732/5.62/-2.1761, 5.62/-2.1761/5.78/-2.1798, 5.78/-2.1798/5.92/-2.1674, 5.92/-2.1674/6.08/-2.2281, 6.08/-2.2281/6.23/-2.2052, 6.23/-2.2052/6.38/-2.2552, 6.38/-2.2552/6.53/-2.2181, 6.53/-2.2181/6.67/-2.2415, 6.67/-2.2415/6.83/-2.2571, 6.83/-2.2571/6.98/-2.2374, 6.98/-2.2374/7.12/-2.2495, 7.12/-2.2495/7.28/-2.2089, 7.28/-2.2089/7.42/-2.3087, 7.42/-2.3087/7.58/-2.2925, 7.58/-2.2925/7.73/-2.2983, 7.73/-2.2983/7.88/-2.2739, 7.88/-2.2739/8.03/-2.3134, 8.03/-2.3134/8.17/-2.3268, 8.17/-2.3268/8.32/-2.2964, 8.32/-2.2964/8.47/-2.3176, 8.47/-2.3176/8.62/-2.3144, 8.62/-2.3144/8.77/-2.3432, 8.77/-2.3432/8.92/-2.3063, 8.92/-2.3063/9.07/-2.3873, 9.07/-2.3873/9.22/-2.3794, 9.22/-2.3794/9.38/-2.3618, 9.38/-2.3618/9.52/-2.3723, 9.52/-2.3723/9.67/-2.3711, 9.67/-2.3711/9.82/-2.3602, 9.82/-2.3602/9.97/-2.3916, 9.97/-2.3916/10.12/-2.3959, 10.12/-2.3959/10.27/-2.4132, 10.27/-2.4132/10.42/-2.4383, 10.42/-2.4383/10.57/-2.4297, 10.57/-2.4297/10.72/-2.4490, 10.72/-2.4490/10.88/-2.4692, 10.88/-2.4692/11.02/-2.4622, 11.02/-2.4622/11.17/-2.4308, 11.17/-2.4308/11.32/-2.4754, 11.32/-2.4754/11.47/-2.4277, 11.47/-2.4277/11.62/-2.4299, 11.62/-2.4299/11.77/-2.3973, 11.77/-2.3973/11.92/-2.4353, 11.92/-2.4353/12.07/-2.4007, 12.07/-2.4007/12.22/-2.5060, 12.22/-2.5060/12.38/-2.5111, 12.38/-2.5111/12.52/-2.4912, 12.52/-2.4912/12.67/-2.5059, 12.67/-2.5059/12.82/-2.4623, 12.82/-2.4623/12.97/-2.4906, 12.97/-2.4906/13.12/-2.5578, 13.12/-2.5578/13.27/-2.4630, 13.27/-2.4630/13.42/-2.5156, 13.42/-2.5156/13.57/-2.5786, 13.57/-2.5786/13.72/-2.5158, 13.72/-2.5158/13.88/-2.5138, 13.88/-2.5138/14.02/-2.5665, 14.02/-2.5665/14.17/-2.5711, 14.17/-2.5711/14.32/-2.5750, 14.32/-2.5750/14.47/-2.5676, 14.47/-2.5676/14.62/-2.5560, 14.62/-2.5560/14.77/-2.5560, 14.77/-2.5560/14.92/-2.5117, 14.92/-2.5117/15.07/-2.5589, 15.07/-2.5589/15.22/-2.5597, 15.22/-2.5597/15.38/-2.6184, 15.38/-2.6184/15.52/-2.5985, 15.52/-2.5985/15.67/-2.6422, 15.67/-2.6422/15.82/-2.6102, 15.82/-2.6102/15.97/-2.5905, 15.97/-2.5905/16.12/-2.6200, 16.12/-2.6200/16.27/-2.6860, 16.27/-2.6860/16.43/-2.6442, 16.43/-2.6442/16.57/-2.5944, 16.57/-2.5944/16.72/-2.6744, 16.72/-2.6744/16.88/-2.6739, 16.88/-2.6739/17.02/-2.6915, 17.02/-2.6915/17.18/-2.6996, 17.18/-2.6996/17.32/-2.7385, 17.32/-2.7385/17.47/-2.7025, 17.47/-2.7025/17.62/-2.6775, 17.62/-2.6775/17.77/-2.7054, 17.77/-2.7054/17.93/-2.7035, 17.93/-2.7035/18.07/-2.7114, 18.07/-2.7114/18.22/-2.6954, 18.22/-2.6954/18.38/-2.7130, 18.38/-2.7130/18.52/-2.7334, 18.52/-2.7334/18.68/-2.6760, 18.68/-2.6760/18.82/-2.6939, 18.82/-2.6939/18.97/-2.6981, 18.97/-2.6981/19.12/-2.7312, 19.12/-2.7312/19.27/-2.7641, 19.27/-2.7641/19.43/-2.7495, 19.43/-2.7495/19.57/-2.7528, 19.57/-2.7528/19.72/-2.7274, 19.72/-2.7274/19.88/-2.8073, 19.88/-2.8073/20.02/-2.7633, 20.02/-2.7633/20.18/-2.7364, 20.18/-2.7364/20.32/-2.8155, 20.32/-2.8155/20.47/-2.7405, 20.47/-2.7405/20.62/-2.7324, 20.62/-2.7324/20.77/-2.7816, 20.77/-2.7816/20.93/-2.8347, 20.93/-2.8347/21.07/-2.7515, 21.07/-2.7515/21.22/-2.7997, 21.22/-2.7997/21.38/-2.8028, 21.38/-2.8028/21.52/-2.8608, 21.52/-2.8608/21.68/-2.7405, 21.68/-2.7405/21.82/-2.8290, 21.82/-2.8290/21.97/-2.7516, 21.97/-2.7516/22.12/-2.8755, 22.12/-2.8755/22.27/-2.8736, 22.27/-2.8736/22.43/-2.8829, 22.43/-2.8829/22.57/-2.8258, 22.57/-2.8258/22.72/-2.8730, 22.72/-2.8730/22.88/-2.8875, 22.88/-2.8875/23.02/-2.7933, 23.02/-2.7933/23.18/-2.8312, 23.18/-2.8312/23.32/-2.8868, 23.32/-2.8868/23.47/-2.8284, 23.47/-2.8284/23.62/-2.7913, 23.62/-2.7913/23.77/-2.8847, 23.77/-2.8847/23.93/-2.9112, 23.93/-2.9112/24.07/-2.8170, 24.07/-2.8170/24.22/-2.8332, 24.22/-2.8332/24.38/-2.8368, 24.38/-2.8368/24.52/-2.8421, 24.52/-2.8421/24.68/-2.8919, 24.68/-2.8919/24.82/-2.8792, 24.82/-2.8792/24.97/-2.8878, 24.97/-2.8878/25.12/-2.9450, 25.12/-2.9450/25.27/-2.9065, 25.27/-2.9065/25.43/-2.8195, 25.43/-2.8195/25.57/-2.8143, 25.57/-2.8143/25.72/-2.9692, 25.72/-2.9692/25.88/-2.9841, 25.88/-2.9841/26.02/-2.9150, 26.02/-2.9150/26.18/-2.7960, 26.18/-2.7960/26.32/-2.9940, 26.32/-2.9940/26.47/-2.9306, 26.47/-2.9306/26.62/-2.9277, 26.62/-2.9277/26.77/-2.9721, 26.77/-2.9721/26.93/-2.9089, 26.93/-2.9089/27.07/-2.9656, 27.07/-2.9656/27.22/-2.9758, 27.22/-2.9758/27.38/-3.0042, 27.38/-3.0042/27.52/-2.9797, 27.52/-2.9797/27.68/-2.9442, 27.68/-2.9442/27.82/-3.0310, 27.82/-3.0310/27.97/-3.0414, 27.97/-3.0414/28.12/-2.9318, 28.12/-2.9318/28.27/-2.9907, 28.27/-2.9907/28.43/-2.9913, 28.43/-2.9913/28.57/-2.9846, 28.57/-2.9846/28.72/-3.0040, 28.72/-3.0040/28.88/-2.9943, 28.88/-2.9943/29.02/-3.0819, 29.02/-3.0819/29.10/-3.0819}{
        \draw[orange,plotline] ({\ax*\HiXScale},{\ay*\HiYScale+\HiYOffset}) -- ({\bx*\HiXScale},{\by*\HiYScale+\HiYOffset});
      }
      
      %% FlyingThings3D
      \foreach \ax/\ay/\bx/\by in {0.00/-3.5644/0.07/-3.5644, 0.07/-3.5644/0.22/-3.0942, 0.22/-3.0942/0.38/-2.8577, 0.38/-2.8577/0.53/-2.7125, 0.53/-2.7125/0.67/-2.6172, 0.67/-2.6172/0.82/-2.5405, 0.82/-2.5405/0.97/-2.4824, 0.97/-2.4824/1.12/-2.4296, 1.12/-2.4296/1.27/-2.3781, 1.27/-2.3781/1.43/-2.3424, 1.43/-2.3424/1.57/-2.3121, 1.57/-2.3121/1.72/-2.2862, 1.72/-2.2862/1.88/-2.2630, 1.88/-2.2630/2.02/-2.2429, 2.02/-2.2429/2.18/-2.2238, 2.18/-2.2238/2.33/-2.2070, 2.33/-2.2070/2.48/-2.1938, 2.48/-2.1938/2.62/-2.1823, 2.62/-2.1823/2.78/-2.1693, 2.78/-2.1693/2.93/-2.1620, 2.93/-2.1620/3.08/-2.1565, 3.08/-2.1565/3.23/-2.1457, 3.23/-2.1457/3.38/-2.1412, 3.38/-2.1412/3.53/-2.1408, 3.53/-2.1408/3.68/-2.1360, 3.68/-2.1360/3.83/-2.1305, 3.83/-2.1305/3.98/-2.1229, 3.98/-2.1229/4.12/-2.1194, 4.12/-2.1194/4.28/-2.1156, 4.28/-2.1156/4.42/-2.1151, 4.42/-2.1151/4.58/-2.1112, 4.58/-2.1112/4.73/-2.1064, 4.73/-2.1064/4.88/-2.1027, 4.88/-2.1027/5.03/-2.0999, 5.03/-2.0999/5.17/-2.0990, 5.17/-2.0990/5.33/-2.0991, 5.33/-2.0991/5.48/-2.0997, 5.48/-2.0997/5.62/-2.1017, 5.62/-2.1017/5.78/-2.1010, 5.78/-2.1010/5.92/-2.0975, 5.92/-2.0975/6.08/-2.0996, 6.08/-2.0996/6.23/-2.1006, 6.23/-2.1006/6.38/-2.1022, 6.38/-2.1022/6.53/-2.1044, 6.53/-2.1044/6.67/-2.1039, 6.67/-2.1039/6.83/-2.1065, 6.83/-2.1065/6.98/-2.1090, 6.98/-2.1090/7.12/-2.1087, 7.12/-2.1087/7.28/-2.1089, 7.28/-2.1089/7.42/-2.1103, 7.42/-2.1103/7.58/-2.1105, 7.58/-2.1105/7.73/-2.1134, 7.73/-2.1134/7.88/-2.1176, 7.88/-2.1176/8.03/-2.1210, 8.03/-2.1210/8.17/-2.1225, 8.17/-2.1225/8.32/-2.1236, 8.32/-2.1236/8.47/-2.1264, 8.47/-2.1264/8.62/-2.1297, 8.62/-2.1297/8.77/-2.1308, 8.77/-2.1308/8.92/-2.1330, 8.92/-2.1330/9.07/-2.1365, 9.07/-2.1365/9.22/-2.1404, 9.22/-2.1404/9.38/-2.1432, 9.38/-2.1432/9.52/-2.1466, 9.52/-2.1466/9.67/-2.1505, 9.67/-2.1505/9.82/-2.1544, 9.82/-2.1544/9.97/-2.1571, 9.97/-2.1571/10.12/-2.1599, 10.12/-2.1599/10.27/-2.1624, 10.27/-2.1624/10.42/-2.1631, 10.42/-2.1631/10.57/-2.1648, 10.57/-2.1648/10.72/-2.1678, 10.72/-2.1678/10.88/-2.1712, 10.88/-2.1712/11.02/-2.1746, 11.02/-2.1746/11.17/-2.1771, 11.17/-2.1771/11.32/-2.1797, 11.32/-2.1797/11.47/-2.1824, 11.47/-2.1824/11.62/-2.1867, 11.62/-2.1867/11.77/-2.1899, 11.77/-2.1899/11.92/-2.1929, 11.92/-2.1929/12.07/-2.1963, 12.07/-2.1963/12.22/-2.2006, 12.22/-2.2006/12.38/-2.2044, 12.38/-2.2044/12.52/-2.2080, 12.52/-2.2080/12.67/-2.2117, 12.67/-2.2117/12.82/-2.2146, 12.82/-2.2146/12.97/-2.2183, 12.97/-2.2183/13.12/-2.2210, 13.12/-2.2210/13.27/-2.2253, 13.27/-2.2253/13.42/-2.2290, 13.42/-2.2290/13.57/-2.2339, 13.57/-2.2339/13.72/-2.2375, 13.72/-2.2375/13.88/-2.2404, 13.88/-2.2404/14.02/-2.2436, 14.02/-2.2436/14.17/-2.2480, 14.17/-2.2480/14.32/-2.2519, 14.32/-2.2519/14.47/-2.2531, 14.47/-2.2531/14.62/-2.2552, 14.62/-2.2552/14.77/-2.2587, 14.77/-2.2587/14.92/-2.2649, 14.92/-2.2649/15.07/-2.2703, 15.07/-2.2703/15.22/-2.2753, 15.22/-2.2753/15.38/-2.2786, 15.38/-2.2786/15.52/-2.2807, 15.52/-2.2807/15.67/-2.2845, 15.67/-2.2845/15.82/-2.2882, 15.82/-2.2882/15.97/-2.2908, 15.97/-2.2908/16.12/-2.2927, 16.12/-2.2927/16.27/-2.2963, 16.27/-2.2963/16.43/-2.3014, 16.43/-2.3014/16.57/-2.3061, 16.57/-2.3061/16.72/-2.3085, 16.72/-2.3085/16.88/-2.3101, 16.88/-2.3101/17.02/-2.3141, 17.02/-2.3141/17.18/-2.3170, 17.18/-2.3170/17.32/-2.3198, 17.32/-2.3198/17.47/-2.3233, 17.47/-2.3233/17.62/-2.3274, 17.62/-2.3274/17.77/-2.3310, 17.77/-2.3310/17.93/-2.3358, 17.93/-2.3358/18.07/-2.3398, 18.07/-2.3398/18.22/-2.3441, 18.22/-2.3441/18.38/-2.3475, 18.38/-2.3475/18.52/-2.3502, 18.52/-2.3502/18.68/-2.3555, 18.68/-2.3555/18.82/-2.3590, 18.82/-2.3590/18.97/-2.3629, 18.97/-2.3629/19.12/-2.3669, 19.12/-2.3669/19.27/-2.3708, 19.27/-2.3708/19.43/-2.3755, 19.43/-2.3755/19.57/-2.3788, 19.57/-2.3788/19.72/-2.3827, 19.72/-2.3827/19.88/-2.3872, 19.88/-2.3872/20.02/-2.3907, 20.02/-2.3907/20.18/-2.3951, 20.18/-2.3951/20.32/-2.3992, 20.32/-2.3992/20.47/-2.4010, 20.47/-2.4010/20.62/-2.4039, 20.62/-2.4039/20.77/-2.4068, 20.77/-2.4068/20.93/-2.4101, 20.93/-2.4101/21.07/-2.4133, 21.07/-2.4133/21.22/-2.4176, 21.22/-2.4176/21.38/-2.4223, 21.38/-2.4223/21.52/-2.4259, 21.52/-2.4259/21.68/-2.4273, 21.68/-2.4273/21.82/-2.4294, 21.82/-2.4294/21.97/-2.4309, 21.97/-2.4309/22.12/-2.4350, 22.12/-2.4350/22.27/-2.4392, 22.27/-2.4392/22.43/-2.4447, 22.43/-2.4447/22.57/-2.4485, 22.57/-2.4485/22.72/-2.4514, 22.72/-2.4514/22.88/-2.4551, 22.88/-2.4551/23.02/-2.4592, 23.02/-2.4592/23.18/-2.4630, 23.18/-2.4630/23.32/-2.4655, 23.32/-2.4655/23.47/-2.4692, 23.47/-2.4692/23.62/-2.4735, 23.62/-2.4735/23.77/-2.4775, 23.77/-2.4775/23.93/-2.4807, 23.93/-2.4807/24.07/-2.4839, 24.07/-2.4839/24.22/-2.4866, 24.22/-2.4866/24.38/-2.4893, 24.38/-2.4893/24.52/-2.4932, 24.52/-2.4932/24.68/-2.4968, 24.68/-2.4968/24.82/-2.5005, 24.82/-2.5005/24.97/-2.5040, 24.97/-2.5040/25.12/-2.5089, 25.12/-2.5089/25.27/-2.5122, 25.27/-2.5122/25.43/-2.5160, 25.43/-2.5160/25.57/-2.5203, 25.57/-2.5203/25.72/-2.5249, 25.72/-2.5249/25.88/-2.5291, 25.88/-2.5291/26.02/-2.5334, 26.02/-2.5334/26.18/-2.5358, 26.18/-2.5358/26.32/-2.5387, 26.32/-2.5387/26.47/-2.5415, 26.47/-2.5415/26.62/-2.5441, 26.62/-2.5441/26.77/-2.5490, 26.77/-2.5490/26.93/-2.5519, 26.93/-2.5519/27.07/-2.5552, 27.07/-2.5552/27.22/-2.5590, 27.22/-2.5590/27.38/-2.5635, 27.38/-2.5635/27.52/-2.5674, 27.52/-2.5674/27.68/-2.5706, 27.68/-2.5706/27.82/-2.5736, 27.82/-2.5736/27.97/-2.5773, 27.97/-2.5773/28.12/-2.5790, 28.12/-2.5790/28.27/-2.5814, 28.27/-2.5814/28.43/-2.5852, 28.43/-2.5852/28.57/-2.5896, 28.57/-2.5896/28.72/-2.5935, 28.72/-2.5935/28.88/-2.5974, 28.88/-2.5974/29.02/-2.6009, 29.02/-2.6009/29.10/-2.6009}{
        \draw[grey,plotline] ({\ax*\HiXScale},{\ay*\HiYScale+\HiYOffset}) -- ({\bx*\HiXScale},{\by*\HiYScale+\HiYOffset});
      }
      
      %2x
      \foreach \ax/\ay/\bx/\by in {0.00/-1.4397/0.07/-1.4397, 0.07/-1.4397/0.22/-1.7425, 0.22/-1.7425/0.38/-1.7447, 0.38/-1.7447/0.53/-1.7891, 0.53/-1.7891/0.67/-1.8044, 0.67/-1.8044/0.82/-1.8229, 0.82/-1.8229/0.97/-1.8490, 0.97/-1.8490/1.12/-1.8572, 1.12/-1.8572/1.27/-1.8792, 1.27/-1.8792/1.43/-1.8645, 1.43/-1.8645/1.57/-1.9114, 1.57/-1.9114/1.72/-1.9181, 1.72/-1.9181/1.88/-1.9386, 1.88/-1.9386/2.02/-1.9243, 2.02/-1.9243/2.18/-1.9502, 2.18/-1.9502/2.33/-1.9467, 2.33/-1.9467/2.48/-1.9611, 2.48/-1.9611/2.62/-1.9792, 2.62/-1.9792/2.78/-1.9959, 2.78/-1.9959/2.93/-1.9978, 2.93/-1.9978/3.08/-2.0022, 3.08/-2.0022/3.23/-2.0293, 3.23/-2.0293/3.38/-2.0170, 3.38/-2.0170/3.53/-2.0243, 3.53/-2.0243/3.68/-2.0398, 3.68/-2.0398/3.83/-2.0654, 3.83/-2.0654/3.98/-2.0556, 3.98/-2.0556/4.12/-2.0454, 4.12/-2.0454/4.28/-2.0696, 4.28/-2.0696/4.42/-2.0907, 4.42/-2.0907/4.58/-2.0922, 4.58/-2.0922/4.73/-2.0848, 4.73/-2.0848/4.88/-2.0637, 4.88/-2.0637/5.03/-2.0758, 5.03/-2.0758/5.17/-2.0921, 5.17/-2.0921/5.33/-2.1129, 5.33/-2.1129/5.48/-2.1067, 5.48/-2.1067/5.62/-2.1111, 5.62/-2.1111/5.78/-2.1237, 5.78/-2.1237/5.92/-2.1426, 5.92/-2.1426/6.08/-2.1147, 6.08/-2.1147/6.23/-2.1413, 6.23/-2.1413/6.38/-2.1424, 6.38/-2.1424/6.53/-2.1623, 6.53/-2.1623/6.67/-2.1700, 6.67/-2.1700/6.83/-2.1603, 6.83/-2.1603/6.98/-2.1880, 6.98/-2.1880/7.12/-2.1715, 7.12/-2.1715/7.28/-2.1761, 7.28/-2.1761/7.42/-2.2090, 7.42/-2.2090/7.58/-2.1981, 7.58/-2.1981/7.73/-2.2006, 7.73/-2.2006/7.88/-2.2098, 7.88/-2.2098/8.03/-2.2484, 8.03/-2.2484/8.17/-2.2357, 8.17/-2.2357/8.32/-2.2102, 8.32/-2.2102/8.47/-2.2302, 8.47/-2.2302/8.62/-2.2660, 8.62/-2.2660/8.77/-2.2636, 8.77/-2.2636/8.92/-2.2710, 8.92/-2.2710/9.07/-2.2810, 9.07/-2.2810/9.22/-2.2778, 9.22/-2.2778/9.38/-2.2917, 9.38/-2.2917/9.52/-2.2858, 9.52/-2.2858/9.67/-2.2898, 9.67/-2.2898/9.82/-2.2854, 9.82/-2.2854/9.97/-2.2979, 9.97/-2.2979/10.12/-2.2852, 10.12/-2.2852/10.27/-2.3012, 10.27/-2.3012/10.42/-2.3147, 10.42/-2.3147/10.57/-2.2990, 10.57/-2.2990/10.72/-2.3056, 10.72/-2.3056/10.88/-2.3042, 10.88/-2.3042/11.02/-2.3305, 11.02/-2.3305/11.17/-2.3261, 11.17/-2.3261/11.32/-2.3756, 11.32/-2.3756/11.47/-2.3862, 11.47/-2.3862/11.62/-2.3565, 11.62/-2.3565/11.77/-2.3192, 11.77/-2.3192/11.92/-2.3163, 11.92/-2.3163/12.07/-2.3780, 12.07/-2.3780/12.22/-2.3635, 12.22/-2.3635/12.38/-2.3182, 12.38/-2.3182/12.52/-2.3562, 12.52/-2.3562/12.67/-2.3288, 12.67/-2.3288/12.82/-2.3690, 12.82/-2.3690/12.97/-2.3722, 12.97/-2.3722/13.12/-2.3742, 13.12/-2.3742/13.27/-2.4224, 13.27/-2.4224/13.42/-2.4140, 13.42/-2.4140/13.57/-2.3893, 13.57/-2.3893/13.72/-2.4040, 13.72/-2.4040/13.88/-2.3940, 13.88/-2.3940/14.02/-2.4181, 14.02/-2.4181/14.17/-2.4074, 14.17/-2.4074/14.32/-2.4011, 14.32/-2.4011/14.47/-2.4220, 14.47/-2.4220/14.62/-2.4078, 14.62/-2.4078/14.77/-2.4454, 14.77/-2.4454/14.92/-2.4724, 14.92/-2.4724/15.07/-2.4665, 15.07/-2.4665/15.22/-2.4517, 15.22/-2.4517/15.38/-2.4456, 15.38/-2.4456/15.52/-2.4691, 15.52/-2.4691/15.67/-2.4712, 15.67/-2.4712/15.82/-2.4445, 15.82/-2.4445/15.97/-2.4694, 15.97/-2.4694/16.12/-2.4590, 16.12/-2.4590/16.27/-2.4921, 16.27/-2.4921/16.43/-2.4888, 16.43/-2.4888/16.57/-2.4908, 16.57/-2.4908/16.72/-2.4976, 16.72/-2.4976/16.88/-2.4774, 16.88/-2.4774/17.02/-2.5149, 17.02/-2.5149/17.18/-2.4978, 17.18/-2.4978/17.32/-2.5056, 17.32/-2.5056/17.47/-2.4678, 17.47/-2.4678/17.62/-2.5205, 17.62/-2.5205/17.77/-2.5077, 17.77/-2.5077/17.93/-2.5161, 17.93/-2.5161/18.07/-2.4914, 18.07/-2.4914/18.22/-2.5660, 18.22/-2.5660/18.38/-2.5578, 18.38/-2.5578/18.52/-2.5521, 18.52/-2.5521/18.68/-2.5338, 18.68/-2.5338/18.82/-2.5490, 18.82/-2.5490/18.97/-2.5607, 18.97/-2.5607/19.12/-2.5809, 19.12/-2.5809/19.27/-2.5557, 19.27/-2.5557/19.43/-2.5767, 19.43/-2.5767/19.57/-2.5703, 19.57/-2.5703/19.72/-2.5503, 19.72/-2.5503/19.88/-2.5480, 19.88/-2.5480/20.02/-2.5639, 20.02/-2.5639/20.18/-2.5718, 20.18/-2.5718/20.32/-2.5960, 20.32/-2.5960/20.47/-2.6163, 20.47/-2.6163/20.62/-2.5703, 20.62/-2.5703/20.77/-2.6242, 20.77/-2.6242/20.93/-2.5976, 20.93/-2.5976/21.07/-2.6103, 21.07/-2.6103/21.22/-2.5837, 21.22/-2.5837/21.38/-2.5737, 21.38/-2.5737/21.52/-2.6116, 21.52/-2.6116/21.68/-2.6301, 21.68/-2.6301/21.82/-2.5987, 21.82/-2.5987/21.97/-2.5875, 21.97/-2.5875/22.12/-2.6234, 22.12/-2.6234/22.27/-2.6294, 22.27/-2.6294/22.43/-2.6298, 22.43/-2.6298/22.57/-2.6336, 22.57/-2.6336/22.72/-2.6433, 22.72/-2.6433/22.88/-2.5729, 22.88/-2.5729/23.02/-2.5989, 23.02/-2.5989/23.18/-2.6224, 23.18/-2.6224/23.32/-2.6144, 23.32/-2.6144/23.47/-2.5722, 23.47/-2.5722/23.62/-2.5859, 23.62/-2.5859/23.77/-2.6204, 23.77/-2.6204/23.93/-2.6279, 23.93/-2.6279/24.07/-2.6831, 24.07/-2.6831/24.22/-2.6243, 24.22/-2.6243/24.38/-2.6602, 24.38/-2.6602/24.52/-2.6812, 24.52/-2.6812/24.68/-2.6891, 24.68/-2.6891/24.82/-2.6693, 24.82/-2.6693/24.97/-2.6495, 24.97/-2.6495/25.12/-2.6703, 25.12/-2.6703/25.27/-2.6980, 25.27/-2.6980/25.43/-2.6770, 25.43/-2.6770/25.57/-2.6602, 25.57/-2.6602/25.72/-2.6676, 25.72/-2.6676/25.88/-2.6791, 25.88/-2.6791/26.02/-2.7063, 26.02/-2.7063/26.18/-2.6695, 26.18/-2.6695/26.32/-2.6943, 26.32/-2.6943/26.47/-2.6972, 26.47/-2.6972/26.62/-2.7089, 26.62/-2.7089/26.77/-2.6919, 26.77/-2.6919/26.93/-2.6920, 26.93/-2.6920/27.07/-2.6774, 27.07/-2.6774/27.22/-2.7366, 27.22/-2.7366/27.38/-2.7356, 27.38/-2.7356/27.52/-2.7445, 27.52/-2.7445/27.68/-2.7104, 27.68/-2.7104/27.82/-2.7417, 27.82/-2.7417/27.97/-2.6942, 27.97/-2.6942/28.12/-2.7228, 28.12/-2.7228/28.27/-2.7687, 28.27/-2.7687/28.43/-2.7267, 28.43/-2.7267/28.57/-2.7787, 28.57/-2.7787/28.72/-2.7438, 28.72/-2.7438/28.88/-2.7237, 28.88/-2.7237/29.02/-2.7559,  29.02/-2.7559/29.10/-2.7559}{
        \draw[red,plotline] ({\ax*\HiXScale},{\ay*\HiYScale+\HiYOffset}) -- ({\bx*\HiXScale},{\by*\HiYScale+\HiYOffset});
      }
      
      %3x
      \foreach \ax/\ay/\bx/\by in {0.00/-1.6155/0.07/-1.6155, 0.07/-1.6155/0.22/-1.8384, 0.22/-1.8384/0.38/-1.9324, 0.38/-1.9324/0.53/-1.8872, 0.53/-1.8872/0.67/-1.9017, 0.67/-1.9017/0.82/-1.9272, 0.82/-1.9272/0.97/-1.9456, 0.97/-1.9456/1.12/-1.9588, 1.12/-1.9588/1.27/-1.9873, 1.27/-1.9873/1.43/-1.9837, 1.43/-1.9837/1.57/-2.0006, 1.57/-2.0006/1.72/-1.9863, 1.72/-1.9863/1.88/-2.0100, 1.88/-2.0100/2.02/-1.9930, 2.02/-1.9930/2.18/-1.9835, 2.18/-1.9835/2.33/-2.0202, 2.33/-2.0202/2.48/-2.0334, 2.48/-2.0334/2.62/-2.0376, 2.62/-2.0376/2.78/-2.0563, 2.78/-2.0563/2.93/-2.0586, 2.93/-2.0586/3.08/-2.0618, 3.08/-2.0618/3.23/-2.0608, 3.23/-2.0608/3.38/-2.0634, 3.38/-2.0634/3.53/-2.0887, 3.53/-2.0887/3.68/-2.0848, 3.68/-2.0848/3.83/-2.0780, 3.83/-2.0780/3.98/-2.0945, 3.98/-2.0945/4.12/-2.0969, 4.12/-2.0969/4.28/-2.1217, 4.28/-2.1217/4.42/-2.1077, 4.42/-2.1077/4.58/-2.1225, 4.58/-2.1225/4.73/-2.1162, 4.73/-2.1162/4.88/-2.1270, 4.88/-2.1270/5.03/-2.1301, 5.03/-2.1301/5.17/-2.1207, 5.17/-2.1207/5.33/-2.1543, 5.33/-2.1543/5.48/-2.1563, 5.48/-2.1563/5.62/-2.1446, 5.62/-2.1446/5.78/-2.1529, 5.78/-2.1529/5.92/-2.1584, 5.92/-2.1584/6.08/-2.1515, 6.08/-2.1515/6.23/-2.1562, 6.23/-2.1562/6.38/-2.1782, 6.38/-2.1782/6.53/-2.1774, 6.53/-2.1774/6.67/-2.1995, 6.67/-2.1995/6.83/-2.1725, 6.83/-2.1725/6.98/-2.1832, 6.98/-2.1832/7.12/-2.2101, 7.12/-2.2101/7.28/-2.1915, 7.28/-2.1915/7.42/-2.1681, 7.42/-2.1681/7.58/-2.1719, 7.58/-2.1719/7.73/-2.1743, 7.73/-2.1743/7.88/-2.1880, 7.88/-2.1880/8.03/-2.2002, 8.03/-2.2002/8.17/-2.2062, 8.17/-2.2062/8.32/-2.2251, 8.32/-2.2251/8.47/-2.2149, 8.47/-2.2149/8.62/-2.2201, 8.62/-2.2201/8.77/-2.2338, 8.77/-2.2338/8.92/-2.2459, 8.92/-2.2459/9.07/-2.2094, 9.07/-2.2094/9.22/-2.2306, 9.22/-2.2306/9.38/-2.2541, 9.38/-2.2541/9.52/-2.2642, 9.52/-2.2642/9.67/-2.2658, 9.67/-2.2658/9.82/-2.2213, 9.82/-2.2213/9.97/-2.2708, 9.97/-2.2708/10.12/-2.2765, 10.12/-2.2765/10.27/-2.2686, 10.27/-2.2686/10.42/-2.2856, 10.42/-2.2856/10.57/-2.2758, 10.57/-2.2758/10.72/-2.2722, 10.72/-2.2722/10.88/-2.2593, 10.88/-2.2593/11.02/-2.2984, 11.02/-2.2984/11.17/-2.2979, 11.17/-2.2979/11.32/-2.2926, 11.32/-2.2926/11.47/-2.2863, 11.47/-2.2863/11.62/-2.2939, 11.62/-2.2939/11.77/-2.2999, 11.77/-2.2999/11.92/-2.3234, 11.92/-2.3234/12.07/-2.3268, 12.07/-2.3268/12.22/-2.3297, 12.22/-2.3297/12.38/-2.3073, 12.38/-2.3073/12.52/-2.3191, 12.52/-2.3191/12.67/-2.3215, 12.67/-2.3215/12.82/-2.3269, 12.82/-2.3269/12.97/-2.3536, 12.97/-2.3536/13.12/-2.3390, 13.12/-2.3390/13.27/-2.3424, 13.27/-2.3424/13.42/-2.3311, 13.42/-2.3311/13.57/-2.3505, 13.57/-2.3505/13.72/-2.3762, 13.72/-2.3762/13.88/-2.3756, 13.88/-2.3756/14.02/-2.3873, 14.02/-2.3873/14.17/-2.3810, 14.17/-2.3810/14.32/-2.3863, 14.32/-2.3863/14.47/-2.3735, 14.47/-2.3735/14.62/-2.3674, 14.62/-2.3674/14.77/-2.3757, 14.77/-2.3757/14.92/-2.4010, 14.92/-2.4010/15.07/-2.3717, 15.07/-2.3717/15.22/-2.3896, 15.22/-2.3896/15.38/-2.4097, 15.38/-2.4097/15.52/-2.4124, 15.52/-2.4124/15.67/-2.3865, 15.67/-2.3865/15.82/-2.3868, 15.82/-2.3868/15.97/-2.4119, 15.97/-2.4119/16.12/-2.3825, 16.12/-2.3825/16.27/-2.4005, 16.27/-2.4005/16.43/-2.4331, 16.43/-2.4331/16.57/-2.4105, 16.57/-2.4105/16.72/-2.3960, 16.72/-2.3960/16.88/-2.4258, 16.88/-2.4258/17.02/-2.4215, 17.02/-2.4215/17.18/-2.4520, 17.18/-2.4520/17.32/-2.4479, 17.32/-2.4479/17.47/-2.4500, 17.47/-2.4500/17.62/-2.4360, 17.62/-2.4360/17.77/-2.4295, 17.77/-2.4295/17.93/-2.3754, 17.93/-2.3754/18.07/-2.4403, 18.07/-2.4403/18.22/-2.4155, 18.22/-2.4155/18.38/-2.4633, 18.38/-2.4633/18.52/-2.4105, 18.52/-2.4105/18.68/-2.4464, 18.68/-2.4464/18.82/-2.4262, 18.82/-2.4262/18.97/-2.4312, 18.97/-2.4312/19.12/-2.4500, 19.12/-2.4500/19.27/-2.4473, 19.27/-2.4473/19.43/-2.4646, 19.43/-2.4646/19.57/-2.4574, 19.57/-2.4574/19.72/-2.4795, 19.72/-2.4795/19.88/-2.5067, 19.88/-2.5067/20.02/-2.5025, 20.02/-2.5025/20.18/-2.5091, 20.18/-2.5091/20.32/-2.4957, 20.32/-2.4957/20.47/-2.5083, 20.47/-2.5083/20.62/-2.4963, 20.62/-2.4963/20.77/-2.5120, 20.77/-2.5120/20.93/-2.4779, 20.93/-2.4779/21.07/-2.4976, 21.07/-2.4976/21.22/-2.5036, 21.22/-2.5036/21.38/-2.5036, 21.38/-2.5036/21.52/-2.4792, 21.52/-2.4792/21.68/-2.5026, 21.68/-2.5026/21.82/-2.5188, 21.82/-2.5188/21.97/-2.5116, 21.97/-2.5116/22.12/-2.4935, 22.12/-2.4935/22.27/-2.5105, 22.27/-2.5105/22.43/-2.5389, 22.43/-2.5389/22.57/-2.5380, 22.57/-2.5380/22.72/-2.5163, 22.72/-2.5163/22.88/-2.5276, 22.88/-2.5276/23.02/-2.5305, 23.02/-2.5305/23.18/-2.5634, 23.18/-2.5634/23.32/-2.5408, 23.32/-2.5408/23.47/-2.5678, 23.47/-2.5678/23.62/-2.5527, 23.62/-2.5527/23.77/-2.5672, 23.77/-2.5672/23.93/-2.5437, 23.93/-2.5437/24.07/-2.5276, 24.07/-2.5276/24.22/-2.5755, 24.22/-2.5755/24.38/-2.5531, 24.38/-2.5531/24.52/-2.5755, 24.52/-2.5755/24.68/-2.5478, 24.68/-2.5478/24.82/-2.5526, 24.82/-2.5526/24.97/-2.5616, 24.97/-2.5616/25.12/-2.5562, 25.12/-2.5562/25.27/-2.5556, 25.27/-2.5556/25.43/-2.5723, 25.43/-2.5723/25.57/-2.5650, 25.57/-2.5650/25.72/-2.5553, 25.72/-2.5553/25.88/-2.5753, 25.88/-2.5753/26.02/-2.5710, 26.02/-2.5710/26.18/-2.5591, 26.18/-2.5591/26.32/-2.5750, 26.32/-2.5750/26.47/-2.5926, 26.47/-2.5926/26.62/-2.5958, 26.62/-2.5958/26.77/-2.5595, 26.77/-2.5595/26.93/-2.6045, 26.93/-2.6045/27.07/-2.6053, 27.07/-2.6053/27.22/-2.5897, 27.22/-2.5897/27.38/-2.6196, 27.38/-2.6196/27.52/-2.6144, 27.52/-2.6144/27.68/-2.6147, 27.68/-2.6147/27.82/-2.6443, 27.82/-2.6443/27.97/-2.6146, 27.97/-2.6146/28.12/-2.5940, 28.12/-2.5940/28.27/-2.6179, 28.27/-2.6179/28.43/-2.6066, 28.43/-2.6066/28.57/-2.6364, 28.57/-2.6364/28.72/-2.6299, 28.72/-2.6299/28.88/-2.6237, 28.88/-2.6237/29.02/-2.6187,  29.02/-2.6187/29.10/-2.6187}{
        \draw[black,plotline] ({\ax*\HiXScale},{\ay*\HiYScale+\HiYOffset}) -- ({\bx*\HiXScale},{\by*\HiYScale+\HiYOffset});
      }
      
    \end{tikzpicture}%
  \end{center}%
  \caption{\textbf{Displacement statistics.} The statistics of the displacement size of the FlyingChairs dataset matches the statistics of the Sintel dataset fairly closely. The FlyingThings3D dataset was not tuned to match any specific histogram and contains relatively few small displacements. Its greater total count of pixels makes its histogram curve appear smoother. The “2x/3x Sintel-like” datasets intentionally exaggerate Sintel’s distribution (see Section~\protect\ref{sec:histograms}).}%
  \label{fig:histograms}%
\end{figure*}
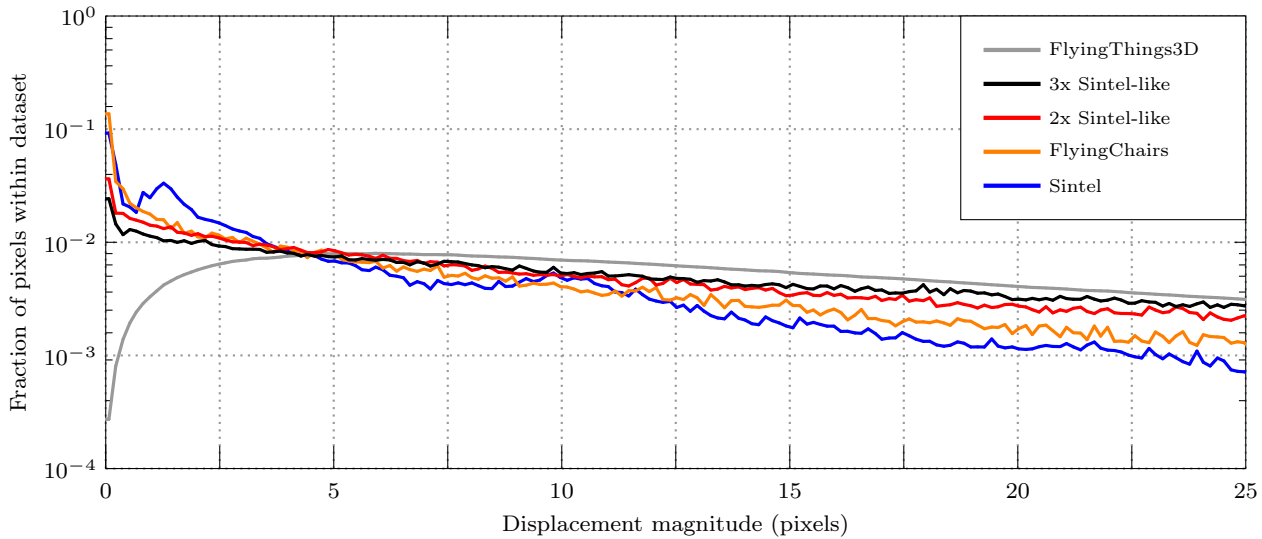%

Classical optical flow estimation methods are sensitive to the displacement magnitude. 
Variational methods \citep{Horn-Schunck-81} have clear advantages in case of small displacements, whereas large displacements require extra treatment and the use of additional combinatorial techniques \citep{BBPW04,ldof}.
When approaching the problem with a deep network, it is not obvious whether similar restrictions apply. 
However, one would expect that the distribution of displacements in the training data plays a crucial role.
This section picks up and expands on experiments done in \citet{flownet2}.

Thus, we evaluated how the performance of the network changes if the displacement distribution of the training set deviates from the distribution of the test set. 
We doubled and tripled the size of the displacements in the FlyingChairs dataset by doubling/tripling all object and background motion distribution parameters (standard deviation for normal distributions, minimum and maximum for uniform distributions).
Fig.~\ref{fig:histograms} shows histogram plots of the datasets’ displacement magnitudes; “2x/3x Sintel-like” are the datasets with exaggerated displacements.

The results are shown in Table~\ref{tab:ablation-histograms}.
Clearly, matching the displacement distribution of the test set is important. 
Using the same displacement statistics as Sintel (denoted as “Sintel-like”) leads to the best results, whereas large deviations from these statistics lead to a large drop in performance. 
In general, such importance of matching the displacement statistics of the test data is disappointing, since one would like to have a network that can deal with all sorts of displacements. This problem can be largely alleviated by the use of learning schedules, as described in Section~\ref{sec:learningschedules}, and by architectures that mix specialized networks, as proposed by \citet{flownet2}.

\rowcolors{4}{white}{gray!20}%
\setlength{\tabcolsep}{2pt}%
\renewcommand{\arraystretch}{1}%
\begin{table}%
  \centering%
  \begin{tabular}{l|P{1.25cm}P{1.25cm}P{1.25cm}P{1.25cm}}%
    \ijcvhruletop
    
               & \multicolumn{4}{c}{Test data} \\[1.5mm]
    Train data & Sintel-like & 2x Sintel-like & 3x Sintel-like & Sintel \\
    
    \ijcvhrulemid
    
      \phantom{2x} Sintel-like   
      & $4.17$ 
      & $11.25$ 
      & $21.84$ 
      & $\mathbf{4.60}$ 
      \\
      2x Sintel-like   
      & $5.04$ 
      & $11.68$ 
      & $20.82$ 
      & $4.99$ 
      \\
      3x Sintel-like   
      & $5.91$ 
      & $12.73$ 
      & $21.70$ 
      & $5.75$ 
      \\
    
    \ijcvhrulebot
  \end{tabular}%
  \caption{\textbf{Flow magnitude distributions.} Results when training on data with different motion statistics. The more the histogram differs from Sintel’s, the worse a trained network performs on Sintel. “2x/3x Sintel-like” are the same as in Fig.~\protect\ref{fig:histograms}.}
  \label{tab:ablation-histograms}%
\end{table}%
\rowcolors{2}{white}{white}%

\rowcolors{4}{white}{gray!20}%
\begin{table}%
  \begin{center}%
    \begin{tabular}{P{2.25cm}|P{1.75cm}|P{1.75cm}|P{1.25cm}}% 
        \ijcvhruletop
        Displacement Range   &  \multicolumn{3}{c}{Partial EPE} \\[1.5mm]
                             & Trained on FlyingChairs & Trained on FlyingThings3D & Difference \\ 
        \ijcvhrulemid
        $\phantom{\infty}0$--$\infty$px$\pz$
        & $\mathbf{4.67}$  
        & $5.76$ 
        & $\phantom{-}1.09$  
        \\ 
        \ijcvhrulemid
        $\pz0$--$10$px 
        & $\mathbf{0.87}$  
        & $1.46$ 
        & $\phantom{-}0.49$ 
        \\ 
        $10$--$40$px     
        & $\mathbf{1.22}$  
        & $1.56$ 
        & $\phantom{-}0.34$ 
        \\ 
        $\pz40$--$160$px    
        & $\mathbf{2.07}$  
        & $2.33$ 
        & $\phantom{-}0.26$
        \\ 
        $\phantom{\infty}160$--$\infty$px$\pz\pz\pz$  
        & $0.51$  
        & $\mathbf{0.42}$ 
        & $-0.09$
        \\ 
        \ijcvhrulebot
    \end{tabular} 
  \end{center}%
   \caption{\textbf{Performance depending on displacement magnitude.} Comparison of error contributions when testing FlowNet on Sintel train clean. Training on FlyingThings3D improves only for displacements larger than 160 pixels but is in all other cases worse. Thus, the displacement statistics from Fig.~\protect\ref{fig:histograms} cannot be the only reason for inferior performance when training on FlyingThings3D.}
  \label{tab:error-fc-fs3d-numbers}
\end{table}%
\rowcolors{2}{white}{white}%

\subsubsection{Lighting}
\label{sec:lightingquality}

In the real world (and in realistically lit synthetic data, such as Sintel), the interaction between light and object surfaces generates reflections, highlights, and shadows.
These phenomena pose a challenge to vision algorithms because they can violate the photoconsistency assumption.
Indeed, highlights and shadows can seem like moving “objects”, but it is generally desired that these effects be ignored.

Potentially, deep networks can learn to distinguish lighting effects from texture and ignore the false motion due to lighting. To this end, they must be provided with training data that shows correct lighting effects. 
We tested whether the FlowNet is able to exploit sophisticated lighting models during training or if optical flow can be estimated just as well with simpler, potentially more efficiently computed lighting.

\setlength{\tabcolsep}{2pt}%
\renewcommand{\arraystretch}{1.5}%
\begin{figure}
  %%
  %% Lighting differences: Example images
  %%
  \begin{tabular}{ccc}%
      \includegraphics[width=.32\linewidth]{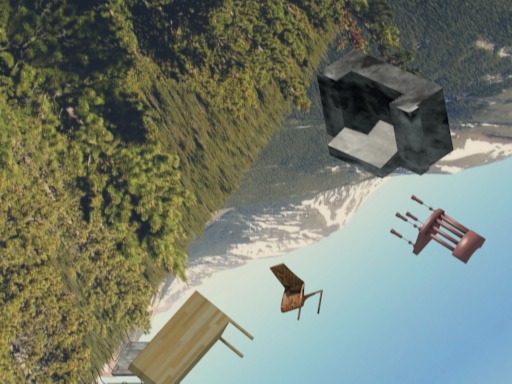}
    & \includegraphics[width=.32\linewidth]{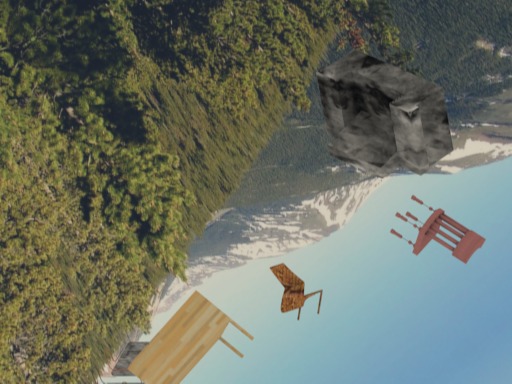}
    & \includegraphics[width=.32\linewidth]{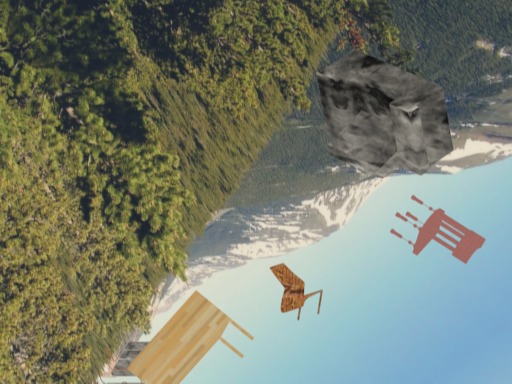}\\
    
      \includegraphics[width=.32\linewidth]{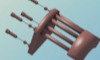}
    & \includegraphics[width=.32\linewidth]{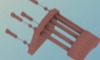}
    & \includegraphics[width=.32\linewidth]{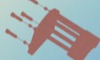}\\
    dynamic & static & shadeless
    
  \end{tabular}%
  \caption{\textbf{Lighting Quality.} Data samples and crops from our “lighting realism” ablation experiment. We used three different render settings, each on the same raw data. The background images are always flat photographs. The crops in the lower row appear blurred only due to upsampling for visualization. See Fig.~\protect\ref{fig:gallery-4} for more examples.}
  \label{fig:lightingquality-examples}
\end{figure}

\rowcolors{3}{white}{gray!20}%
\setlength{\tabcolsep}{2pt}%
\renewcommand{\arraystretch}{1}%
\begin{table}%
  %%
  %% Sintel train final
  %%
  \begin{center}%
    \begin{tabular}{l|P{1.25cm}P{1.25cm}P{1.25cm}P{1.25cm}}%
      \ijcvhruletop
      & \multicolumn{4}{c}{Test data} \\[1.5mm]
      Train data
      & shadeless
      & static
      & dynamic
      & Sintel 
      \\
      \ijcvhrulemid
        shadeless   %% l2
      & $2.71$ %893$
      & $2.77$ %445$
      & $2.93$ %014$
      & $4.41$
      \\
        static  %% l0
      & $2.69$ %92$
      & $2.67$ %603$
      & $2.79$ %736$
      & $\mathbf{4.16}$
      \\
        dynamic   %% l1
      & $2.94$ %794$
      & $2.85$ %293$
      & $2.74$ %166$
      & $4.22$
      \\
        mixture   %%  l0+l1, 1:1, 50% each
      & $2.83$
      & $2.80$
      & $2.86$
      & $4.24$
      \\
      \ijcvhrulebot
    \end{tabular}%
  \end{center}%
\caption{\textbf{Lighting.} Comparison of the lighting models shown in Fig.~\protect\ref{fig:lightingquality-examples}. A network trained on realistic dynamic lighting also expects realistic lighting in the test data. A network trained on static lighting cannot exploit dynamic lighting features, but performs best on Sintel whose realistic lighting setup generates both “static” and “dynamic” effects. A 1:1 static/dynamic mixture does not perform better.}
  \label{fig:lightingquality-numbers}
\end{table}%
\rowcolors{2}{white}{white}%

For this experiment, we used the FlyingChairs dataset re-modeled in 3D, as described in Section~\ref{sec:flyingstuff3d}, in order to apply lighting effects. 
We rendered the three different lighting settings shown in Fig.~\ref{fig:lightingquality-examples}: shadeless, static, and dynamic.
More examples are shown in Fig.~\ref{fig:gallery-4} in the appendix.

The objects in the shadeless variant show no lighting effects at all.
Some object models are textured, but many simply have a uniform flat color.
Estimating optical flow for these objects is hard because the only usable features are given by the shape of the object contour.

Statically lit objects are shaded by environmental lighting which is uniform in all directions and does not consider self-shadowing or other objects.
This corresponds to applying a fixed “shadow texture” to each object.
Known as \emph{ambient occlusion baking}, this approach is often used in computer graphics to make objects look more realistic without having to recompute all lighting whenever an object or light moves.

The dynamic lighting scenario uses raytracing and a light source (lamp), shining into the scene from a randomized direction.
Objects in this case cast realistic shadows and if an object rotates, the shading on its surface changes accordingly.
The chairs also show specular highlights.
This scenario presents the most realistic effects. 
It allows the network to learn about lighting effects, but this also makes the learning task harder: the network must learn to distinguish between different materials and that in case of specular materials, the changing position of a highlight is not supposed to induce optical flow.

We trained a separate network from scratch on each scenario and tested them on each other’s test data, as well as on Sintel’s clean training data.
The results in Table \ref{fig:lightingquality-numbers} show that the network can exploit more complex lighting in the training data to perform better on test data. 
The largest effect of lighting, though, is the larger number of visual features on object surfaces.
The effect of dynamic lighting is much smaller because highlights are sparse in the image. 

The results also show that the network trained with dynamic lighting gets confused by objects that have not been generated with dynamic lighting. 
This can also explain why this network performs marginally worse on Sintel than the network trained with static lighting: if there is no strong directional light source in a Sintel scene, the objects have “static” features but not the hard shadows produced in our “dynamic” setup.
Some surfaces in Sintel and in real scenes are Lambertian, others shiny. The network must learn to distinguish the different surface materials to fully exploit the lighting cue.
A training set that only contains chairs is possibly not sufficient to capture different surface materials.  
Moreover, this makes the learning task more difficult. The latter would also partially explain why the FlyingThings3D dataset, which comes with realistic lighting and many object categories, leads to a network that is inferior to one trained on FlyingChairs. This interpretation is supported by the experiments in Section~\ref{sec:learningschedules}.

\rowcolors{4}{gray!20}{white}%
\setlength{\tabcolsep}{2pt}%
\renewcommand{\arraystretch}{1}%
\begin{table}%
  \begin{center}%
    \begin{tabular}{P{1.25cm}P{1.25cm}P{1.25cm}P{1.25cm}|c}%
      \ijcvhruletop

      \multicolumn{2}{c|}{Color changes}& \multicolumn{2}{c|}{Flow changes} & \\[1.5mm]
      \multicolumn{1}{c|}{on both}   & \multicolumn{1}{c|}{between} & \multicolumn{1}{c|}{on both}  & \multicolumn{1}{c|}{between} & \\
      \multicolumn{1}{c|}{frames} & \multicolumn{1}{c|}{frames}  & \multicolumn{1}{c|}{frames}& \multicolumn{1}{c|}{frames}  & Sintel \\

      \ijcvhrulemid

          &     &     &     & $7.66$ \\  %% no augmentation
      \my &     &     &     & $6.30$ \\  %% same color change to both images
      \my & \my &     &     & $6.12$ \\  %% +different color change to second image
          &     & \my &     & $5.25$ \\  %% same geometry change to both images
          &     & \my & \my & $5.33$ \\  %% +different geometry change to second image
      \my &     & \my &     & $5.11$ \\  %% same color and geometry change to both images
      \my & \my & \my & \my & $\mathbf{4.60}$ \\  %% full augmentation
      
      \ijcvhrulebot
    \end{tabular}%
  \end{center}%
  \caption{\textbf{Data augmentation.} The results show the effect on FlowNet performance when different types of augmentation are applied to training data (here we used “\printasis{+}Thin objects” from Table~\protect\ref{tab:data-variation-types}). The data can be augmented by applying color and/or geometry changes. One can also choose (a) to only apply the same changes to both input images, or (b) to additionally apply a second set of incremental changes to the second image.}
\label{tab:augmentation}
\end{table}%
\rowcolors{2}{white}{white}%

\subsubsection{Data augmentation}
\label{sec:augmentation}

We split the data augmentation into augmentations on color and augmentations on geometry
(these augmentations were used in our previous works \citep{flownet,dispnet,flownet2}, but there we did not analyze the effects of the individual options).
In Table \ref{tab:augmentation} we show an ablation study on the full set of augmentation options: no augmentation, augmentations on either color or geometry, or both.
We further split the experiments by whether both images of a sample pair receive only the same augmentation, or whether a second, incremental transformation is additionally applied to the second image.

The evaluation shows that almost all types of data augmentation are complementary and important to improve the performance of the network. 
The notable exception is additional geometry augmentation between the frames of a sample. The raw dataset seems to already contain sufficiently varied displacements, such that additional augmentation between frames cannot generate helpful new data. Contrary to this, color changes between frames always introduce new data: the unaugmented dataset does not contain this effect, and applying the same color augmentation to both frames does not change this.

\begin{figure}%
  \begin{tikzpicture}[%
    font=\footnotesize,%
    xscale=0.925,yscale=0.4,%
    cpunode/.style={fill=blue,circle,scale=.6},%
    gpunode/.style={fill=orange,circle,scale=.6},%
    ournode/.style={fill=orange,draw,thick,circle,scale=.6},%
    l/.style={fill=white,inner sep=1pt}%
  ]%
    %% Grid
    \draw[dotted] (0,0) grid (8,10);%
    %% Frame
    \draw[thick] (0,0) -- ++(8,0) -- ++(0,10) -- ++(-8,0) -- cycle;%
    %% Axes labels
    \node[rotate=90] at (-.8,5) {EPE on Sintel train clean};%
    \node at (4,-1.7) {\#Training Samples};%
    %% Y base ticks
    \foreach \y in {2,3,...,12} {%
      \node at (0,{\y-2}) [anchor=east] {$\y$};%
      \draw (0,{\y-2}) -- ++(.05,0);%
      \draw (8,{\y-2}) -- ++(-.05,0);%
    }%
    %% X base ticks
    \foreach \x in {0,1,...,8} {%
      %\node at (\x,-.3) {$10^{\x}$};%
      \draw (\x,0) -- ++(0,.2);%
      \draw (\x,10) -- ++(0,-.2);%
    }%
    \node at (1,-.6) {$3.125$k};%
    \node at (2,-.6) {$6.25$k};%
    \node at (3,-.6) {$12.5$k};%
    \node at (4,-.6) {$25$k};%
    \node at (5,-.6) {$50$k};%
    \node at (6,-.6) {$100$k};%
    \node at (7,-.6) {\phantom{*}$\infty$};%

    %% Points
    \node[cpunode] at (1,{4.99 -2}) {};%
    \node[cpunode] at (2,{4.89 -2}) {};%
    \node[cpunode] at (3,{4.73 -2}) {};%
    \node[cpunode] at (4,{4.63 -2}) {};%
    \node[cpunode] at (5,{4.61 -2}) {};%
    \node[cpunode] at (6,{4.62 -2}) {};%
    \node[cpunode] at (7,{4.33 -2}) {};%
    
    \node[gpunode] at (1,{11.0 -2}) {};%
    \node[gpunode] at (2,{9.66 -2}) {};%
    \node[gpunode] at (3,{8.71 -2}) {};%
    \node[gpunode] at (4,{7.58 -2}) {};%
    \node[gpunode] at (5,{7.25 -2}) {};%
    \node[gpunode] at (6,{6.55 -2}) {};%
    \node[gpunode] at (7,{6.05 -2}) {};%
    
    %% Legend
    \draw[thick,fill=white] (4,7) rectangle ++(4,3);%
    \node[cpunode] at (4.25,8) {};%
    \node at (4.35,8) [anchor=west] {with augmentation};%
    \node[gpunode] at (4.25,9) {};%
    \node at (4.35,9) [anchor=west] {without augmentation};%
  \end{tikzpicture}%
  \caption{%
   \textbf{Amount of Training Data.} $\infty$ indicates as many training samples as training iterations (no training sample was ever used twice). This figure complements the augmentation modes experiment in Table~\protect\ref{tab:augmentation} and uses the same training data setting. A FlowNet trained without data augmentation gets much greater relative benefits from more training data, compared to augmented training. However, even with infinite data, networks without augmentation perform far worse.
  }%
  \label{fig:amountoftrainingdata}
\end{figure}
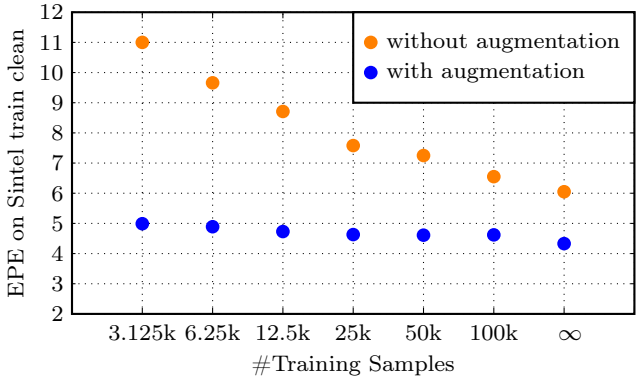

\subsubsection{Amount of training data}
\label{sec:amountoftrainingdata}

It is generally accepted that more training data leads to better results in supervised training.
Data augmentation increases the effective size of the training dataset. However, an augmented existing sample is not the same as a truly new sample: for example, augmentation as we use it cannot create new textures or change the relative positions within a group of objects.

We compared the effect of augmentation to truly changing the size of the dataset. To evaluate this, we trained FlowNets on different amounts of data and either allowed full data augmentation or none (corresponding to the first and last rows in Table~\ref{tab:augmentation}).
The results are given in Fig.~\ref{fig:amountoftrainingdata} and show that while more data is indeed always better,
randomized augmentation yields another huge improvement even for an infinite amount of training samples, \ie as many samples as training iterations (corresponding to $2.4$M samples for $600$k minibatches and batch-size $4$). Therefore, augmentation allows for a $\sim\!100$-fold reduction of the training data and still provides better results.

From this section and the previous Sec.~\ref{sec:augmentation}, we conclude that data augmentation serves two purposes: (a) if it only replicates the effects found in the original data, then augmentation increases the effective size of the training set without changing the domain of the data. This can help if the dataset is just not big enough. On the other hand, (b) augmentation may be able to cover types of data variation which are complementary to that of the original data (depending on the task and data domain), and in this case a network can learn to cope with a wider range of inputs. However, augmentation has inherent limits, and we achieved the best results when using both augmentation and as much raw data as possible.

\subsection{Learning schedules with multiple datasets}
\label{sec:learningschedules}

The results from Table~\ref{tab:introexperiment} in Section~\ref{sec:good_training_dataset} show that training a network on the FlyingThings3D dataset~\citep{flownet2} yields worse results on Sintel than training the same network on FlyingChairs~\citep{flownet}, although the former is more diverse and uses more realistic modeling. The experimental results presented so far do not yet give an explanation for this behavior.
This section complements the dataset schedule experiments from \citet{flownet2}.

Fig.~\ref{fig:histograms} reveals that FlyingThings3D has much fewer small displacements (smaller than 5 pixels).   Moreover, Table~\ref{tab:error-fc-fs3d-numbers} shows that the displacement statistics cannot be the only reason for the inferior performance of FlyingThings3D over FlyingChairs: although displacements between 40 and 160 pixels are well represented in FlyingThings3D, it still performs worse than FlyingChairs in that displacement range.

\begin{figure}%
  \centering%
  \includegraphics[width=\linewidth]{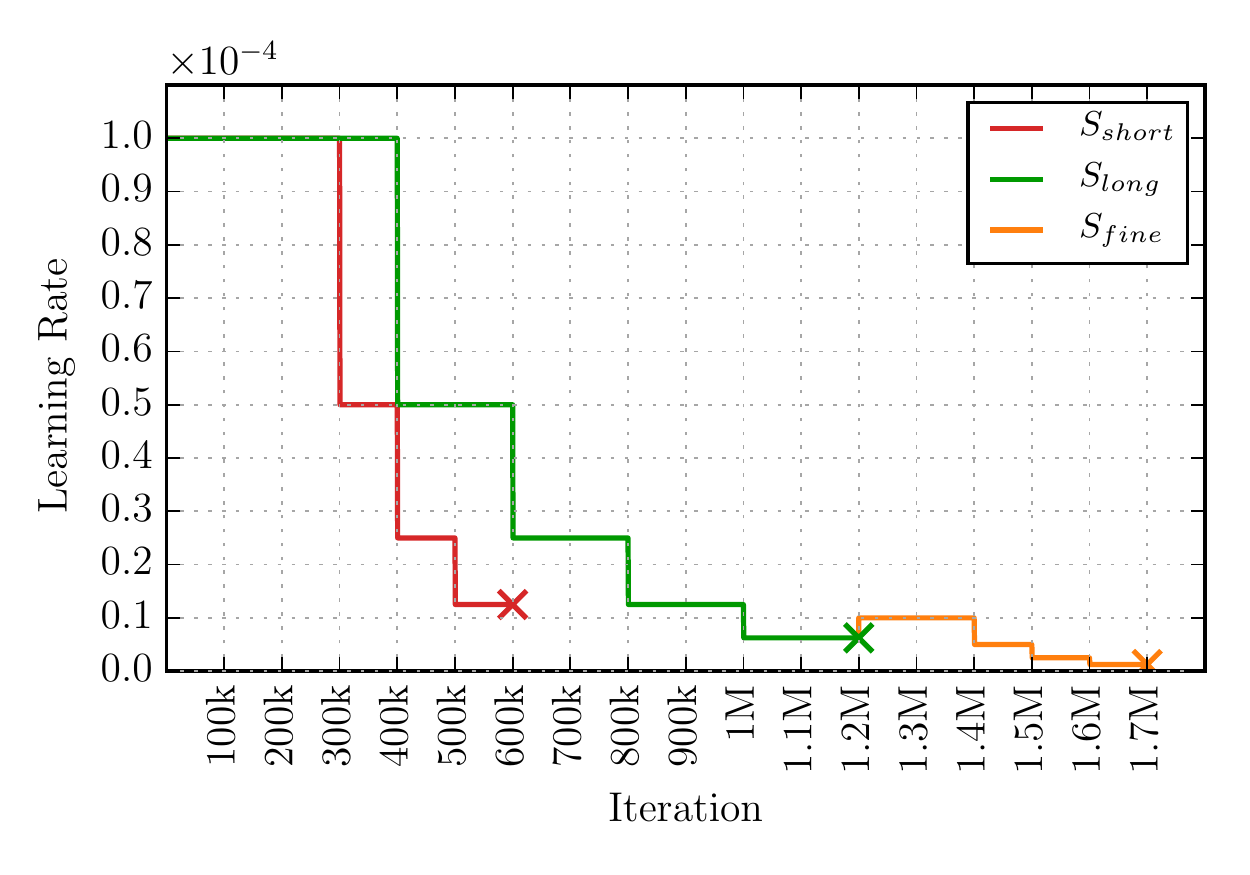}\\[-2mm]
  \caption{\textbf{Learning rate schedules.} In most experiments of this work $S_\text{short}$ is used due to its lower computational cost. Section~\protect\ref{sec:learningschedules} uses $S_\text{long}$ in combination with $S_\text{fine}$, where $S_\text{fine}$ is used for fine-tuning on a second dataset (as done in \citet{flownet2} where these schedules were introduced).}
\label{fig:lr-schedules}%
\end{figure}

To investigate the effects of both datasets further, we trained FlowNets on combinations of both datasets, using the $S_\text{long}$ and $S_\text{fine}$ schedules (see Fig.~\ref{fig:lr-schedules}). The results are given in Table~\ref{tab:learningschedules} and show that, in fact, using a combination of FlyingChairs and FlyingThings3D training data yields the best results. Moreover, this only holds if the datasets are used separately and in the specific order of training on FlyingChairs first and then fine-tuning on FlyingThings3D. This reveals that not only the content of the data matters, but also the point of time at which it is presented to the network during the training phase. This is strongly connected to curriculum learning, which has previously been applied to shape recognition \citep{curriculumlearning,startingsmall} and other (non-vision) tasks. From this observation, we conclude that presenting a too sophisticated dataset too early is disadvantageous, but at a later stage it actually helps. A reason might be that the simpler FlyingChairs dataset helps the network learn the general concept of finding correspondences without developing possibly confusing priors for 3D motion too early.

\rowcolors{2}{gray!20}{white}%
\begin{table}%
  \begin{center}%
    \begin{tabular}{P{4.5cm}|P{1.25cm}|P{1.25cm}}% 
        \ijcvhruletop
        Datasets & $S_\text{long}$ & $S_\text{fine}$ \\ 
        \ijcvhrulemid
        FlyingChairs & $4.22$ & $4.15$ \\ 
        FlyingThings3D & $5.14$ & $4.93$ \\ 
        Mixture & $4.31$ & $4.25$ \\ 
        FlyingChairs $\!\to\!$ FlyingThings3D & $4.22$  & $\mathbf{4.05}$ \\ 
        FlyingThings3D $\!\to\!$ FlyingChairs & $5.14$  & $4.60$ \\ 
        \ijcvhrulebot
    \end{tabular} 
  \end{center}%
  \caption{\textbf{Learning schedules.} Results on Sintel train clean of training FlowNet on mixtures of FlyingChairs and FlyingThings3D with schedules $S_\text{long}$ and $S_\text{fine}$. “Mixture” is a 1:1 mix of both entire datasets. One can observe that only the sequential combination FlyingChairs $\!\to$ FlyingThings3D gives significantly better results.}
  \label{tab:learningschedules}
\end{table}%
\rowcolors{2}{white}{white}%

\setlength{\tabcolsep}{2pt}%
\renewcommand{\arraystretch}{1}%
\begin{figure*}[!h]%
  \begin{tabular}{ccc}%
    \includegraphics[width=.325\linewidth]{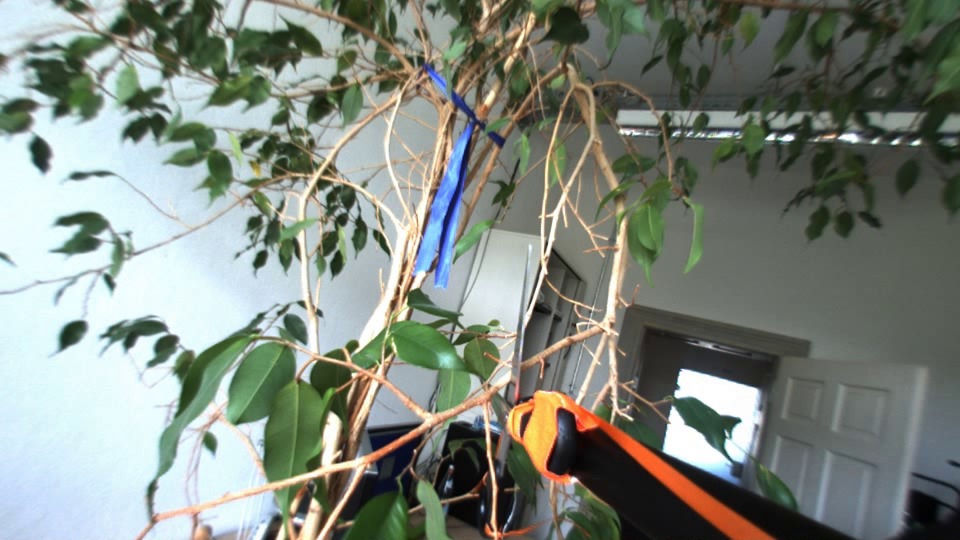} &
    \includegraphics[width=.325\linewidth]{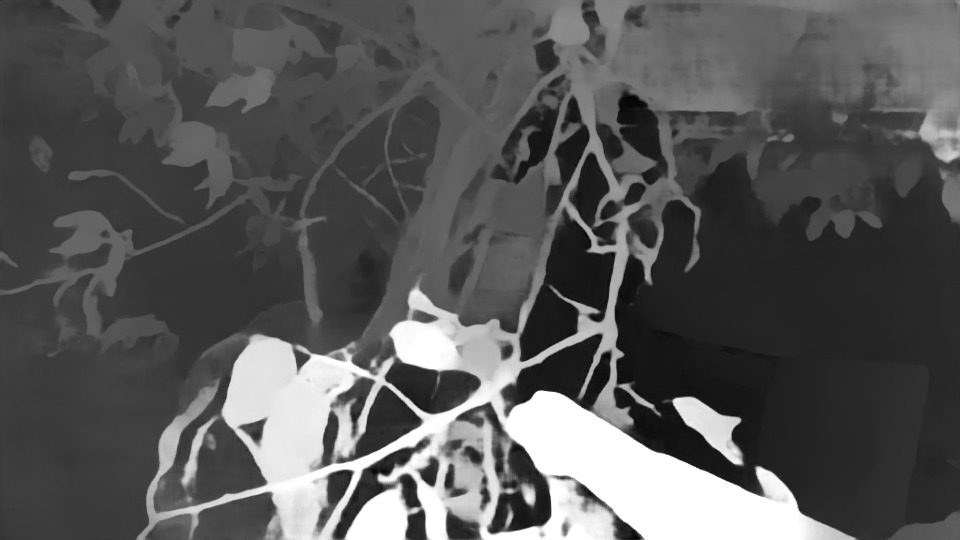} &
    \includegraphics[width=.325\linewidth]{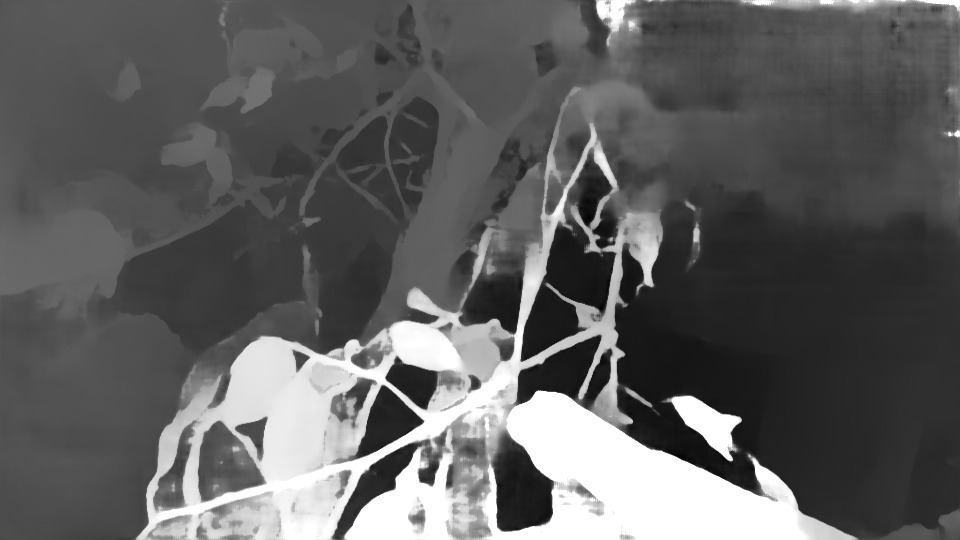} \\
    (a) Left-eye input & (b) DispNet output (trained on (e)) & (c) DispNet output (trained on (f)) \\[2mm]
    
    \includegraphics[width=.325\linewidth]{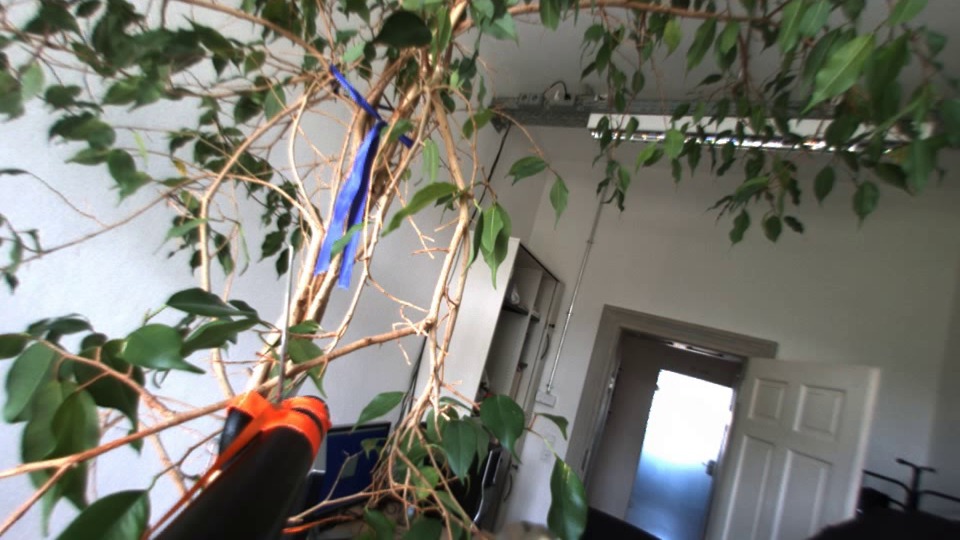} &
    \includegraphics[width=.325\linewidth]{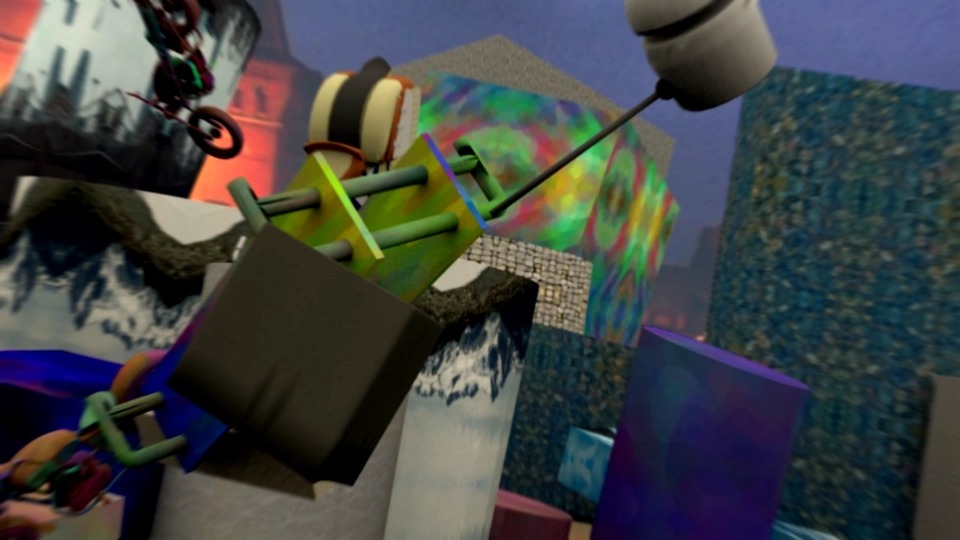} &
    \includegraphics[width=.325\linewidth]{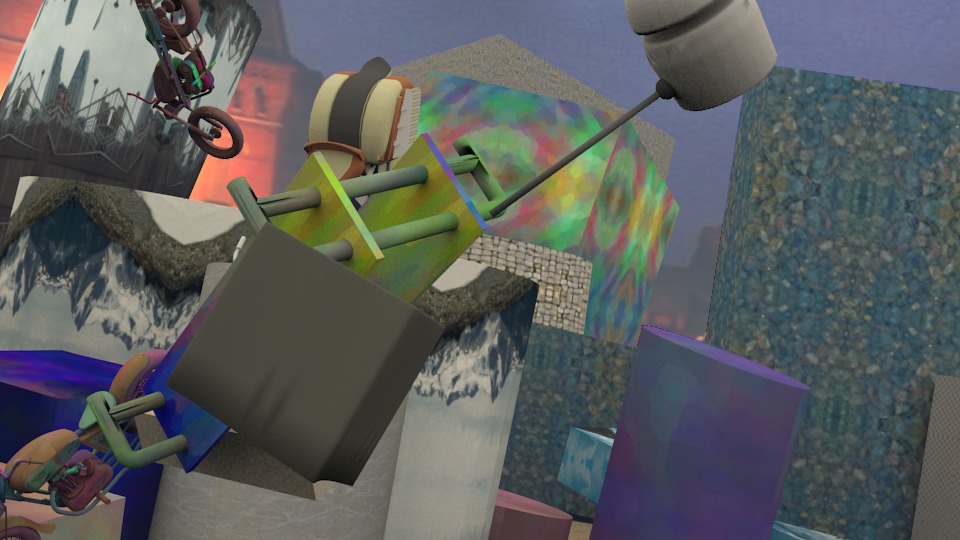} \\
    (d) Right-eye input & (e) Degraded training data & (f) Original training data \\
  \end{tabular}%
      \caption{\textbf{Lens distortion and blur in training data.} A DispNet trained on clean synthetic data (f) does not perform well out of the box on images from a real camera; the input images from a Bumblebee stereo camera (a,d) show significant radial blur after undistortion, as well as general blur. Adding these degradation effects to the training data (e) to make it look more like the images from this camera leads to significantly more detailed disparity estimates (b) than without these effects (c), especially far away from the image center, where the radial blur is the strongest. In this case, the network learns to deal with the faults in the training data and can transfer this knowledge to the real images.}
  \label{fig:datadeg-bumblebee}%
\end{figure*}%

\subsection{Synthesizing the defects of real imaging}
\label{sec:datadegradation}

In this section, we focus on the artifacts that real-world imaging systems add to the data.
The effects of the imaging process on the obtained images have recently been studied by \citet{simulatingcameras}. 
There, a rendering pipeline was designed to artificially apply the same artifacts that the real camera produces to rendered objects.
As a result, augmented-reality objects blended well into the real images; without this adjustment, the rendered objects stood out because they looked unnaturally clean and clear.
The OVVV dataset by \citet{ovvv} uses a comparable approach, simulating analog video effects to make their evaluation data for video surveillance systems more realistic.

We took this idea and applied it to \emph{training} data: we usually train networks on synthetic data because real data with good ground truth is not always available in quantity or even possible; yet the fact that this synthetic data was not recorded by a physical camera already makes it different.
We investigated what happens when we corrupt the training data with the same flaws of real cameras\footnote{Visual effects artists have long been using the knowledge that “perfect” pictures are not perceived as “real” by humans; hence artificial film grain, chromatic aberrations, and lens flare effects are applied in movies and computer games.}.

Applying such data degradation leads to two possible, counteracting effects: 
(a) in artificially degraded data, the relationship between image pixels and ground truth labels is muddied by blur etc. The network already has to cope with the inherent discretization artifacts of raster images in any case, but complicating this further could lead to worse performance.
On the other hand, (b) the network should be able to learn that certain flaws or artifacts convey no meaning and subsequently be able to ignore such effects.
If the network can transfer this knowledge to real-world test data, its performance should improve.

We performed two disparity estimation experiments to test our conjectures: we observed the qualitative effects of degraded synthetic training data on images taken with a commercial stereo camera with low-quality wide-angle lenses; and we quantified whether training data adapted to camera artifacts in KITTI can improve DispNet performance on the KITTI2015 dataset (one of the few real-world datasets with ground truth).
In both cases, the degradation was tuned to simulate the actual lens and camera effects seen in the respective real data.

Note that this section’s view on data is orthogonal to that of the preceding sections, and we consider the experiments disjoint: above, we looked at \emph{features and content} in the data (motion types, texture patterns, object shapes etc.). In this section, we rather look at \emph{pixel-level} data characteristics that are independent from what the images actually depict. 
To perform these experiments, we used the established datasets from our prior works, not the additional experimental ones from this paper.

\subsubsection{Lens distortion and blur}
The notable degradation of images produced by a Bumblebee\footnote{Bumblebee2 BB2-08S2C} stereo camera is due to radial blur (caused by undistortion from a wide-angle lens), some general blur, and over-saturated colors (\ie strong contrast).
We replicated these effects offline for the entire FlyingThings3D training dataset \citep{dispnet}, using filters in the GIMP image processing tool.

To test how the degradation affects performance, we trained two DispNetCorr1D networks from scratch \citep{dispnet}: one on the original dataset and the other on the degraded data.
Fig.~\ref{fig:datadeg-bumblebee} highlights the qualitative differences in the disparity maps estimated by the two networks: training on artificially degraded data produces much finer details, especially in areas towards the image boundaries, where the lens distortion has the largest effect.
The network even makes more sensible guesses for areas which are occluded in the other view: the network seems to have learned that object boundaries can be blurry.
This cannot be inferred from the original training data with crisp contours.

\setlength{\tabcolsep}{2pt}%
\renewcommand{\arraystretch}{1}%
\begin{figure*}[!h]%
  \begin{tabular}{cccc}%
    \multicolumn{2}{c}{\includegraphics[width=.49\linewidth,trim=0 32 0 0,clip]{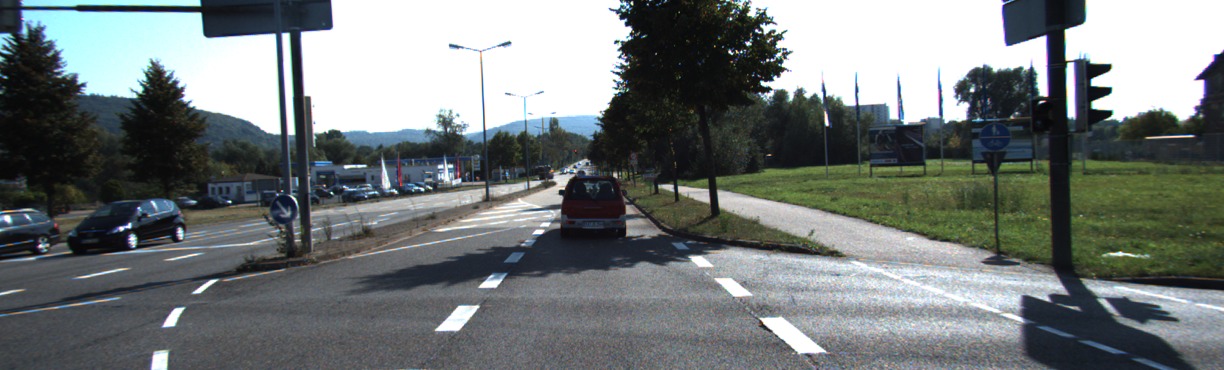}} &
    \includegraphics[width=.24\linewidth]{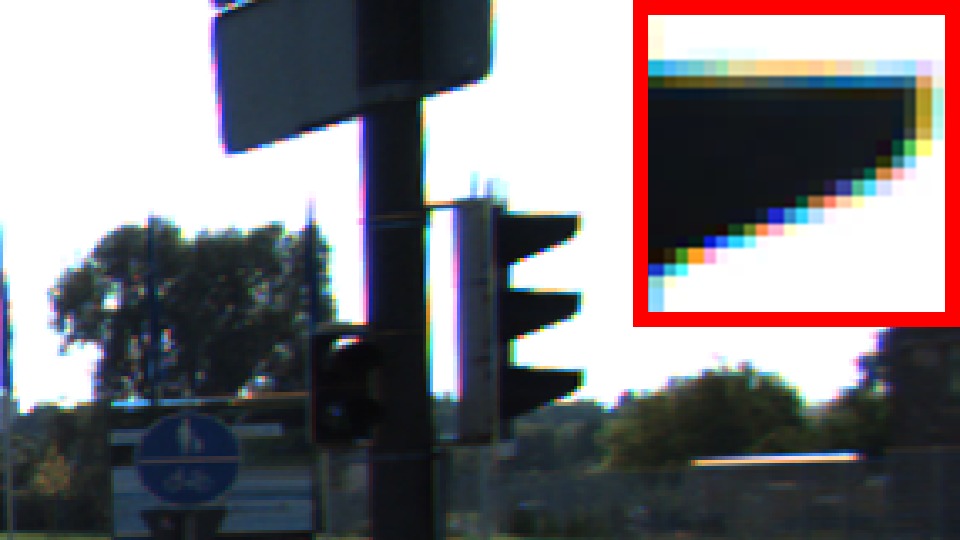} &
    \includegraphics[width=.24\linewidth]{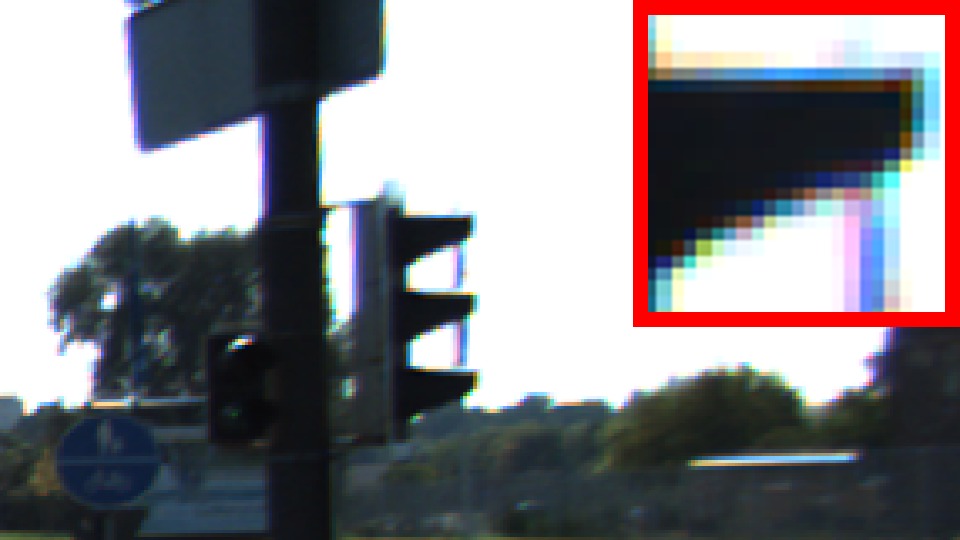} \\
    \multicolumn{2}{c}{KITTI 2015 example, left image} &
    Crop from left image &
    Crop from right image \\[2mm]
    
    \multicolumn{2}{c}{\includegraphics[width=.49\linewidth,trim=0 32 0 0,clip]{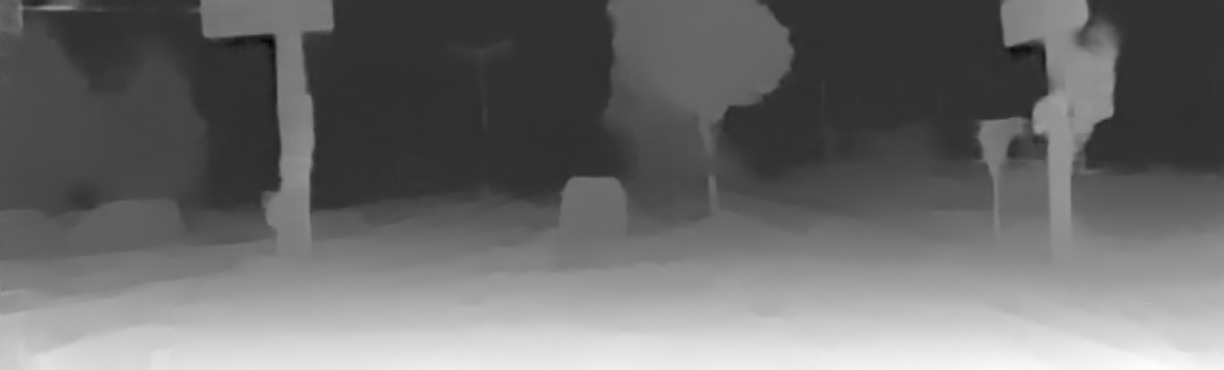}} &
    \includegraphics[width=.24\linewidth]{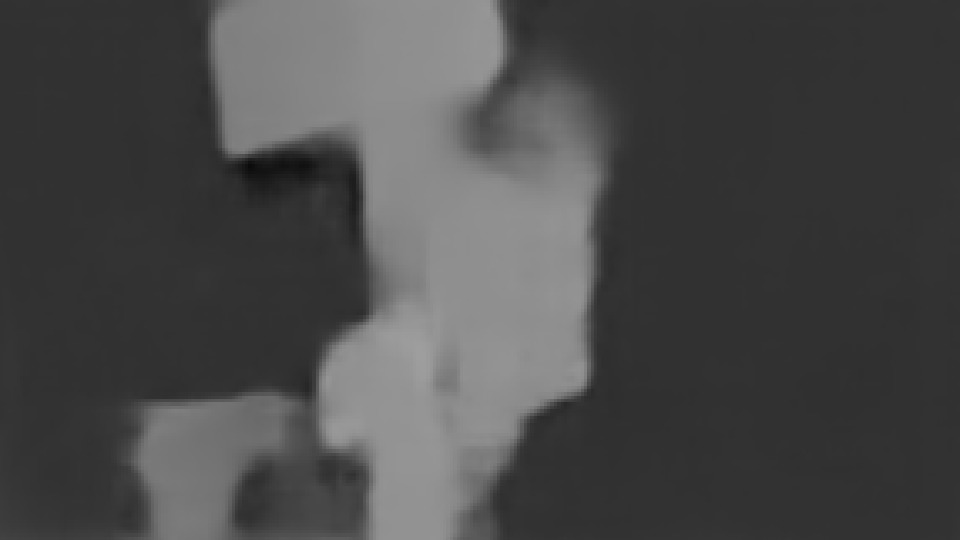} &
    \includegraphics[width=.24\linewidth]{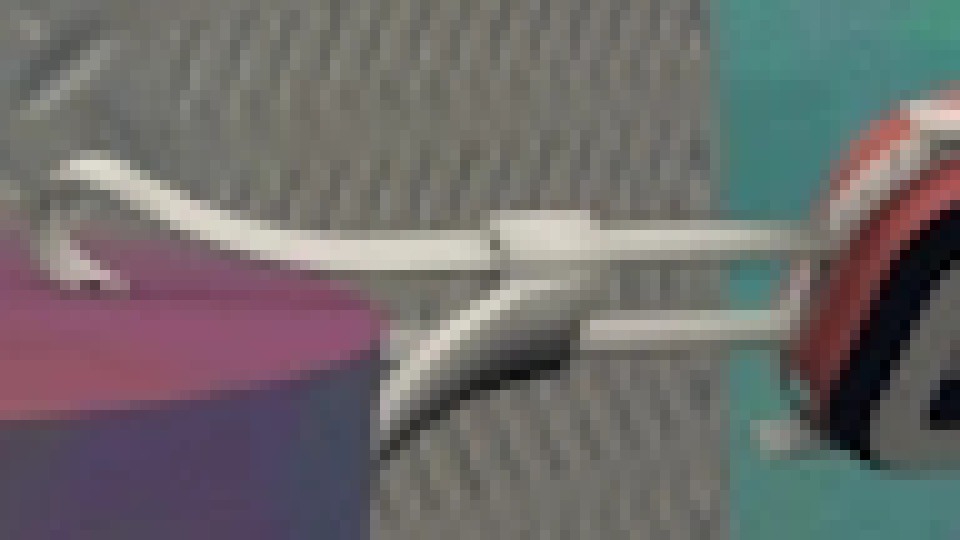} \\
    \multicolumn{2}{c}{Disparity estimate trained on FlyingThings3D} &
    Crop from disparity map &
    Crop from FlyingThings3D \\[2mm]
    
    \multicolumn{2}{c}{\includegraphics[width=.49\linewidth,trim=0 32 0 0,clip]{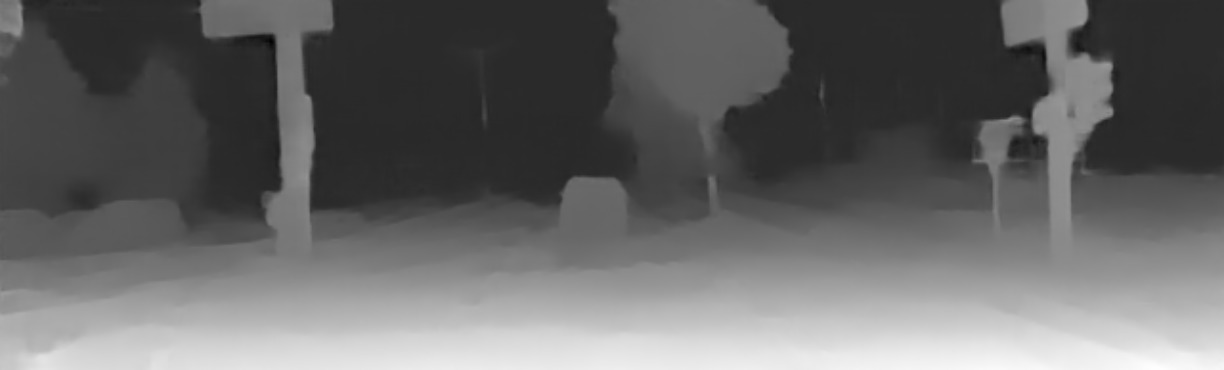}} &
    \includegraphics[width=.24\linewidth]{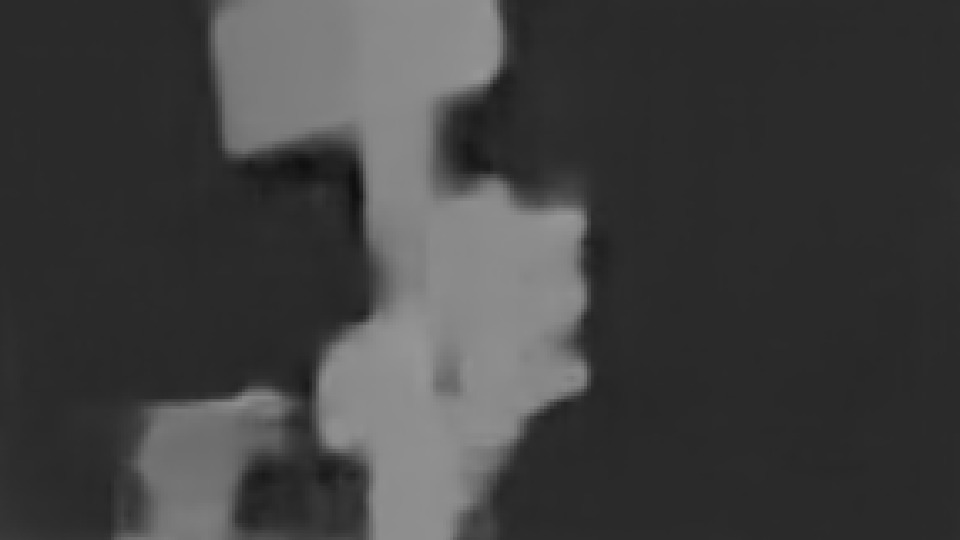} &
    \includegraphics[width=.24\linewidth]{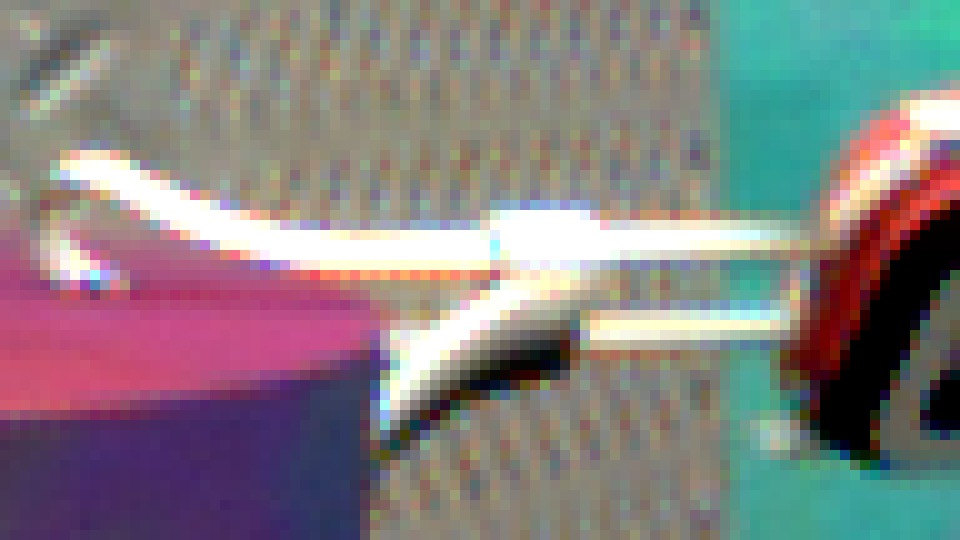} \\
    \multicolumn{2}{c}{Disparity estimate trained on degraded FlyingThings3D} &
    Crop from disparity map &
    Crop from degraded data \\[2mm]
    
    \multicolumn{2}{c}{\begin{tabular}[t]{c}%
                          \fbox{\includegraphics[width=.49\linewidth,trim=0 32 0 0,clip]{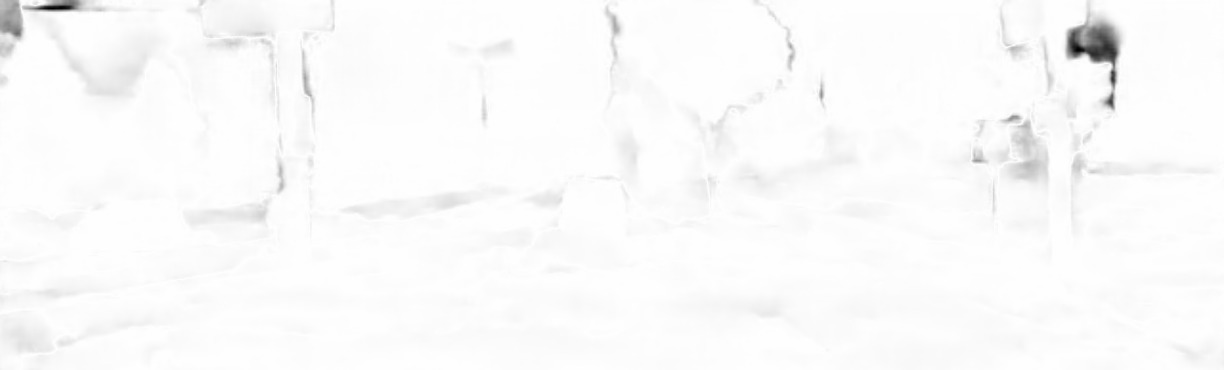}} \\
                          Difference image of disparity maps
                       \end{tabular}} &
    \multicolumn{2}{c}{\raisebox{.75\height}{\begin{tabular}{p{2.5cm}p{2.5cm}p{2.5cm}}%
                          \ijcvhruletop
                          \multicolumn{3}{c}{Test on KITTI2015 (training set), trained on...} \\[0.4mm]
                          FT3D & FT3D+deg & (KITTI2012) \\
                          \ijcvhrulemid
                          $1.62$ & $1.54$ & $(1.28)$ \\
                          \ijcvhrulebot \\
                        \end{tabular}}} \\
  \end{tabular}%
  \caption{\textbf{Bayer-interpolation artifacts in training data.} Color fringing (first row) from naive Bayer-pattern interpolation poses problems to a DispNet trained+finetuned on clean synthetic FlyingThings3D data (second row). Finetuning on a purposefully degraded version of the same dataset instead helps the network deal with this effect (third row). The disparity difference image highlights that the two networks’ outputs differ mostly along object contours where the artifacts have the greatest impact. The accompanying table shows that degraded training data produces better results than the artifact-free original data. The (rarely possible) best-case scenario of training with ground truth data from the target domain still yields the best results; this is expected as this experiment only targets low-level data characteristics, and FlyingThings3D data (degraded or not) cannot teach disparity priors for KITTI’s street scenes.}%
  \label{fig:datadeg-kitti}%
\end{figure*}%

\subsubsection{Bayer-interpolation artifacts}
Imaging sensors can generally sense only light intensity, not color.
To get color images from such a sensor, the most common strategy is to cover the sensor with a repeating pattern of color filters, such that each pixel becomes sensitive to a certain part of the color spectrum.
The classic way to do this is the Bayer pattern.
To reconstruct a full-color image, each pixel’s color channels must be inferred from within a neighborhood, which can lead to significant artifacts.
To test whether a network can be preconditioned to deal with Bayer-interpolation artifacts, we emulated them on the FlyingThings3D dataset by first simulating how a Bayer sensor would perceive the scene and then interpolating an RGB image from this virtual sensor image.
We evaluated the effects by finetuning a DispNet on FlyingThings3D data with or without Bayer artifacts and then testing the network on the KITTI 2015 training set.

Fig.~\ref{fig:datadeg-kitti} shows examples of our original and degraded data, as well as the effects on the network output.
Our degraded data improves the KITTI 2015 score by 5\%. This shows that adding these effects helps, but from the qualitative results one might expect a larger difference.
This is because the KITTI ground truth is sparse and has gaps along object contours, due to the nature of its acquisition method using a laser scanner and the perspective difference between the scanner and the camera.
A difference image between the disparity estimates of our two synthetic-data networks reveals that most changes actually occur at object boundaries, and the highlighted crops are located in an image region for which there is no ground truth at all.

\section{Conclusion}

In this paper, we performed a detailed analysis of the synthetic datasets for deep network training that we introduced in earlier conference papers for optical flow and disparity estimation. This analysis led to several findings, of which we summarize the most important ones here: (1)~Diversity is important. A network trained on specialized data generalizes worse to other datasets than a network trained on diverse data. (2)~Realism is overrated.  Most of the learning task can be accomplished via simplistic data and data augmentation. Realistic effects, such as sophisticated lighting models, are not important to learn basic optical flow, but merely induce minor improvements. (3)~Learning schedules matter. Learning schedules that combine multiple different datasets, especially simpler and more complex ones, are a way to greatly improve the generic performance of the trained networks. (4)~Camera knowledge helps. Modeling distortions of the camera in the training data largely improves the network’s performance when run with this camera. 

According to our evaluation, these findings are valid for optical flow (1--3) and disparity estimation~(4).
They may not hold for high-level tasks such as object recognition. With an increasing necessity to jointly solve low-level and high-level tasks, we see a major future challenge in designing datasets that can serve both regimes and in finding learning procedures that can efficiently combine the advantages of real-world and synthetic data.

\bibliographystyle{spbasic}
\bibliography{bibliography}

%% Set rowcolor to start "impossibly" late because it interferes with the multirow rotatebox
\rowcolors{100}{gray!20}{white}%
\setlength{\tabcolsep}{5pt}%
\begin{figure*}%
  \begin{center}%
    \begin{tabular}{ccccc}%
      \multirow{2}{*}{\rotatebox[origin=c]{90}{Boxes}} &
      \includegraphics[width=0.22\linewidth]{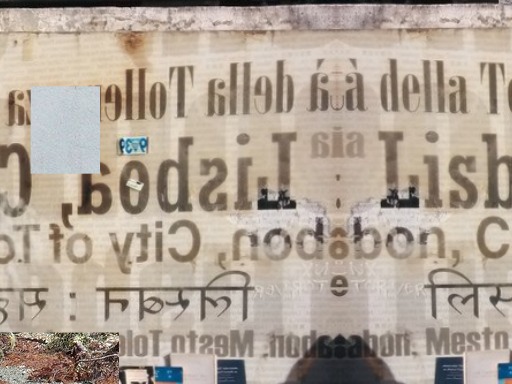} &
      \includegraphics[width=0.22\linewidth]{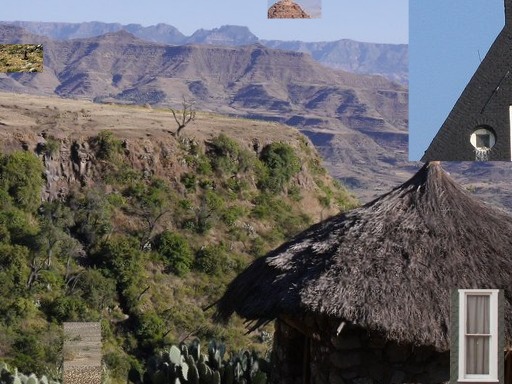} &
      \includegraphics[width=0.22\linewidth]{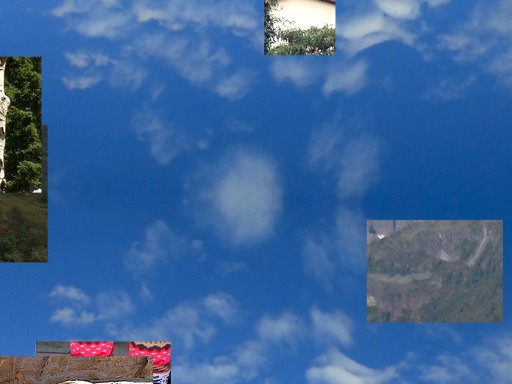} &
      \includegraphics[width=0.22\linewidth]{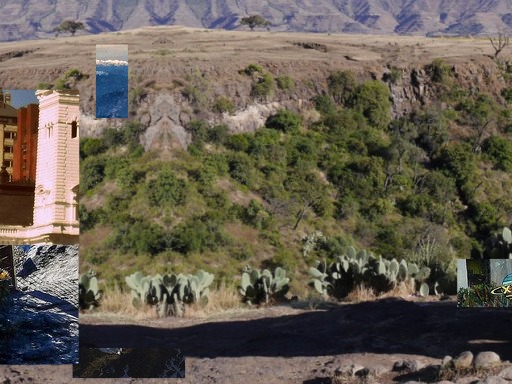} \\
      &
      \includegraphics[width=0.22\linewidth]{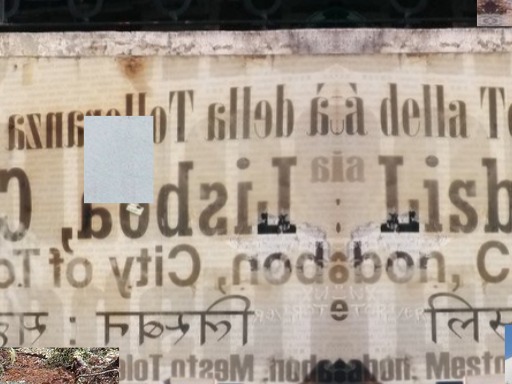} &
      \includegraphics[width=0.22\linewidth]{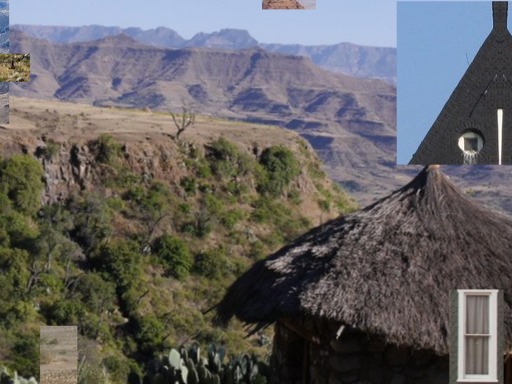} &
      \includegraphics[width=0.22\linewidth]{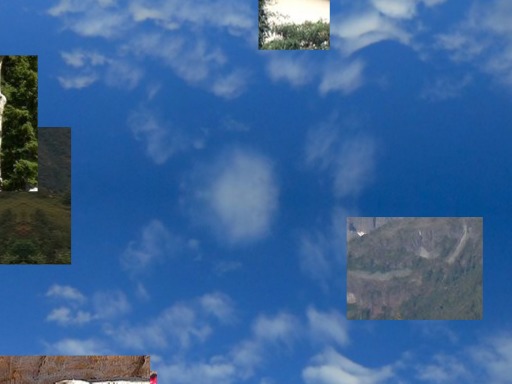} &
      \includegraphics[width=0.22\linewidth]{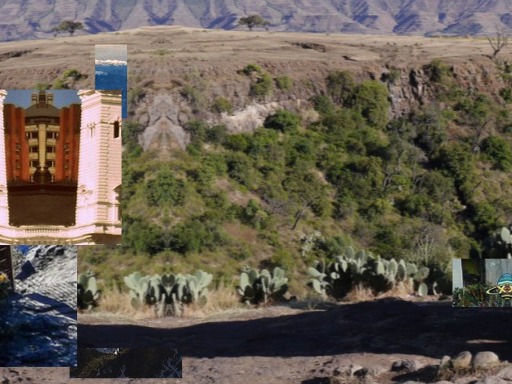} \\[-1.25mm]
      
      \ijcvhrulemid
      
      \multirow{2}{*}{\rotatebox[origin=c]{90}{Polygons}} &
      \includegraphics[width=0.22\linewidth]{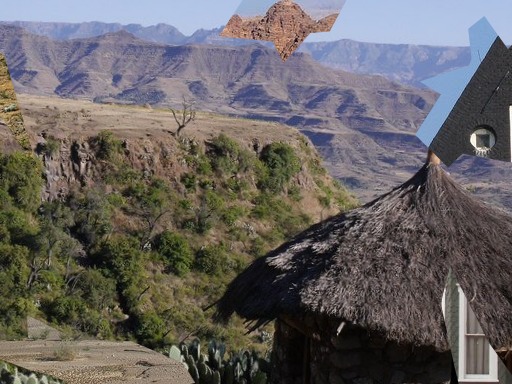} &
      \includegraphics[width=0.22\linewidth]{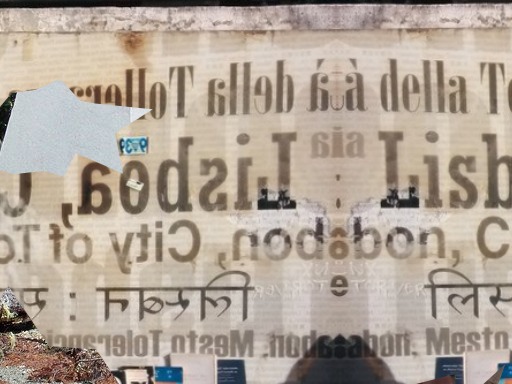} &
      \includegraphics[width=0.22\linewidth]{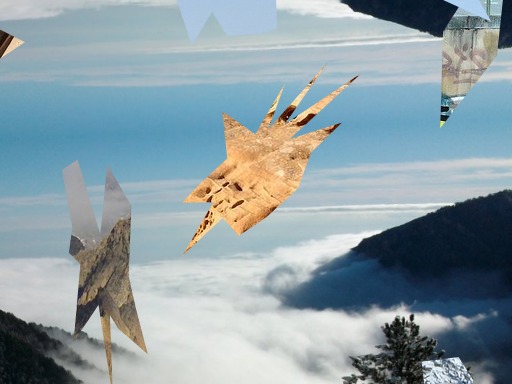} &
      \includegraphics[width=0.22\linewidth]{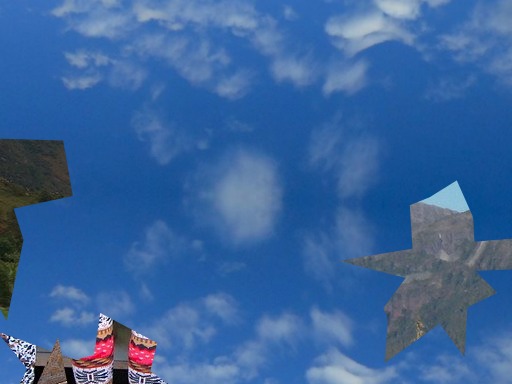} \\
      &                                       
      \includegraphics[width=0.22\linewidth]{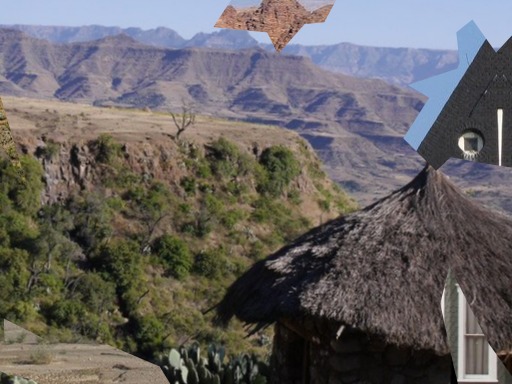} &
      \includegraphics[width=0.22\linewidth]{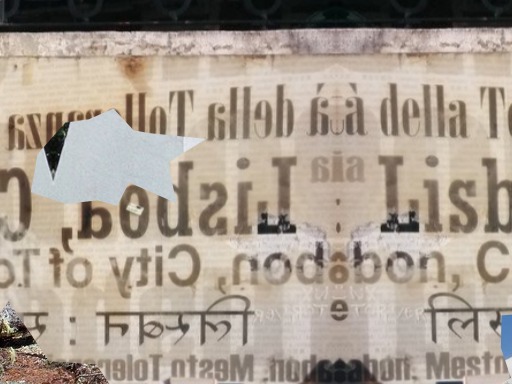} &
      \includegraphics[width=0.22\linewidth]{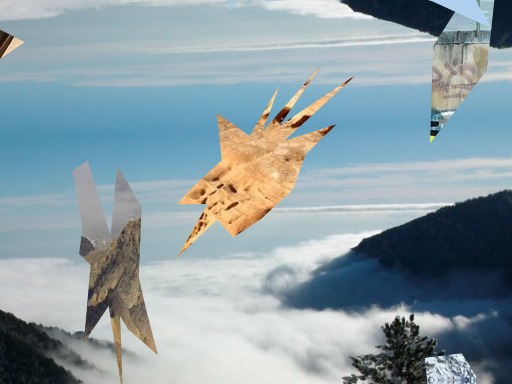} &
      \includegraphics[width=0.22\linewidth]{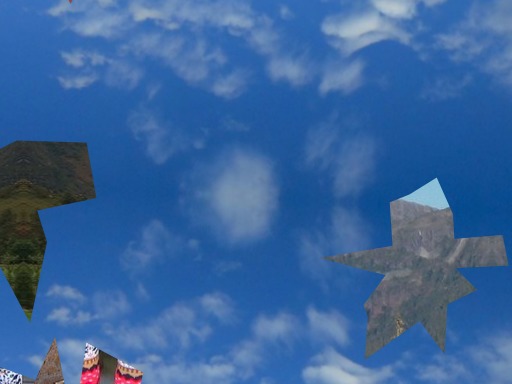} \\[-1.25mm]
      
      \ijcvhrulemid
      
      \multirow{2}{*}{\rotatebox[origin=c]{90}{Ellipses}} &
      \includegraphics[width=0.22\linewidth]{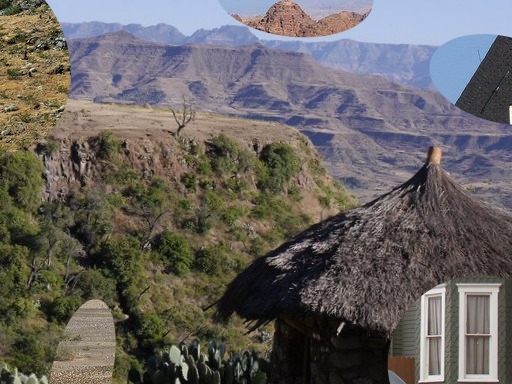} &
      \includegraphics[width=0.22\linewidth]{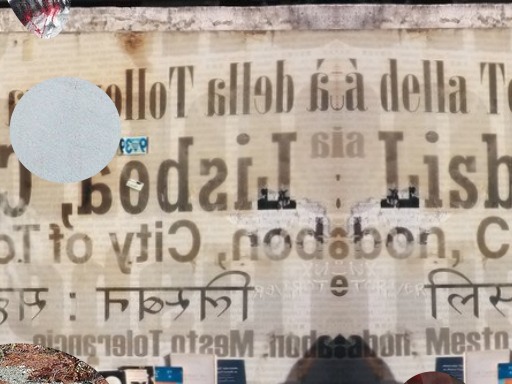} &
      \includegraphics[width=0.22\linewidth]{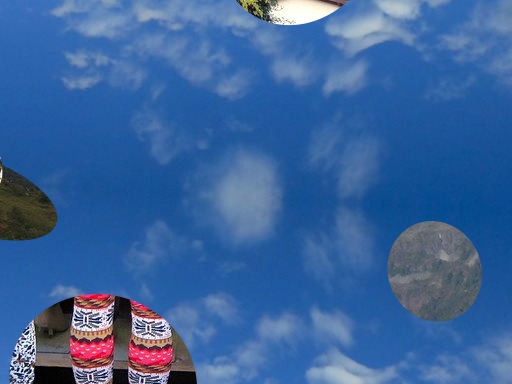} &
      \includegraphics[width=0.22\linewidth]{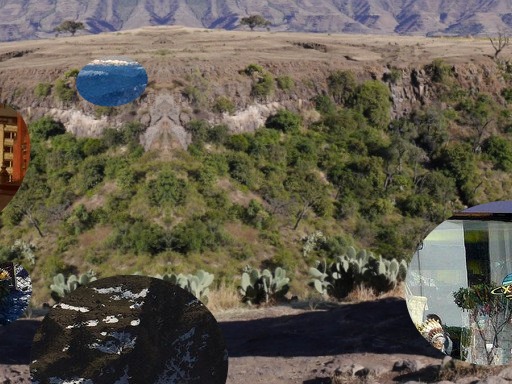} \\
      &                                       
      \includegraphics[width=0.22\linewidth]{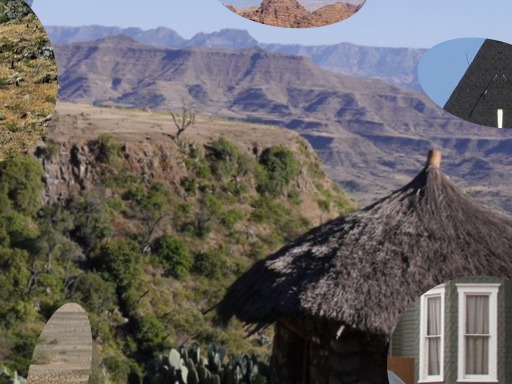} &
      \includegraphics[width=0.22\linewidth]{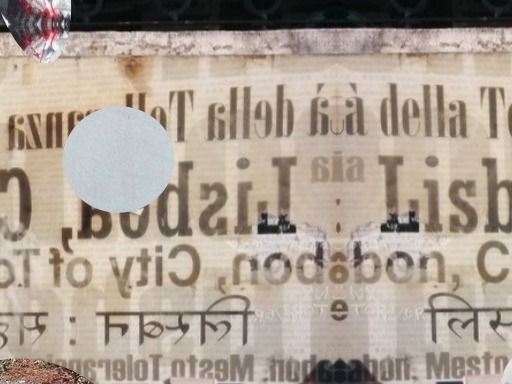} &
      \includegraphics[width=0.22\linewidth]{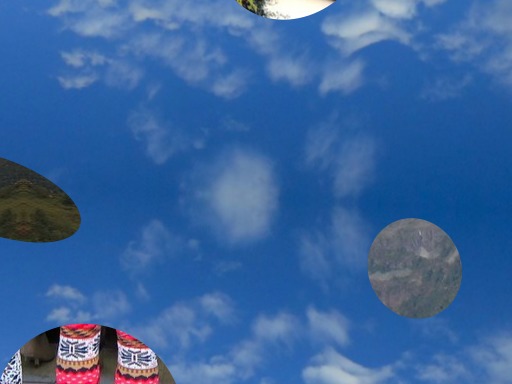} &
      \includegraphics[width=0.22\linewidth]{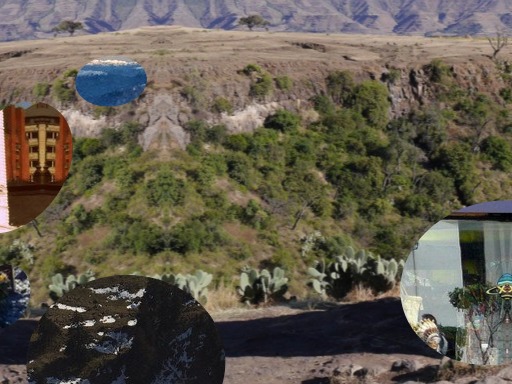} \\
    \end{tabular}%
  \end{center}%
  \caption{\textbf{Dataset gallery 1/4:} Boxes, Polygons and Ellipses variants as used in Sec.~\protect\ref{sec:shape-and-motion} (see also Table~\protect\ref{tab:data-variation-types}).}%
  \label{fig:gallery-1}%
\end{figure*}%

\begin{figure*}%
  \begin{center}%
    \begin{tabular}{ccccc}%
      \multirow{2}{*}{\rotatebox[origin=c]{90}{Polygon+Ellipses}} &
      \includegraphics[width=0.22\linewidth]{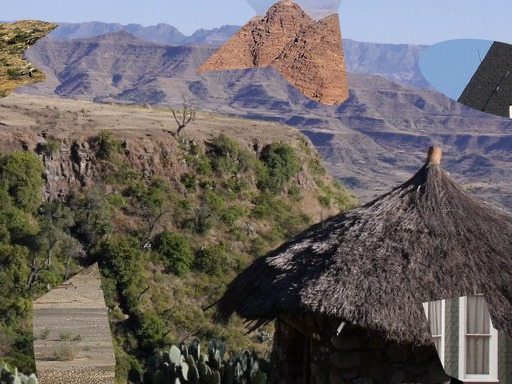} &
      \includegraphics[width=0.22\linewidth]{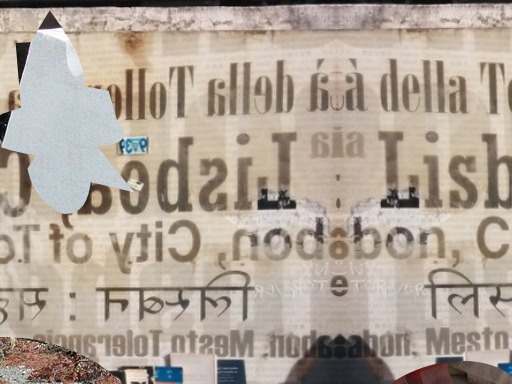} &
      \includegraphics[width=0.22\linewidth]{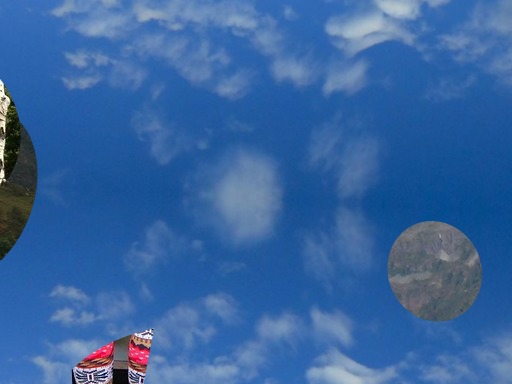} &
      \includegraphics[width=0.22\linewidth]{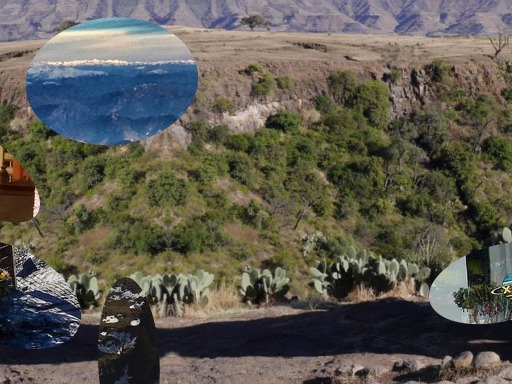} \\
      &
      \includegraphics[width=0.22\linewidth]{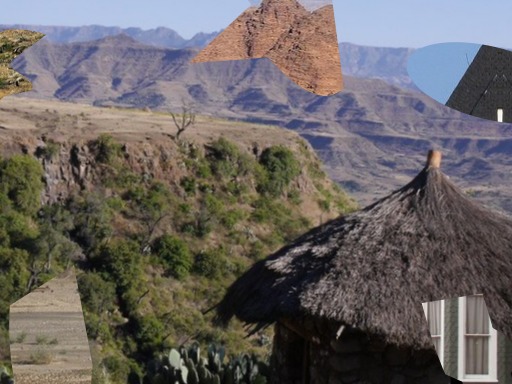} &
      \includegraphics[width=0.22\linewidth]{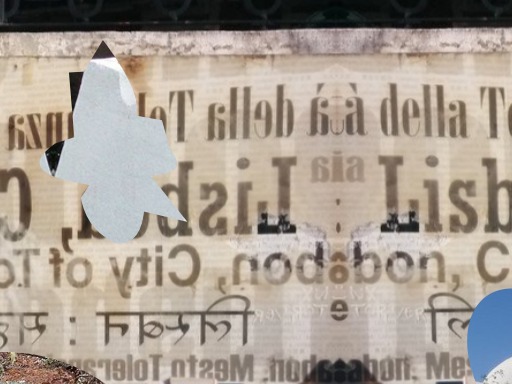} &
      \includegraphics[width=0.22\linewidth]{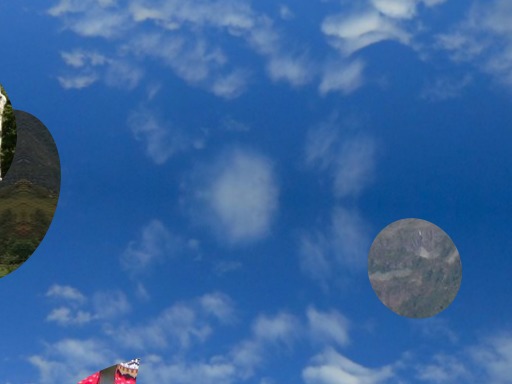} &
      \includegraphics[width=0.22\linewidth]{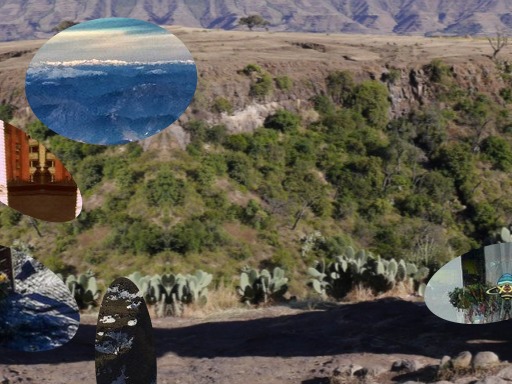} \\[-1.25mm]
      
      \ijcvhrulemid
      
      \multirow{2}{*}{\rotatebox[origin=c]{90}{+Deformations}} &
      \includegraphics[width=0.22\linewidth]{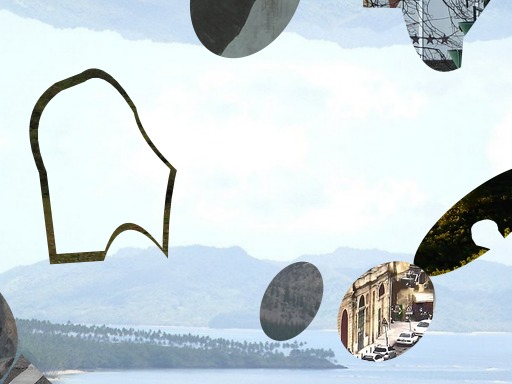} &
      \includegraphics[width=0.22\linewidth]{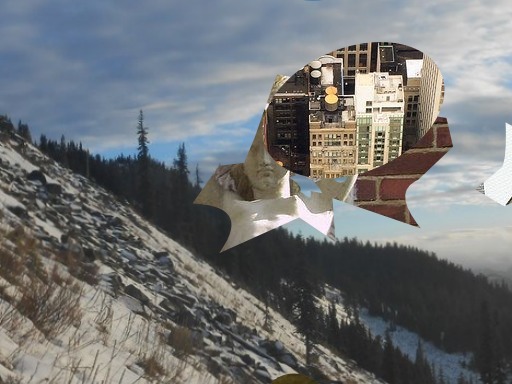} &
      \includegraphics[width=0.22\linewidth]{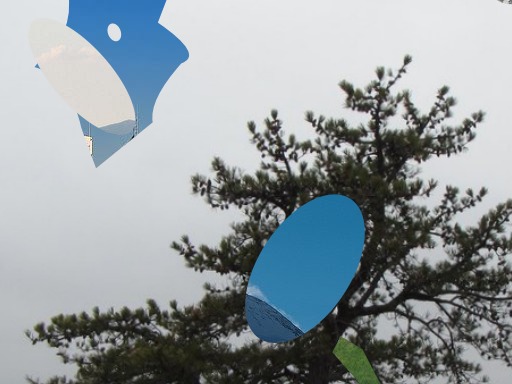} &
      \includegraphics[width=0.22\linewidth]{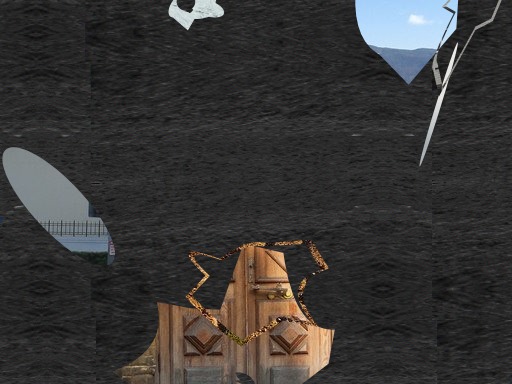} \\
      &                                       
      \includegraphics[width=0.22\linewidth]{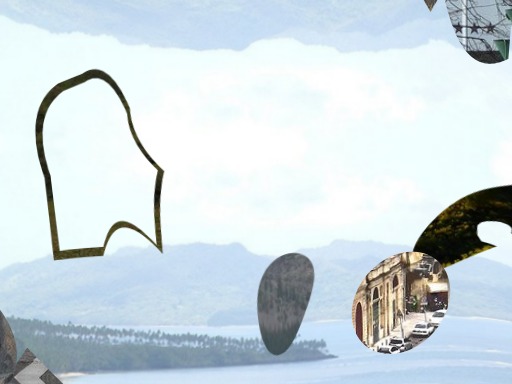} &
      \includegraphics[width=0.22\linewidth]{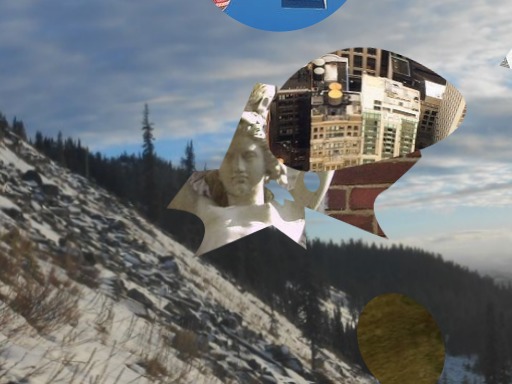} &
      \includegraphics[width=0.22\linewidth]{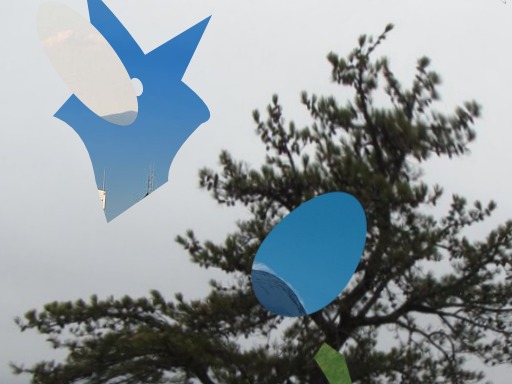} &
      \includegraphics[width=0.22\linewidth]{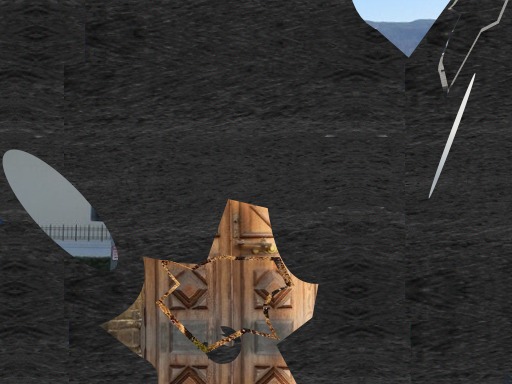} \\[-1.25mm]
      
      \ijcvhrulemid
      
      \multirow{2}{*}{\rotatebox[origin=c]{90}{“Plasma” textures}} &
      \includegraphics[width=0.22\linewidth]{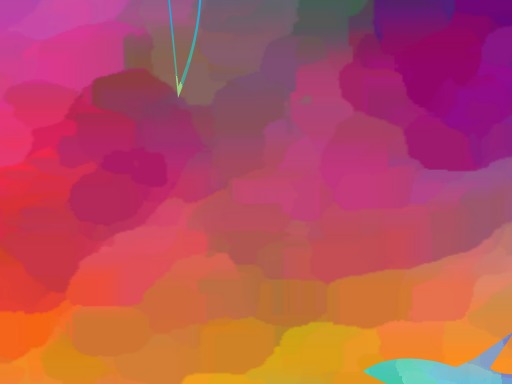} &
      \includegraphics[width=0.22\linewidth]{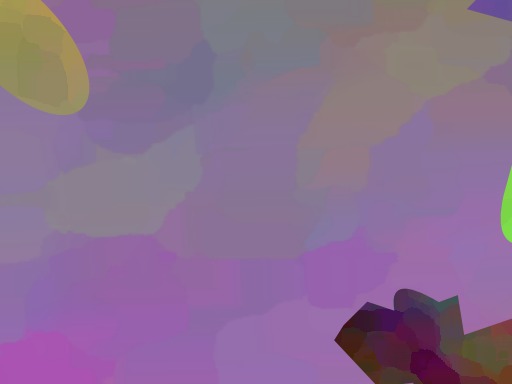} &
      \includegraphics[width=0.22\linewidth]{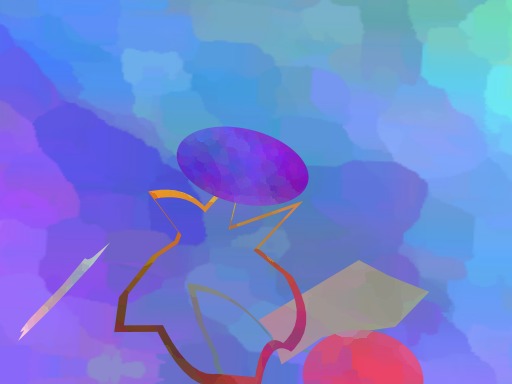} &
      \includegraphics[width=0.22\linewidth]{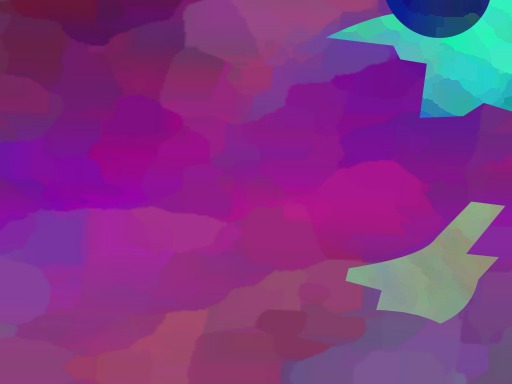} \\
      &                                   
      \includegraphics[width=0.22\linewidth]{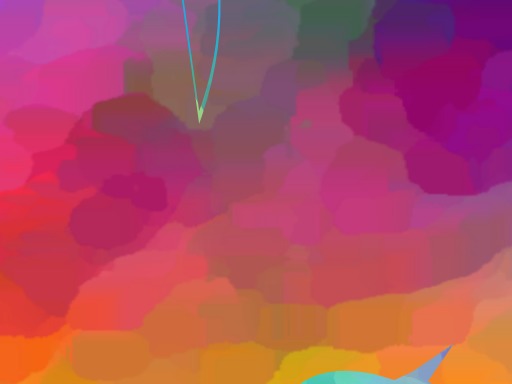} &
      \includegraphics[width=0.22\linewidth]{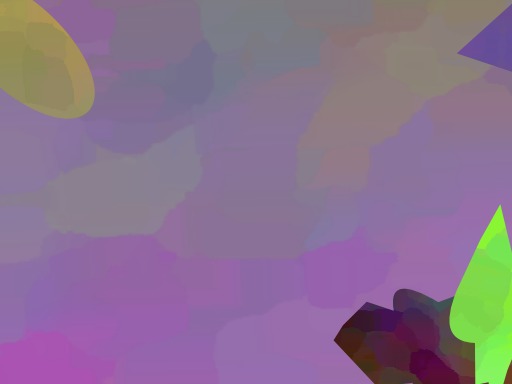} &
      \includegraphics[width=0.22\linewidth]{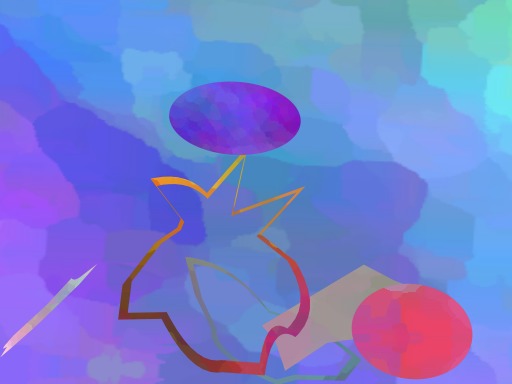} &
      \includegraphics[width=0.22\linewidth]{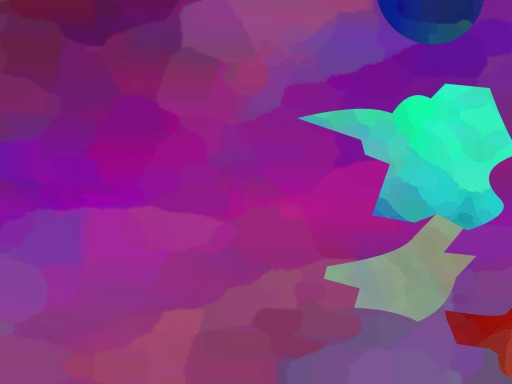} \\
    \end{tabular}%
  \end{center}%
  \caption{\textbf{Dataset gallery 2/4:} Polygons+Ellipses and Polygons+Ellipses+Deformations variants as used in Sec.~\protect\ref{sec:shape-and-motion} (see also Table~\protect\ref{tab:data-variation-types}); “Plasma” textures as used in Sec.~\protect\ref{sec:textures}.}%
  \label{fig:gallery-2}%
\end{figure*}%

\begin{figure*}%
  \begin{center}%
    \begin{tabular}{ccccc}%
      \multirow{2}{*}{\rotatebox[origin=c]{90}{“Clouds” textures}} &
      \includegraphics[width=0.22\linewidth]{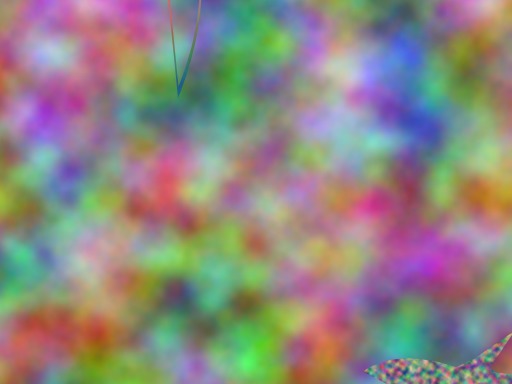} &
      \includegraphics[width=0.22\linewidth]{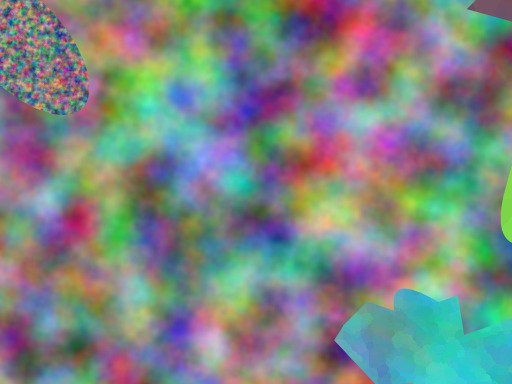} &
      \includegraphics[width=0.22\linewidth]{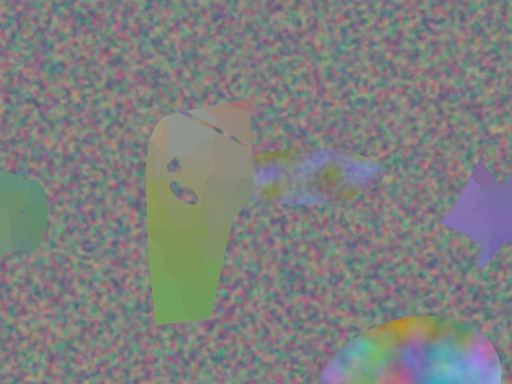} &
      \includegraphics[width=0.22\linewidth]{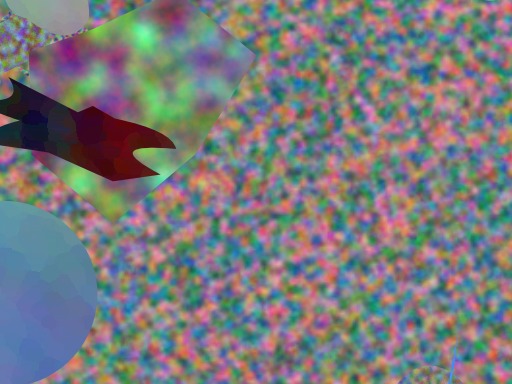} \\
      &                                   
      \includegraphics[width=0.22\linewidth]{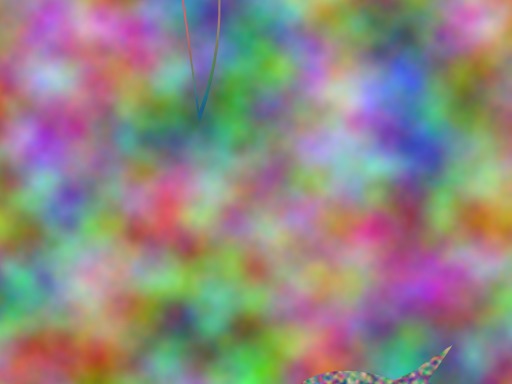} &
      \includegraphics[width=0.22\linewidth]{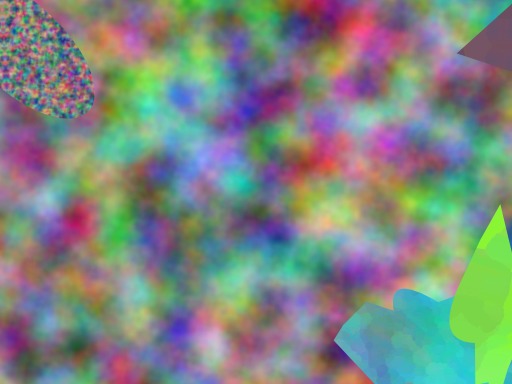} &
      \includegraphics[width=0.22\linewidth]{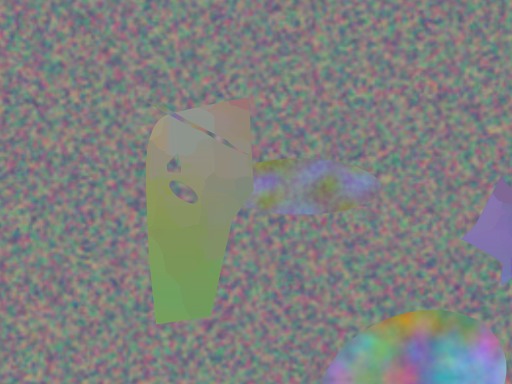} &
      \includegraphics[width=0.22\linewidth]{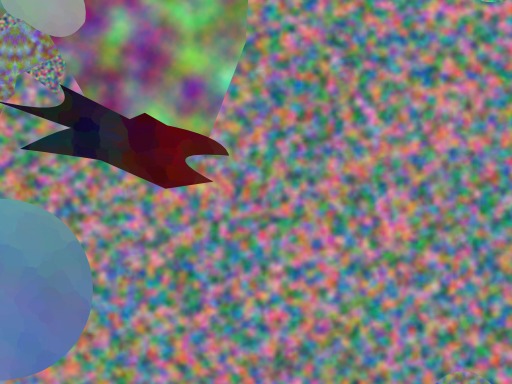} \\
      
      \ijcvhrulemid
      
      \multirow{2}{*}{\rotatebox[origin=c]{90}{FlyingChairs}} &
      \includegraphics[width=0.22\linewidth]{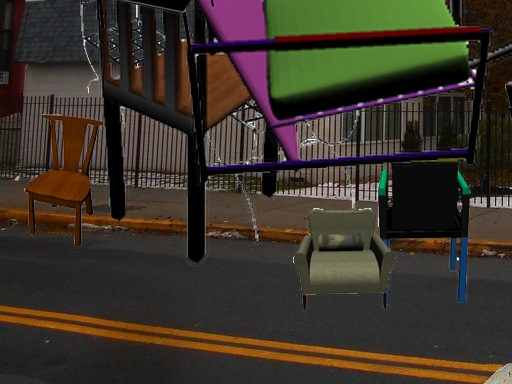} &
      \includegraphics[width=0.22\linewidth]{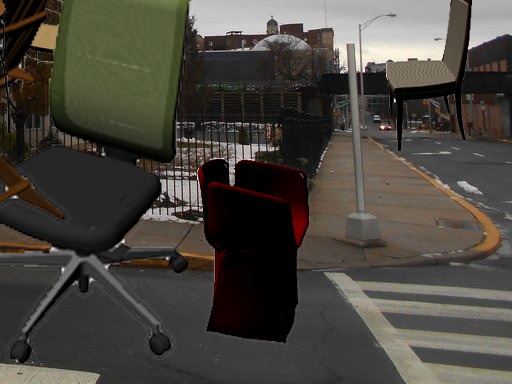} &
      \includegraphics[width=0.22\linewidth]{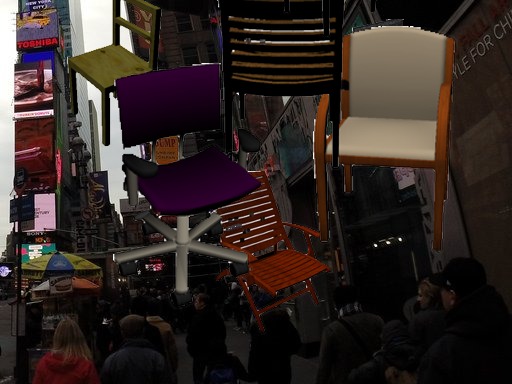} &
      \includegraphics[width=0.22\linewidth]{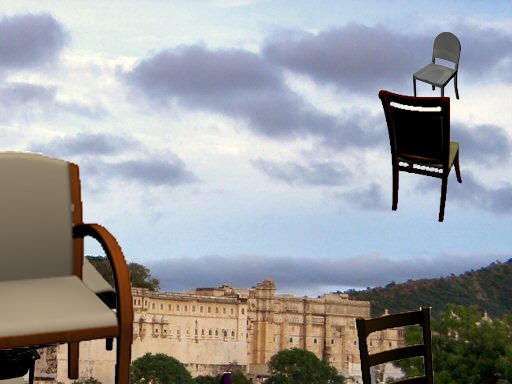} \\
      &  
      \includegraphics[width=0.22\linewidth]{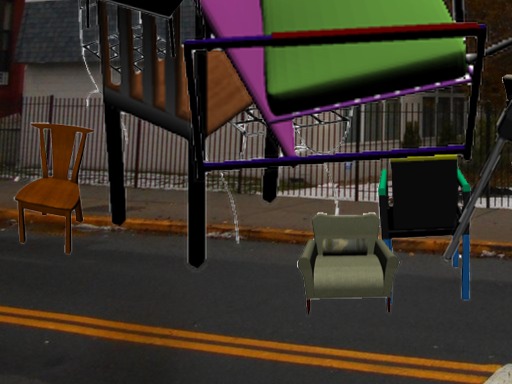} &
      \includegraphics[width=0.22\linewidth]{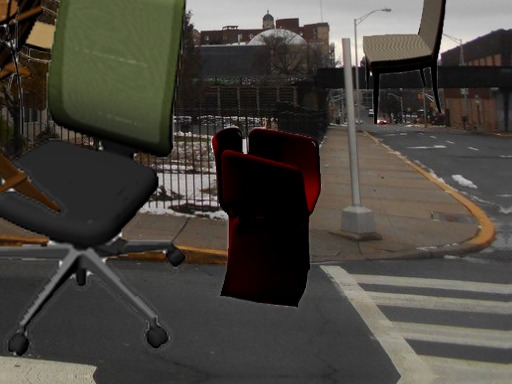} &
      \includegraphics[width=0.22\linewidth]{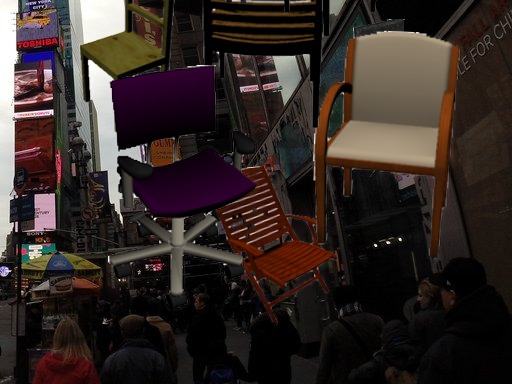} &
      \includegraphics[width=0.22\linewidth]{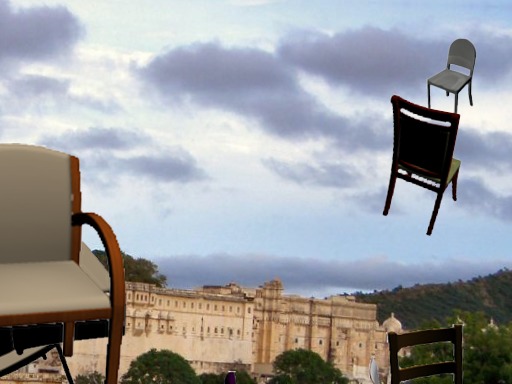} \\[-1.25mm]
      
      \ijcvhrulemid
      
      \multirow{2}{*}{\rotatebox[origin=c]{90}{FlyingThings3D}} &
      \includegraphics[width=0.22\linewidth]{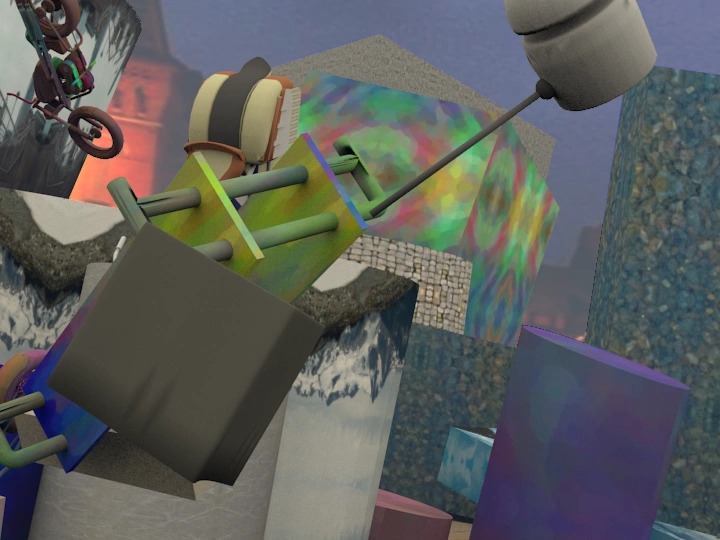} &
      \includegraphics[width=0.22\linewidth]{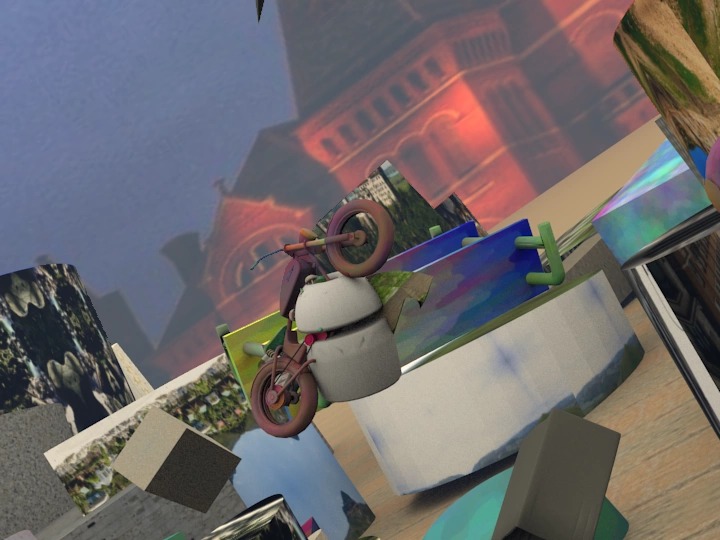} &
      \includegraphics[width=0.22\linewidth]{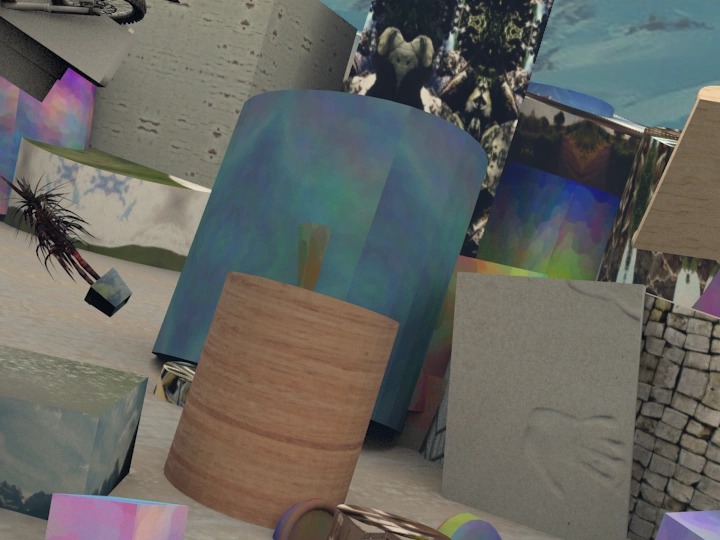} &
      \includegraphics[width=0.22\linewidth]{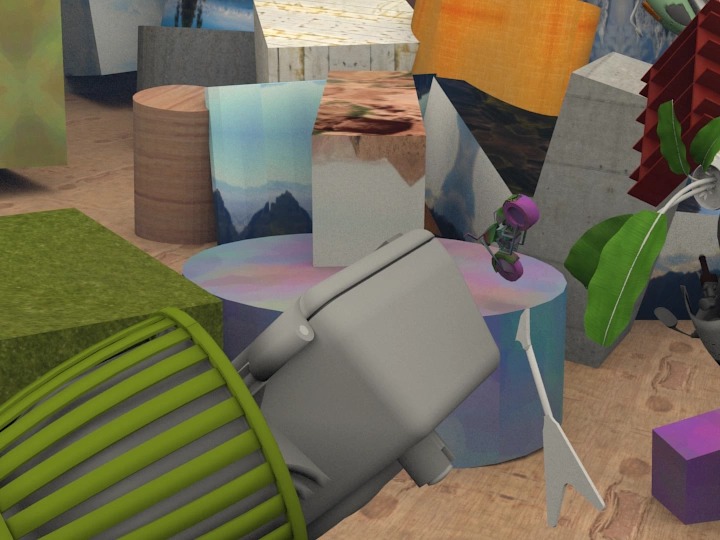} \\
      &                                       
      \includegraphics[width=0.22\linewidth]{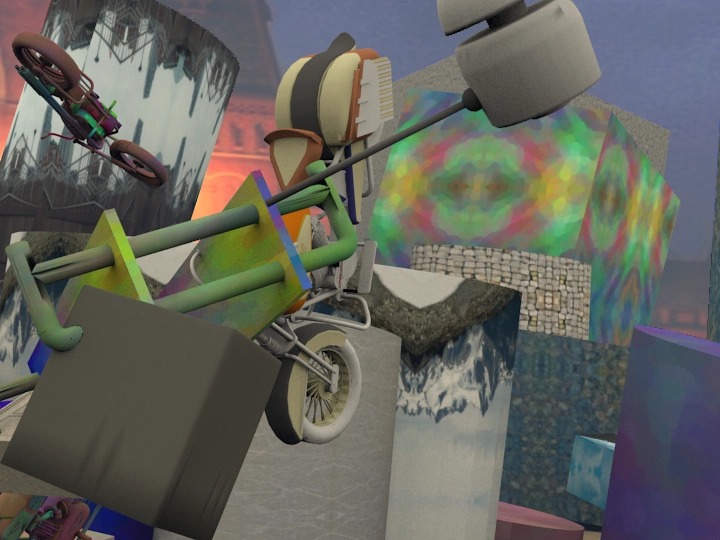} &
      \includegraphics[width=0.22\linewidth]{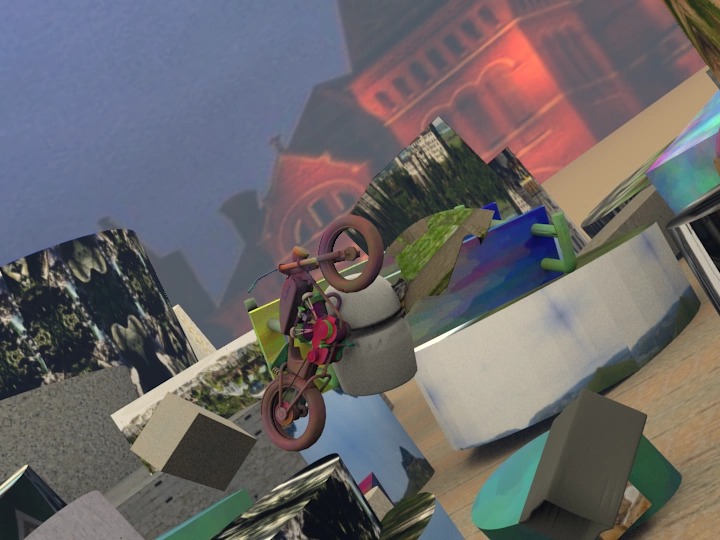} &
      \includegraphics[width=0.22\linewidth]{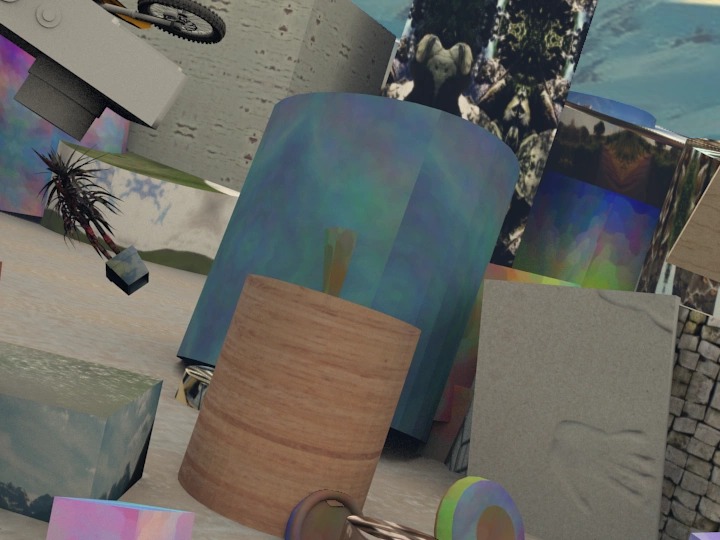} &
      \includegraphics[width=0.22\linewidth]{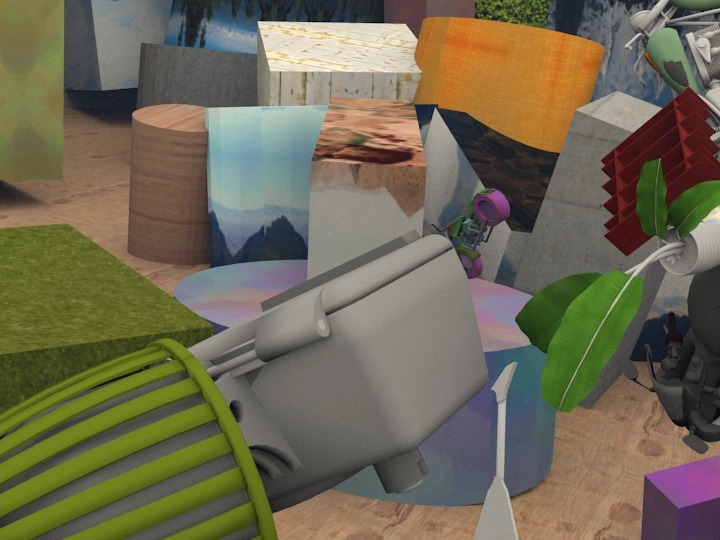} \\[-1.25mm]
    \end{tabular}%
  \end{center}%
  \caption{\textbf{Dataset gallery 3/4:} “Clouds” textures as used in Sec.~\protect\ref{sec:textures}; FlyingChairs from \citet{flownet}; FlyingThings3D (cropped to the same 4:3 aspect ratio as the other datasets for display purposes) from \citet{dispnet}.}%
  \label{fig:gallery-3}%
\end{figure*}%

\begin{figure*}%
  \begin{center}%
    \begin{tabular}{ccccc}%
      \multirow{2}{*}{\rotatebox[origin=c]{90}{Shadeless}} &
      \includegraphics[width=0.22\linewidth]{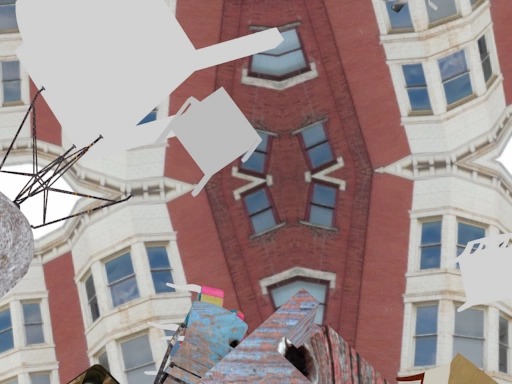} &
      \includegraphics[width=0.22\linewidth]{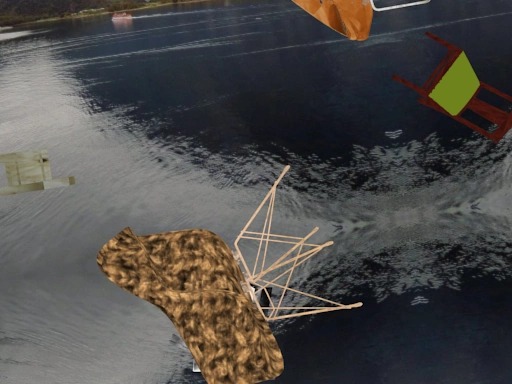} &
      \includegraphics[width=0.22\linewidth]{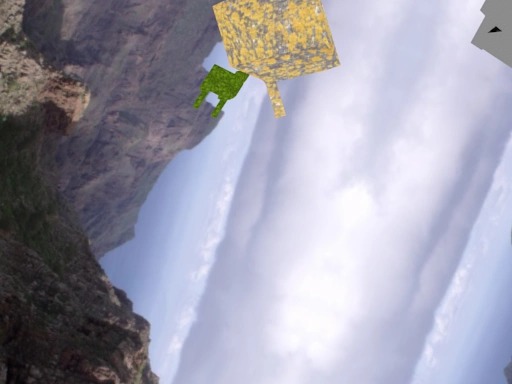} &
      \includegraphics[width=0.22\linewidth]{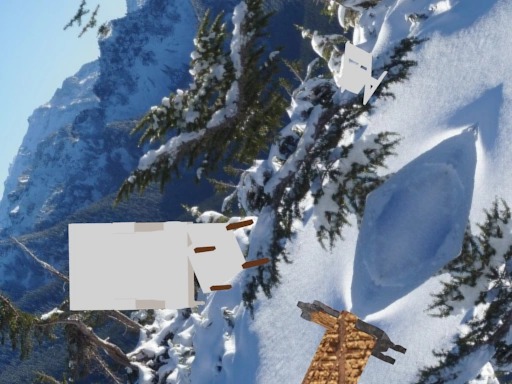} \\
      &
      \includegraphics[width=0.22\linewidth]{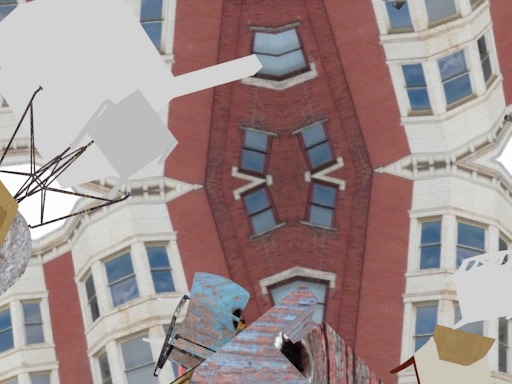} &
      \includegraphics[width=0.22\linewidth]{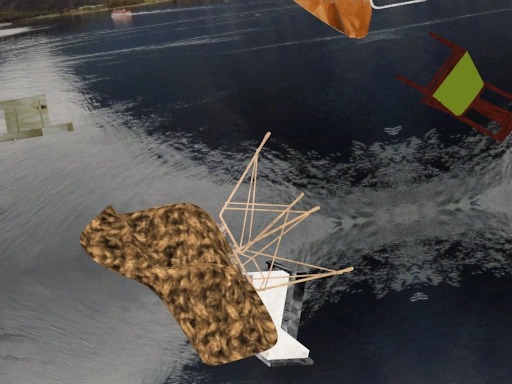} &
      \includegraphics[width=0.22\linewidth]{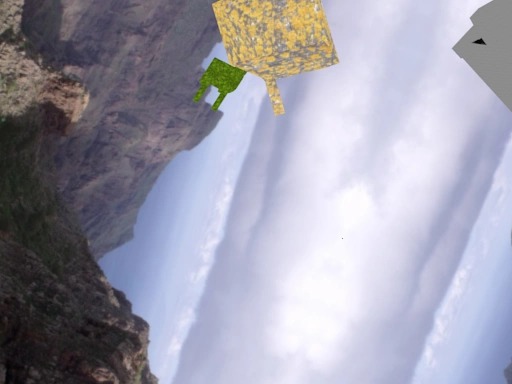} &
      \includegraphics[width=0.22\linewidth]{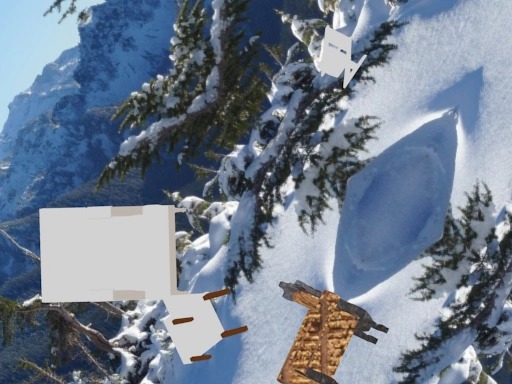} \\[-1.25mm]
      
      \ijcvhrulemid
      
      \multirow{2}{*}{\rotatebox[origin=c]{90}{Static}} &
      \includegraphics[width=0.22\linewidth]{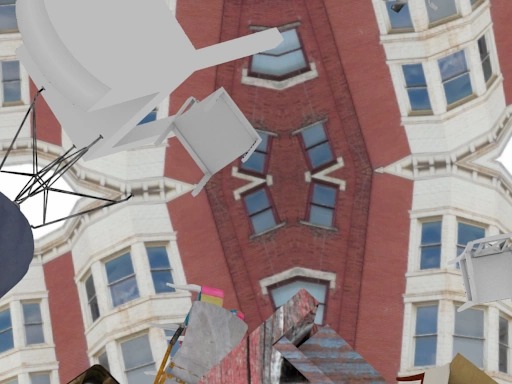} &
      \includegraphics[width=0.22\linewidth]{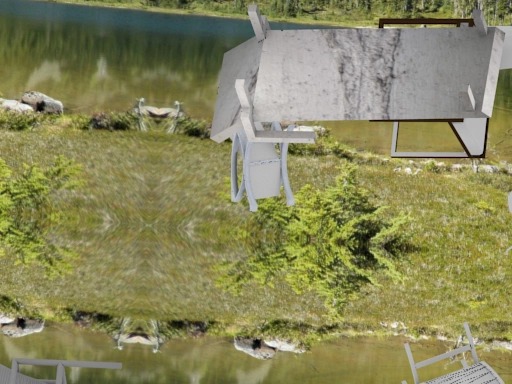} &
      \includegraphics[width=0.22\linewidth]{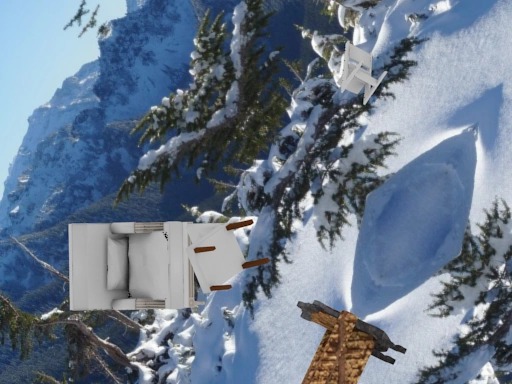} &
      \includegraphics[width=0.22\linewidth]{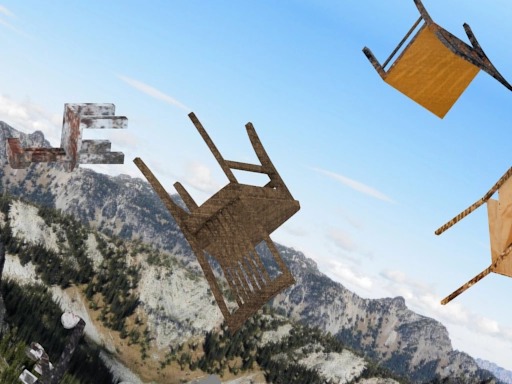} \\
      &
      \includegraphics[width=0.22\linewidth]{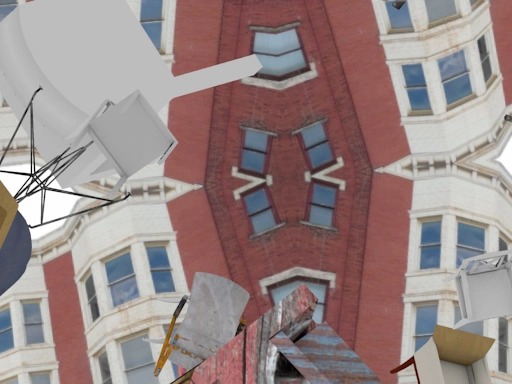} &
      \includegraphics[width=0.22\linewidth]{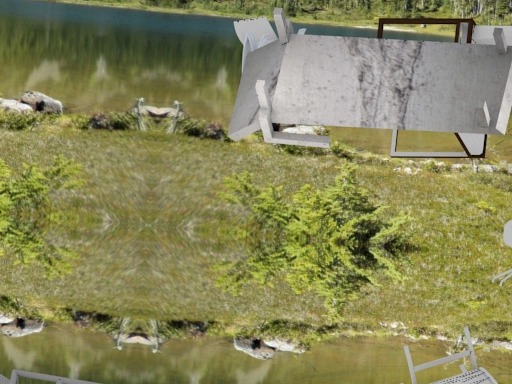} &
      \includegraphics[width=0.22\linewidth]{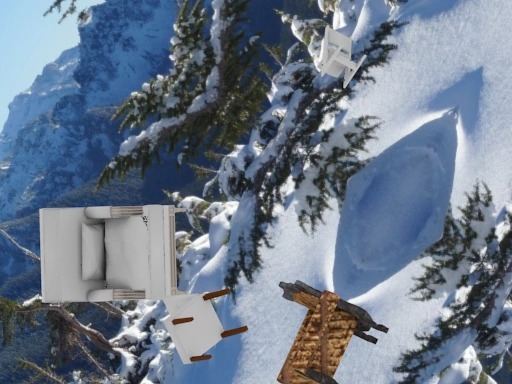} &
      \includegraphics[width=0.22\linewidth]{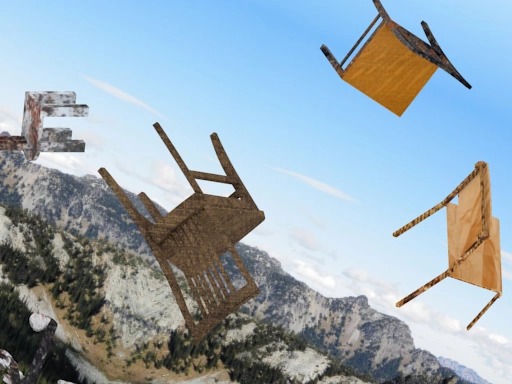} \\[-1.25mm]
      
      \ijcvhrulemid
      
      \multirow{2}{*}{\rotatebox[origin=c]{90}{Dynamic}} &
      \includegraphics[width=0.22\linewidth]{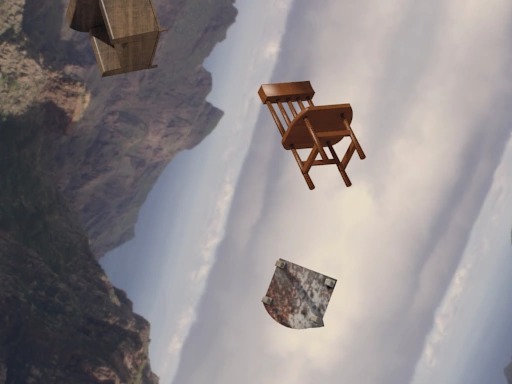} &
      \includegraphics[width=0.22\linewidth]{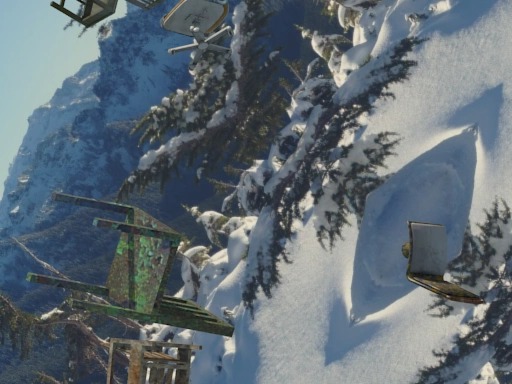} &
      \includegraphics[width=0.22\linewidth]{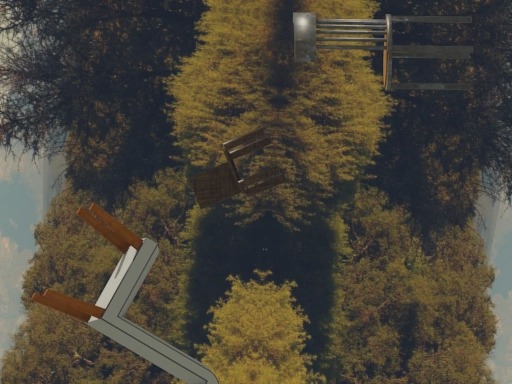} &
      \includegraphics[width=0.22\linewidth]{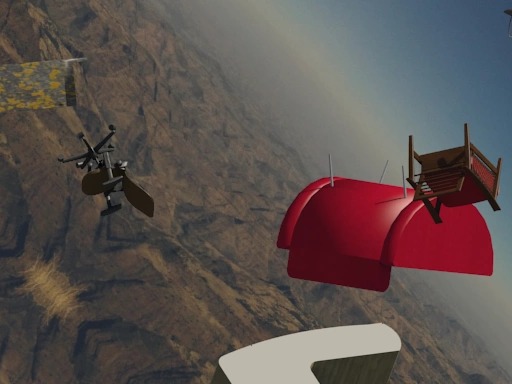} \\
      &
      \includegraphics[width=0.22\linewidth]{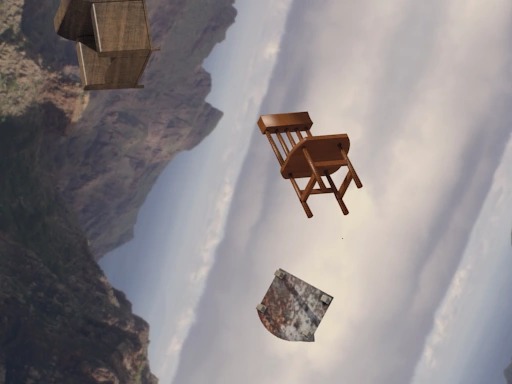} &
      \includegraphics[width=0.22\linewidth]{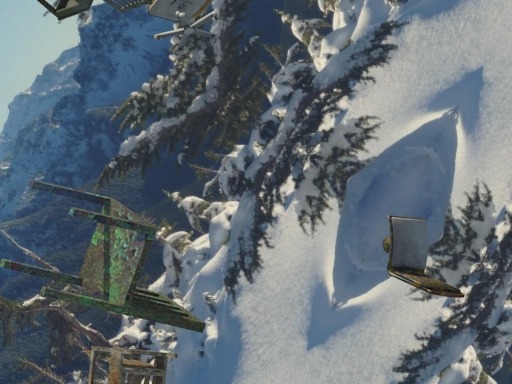} &
      \includegraphics[width=0.22\linewidth]{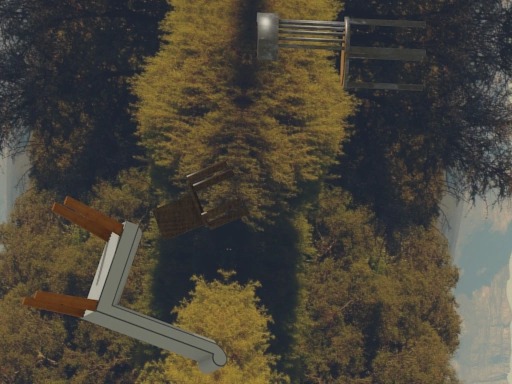} &
      \includegraphics[width=0.22\linewidth]{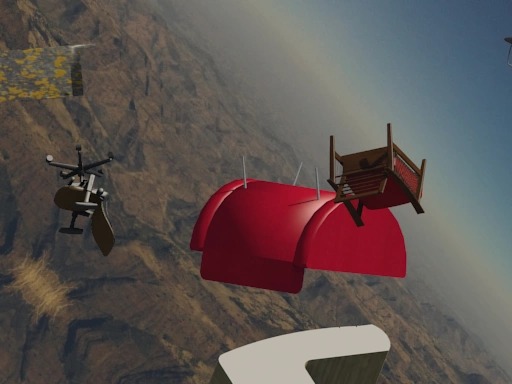} \\
    \end{tabular}%
  \end{center}%
  \caption{\textbf{Dataset gallery 4/4:} Shadeless, static and dynamic lighting variants as used in Sec.~\protect\ref{sec:lightingquality}.}%
  \label{fig:gallery-4}%
\end{figure*}%

\end{document}